**Title:** Modular magnetic soft robotic catheter enables visualized intervention in the pancreatobiliary tree

**Authors:**
B. Calmé*, N. J. Greenidge, A. Metcalf, A. Bacchetti, G. Loza, D. Kpeglo, P. Lloyd, V. Pensabene, J. H. Chandler, P. Valdastri

**Affiliations:**
STORM Lab, University of Leeds, Leeds, United Kingdom.
School of Electronic and Electrical Engineering and Pollard Institute, University of Leeds, Leeds, United Kingdom.
Leeds Institute of Medical Research at St James' University Hospital, University of Leeds, Leeds, United Kingdom.

*Corresponding author. Email: b.p.calme@leeds.ac.uk

**Abstract:** Diagnosis and therapy in the pancreatobiliary tree are limited by access. Guidewires reach deep into the ducts but are blind, whereas cholangioscopes provide vision but are too large to readily enter the pancreatic body and tail, the cystic duct, or the gallbladder. At the millimeter scale of these ducts, every function added to a steerable catheter competes with steering for cross-section and adds bending stiffness. We present a 1.47-mm modular magnetic soft robotic catheter in which each function is built into the magnetic body, onto its surface, or onto the footprint of the camera, and we measured what each function costs in bending under bedside fields. Imaging and fiber Bragg grating (FBG) shape sensing accounted for most of this cost, whereas a compliant gripper, magnetic anchoring balloons, and a micrometer-thick ultrasound-triggered release shell together reduced the tip angle at 8 mT only from 70° to 64°. The anchoring balloon increased the gripping force about threefold without stiffening the catheter during navigation. In live pigs, the robotic catheter performed operator-initiated task-autonomous cannulation of the native papilla without sphincterotomy and was teleoperated under onboard vision 75 mm into the pancreatic duct and, without a guidewire, through the cystic duct into the gallbladder, whose lumen it imaged. Anchoring, grasping, and payload release were each demonstrated in vivo. Adding each function at a measured cost to steering allows a soft robot small enough for these ducts to intervene as well as navigate.

**One-Sentence Summary:** A 1.47-mm magnetic soft robotic catheter carries vision, anchoring, and tools to the pancreatic duct and gallbladder in vivo.

## INTRODUCTION

The pancreatobiliary system remains one of the most difficult anatomical regions to reach with endoluminal tools. Pancreatic cancer is now the sixth leading cause of cancer-related death worldwide, with a five-year survival rate of approximately 10% [1], and biliary tract cancers and gallbladder disease account for a large population of patients who require diagnosis or therapy deep within small, tortuous ducts [2]. In endoscopic retrograde cholangiopancreatography (ERCP), a side-viewing duodenoscope is positioned in front of the papilla, a small and mobile natural orifice, and a catheter or guidewire is used to cannulate the ducts selectively. Selective biliary cannulation succeeds in roughly 90%–95% of native papilla cases [3], [4], but difficult cannulation is associated with post-ERCP pancreatitis [3], [5], which occurs in about 10% of patients overall and 15% of those at high risk [6], and with cholangitis [5], [7]. Cannulation is also acquired slowly: trainees need between 70 and 400 supervised procedures to reach 80% success [8]. Robotic cannulation could make its outcome less dependent on operator experience.

Beyond the papilla, current tools force a trade-off between reach and information. Guidewires are thin enough to follow the ducts but provide no visualization, relying on fluoroscopy and tactile feedback. Single-operator cholangioscopes such as the SpyScope (Boston Scientific, USA) provide direct visualization and a working channel, but their insertion is often facilitated by sphincterotomy [9], and they do not readily enter the pancreatic body and tail or the cystic duct. The gallbladder illustrates this gap: because no endoscopic route routinely reaches its lumen, lesions of its wall are characterized only by ultrasound and cross-sectional imaging, and cholecystectomy is recommended for polypoid lesions of 10 mm or more [10]. Yet in one series 97% of gallbladders removed for suspected polyps contained no neoplastic lesion [11], whereas gallbladder cancers are often found only after cholecystectomy for presumed benign disease [12]. Transpapillary drainage of the gallbladder requires a guidewire to traverse the cystic duct under fluoroscopy, without direct visualization, with a pooled technical success of about 83% [13], and direct visualization of the gallbladder lumen through the papilla has been reported mainly in case reports and a small series, usually with a cholangioscope advanced over a guidewire already placed in the gallbladder [14], [15], [16], [17]. No current tool combines the diameter of a guidewire with onboard vision, active steering, and the ability to act at the target.

Soft robotic catheters are well placed to fill this gap. Their compliance allows safe interaction with delicate tissue [18], [19], and magnetic soft continuum robots (mSCRs), in which hard-magnetic microparticles such as NdFeB (neodymium-iron-boron) are embedded in an elastomer, can be programmed for complex shapes [20], [21], [22], [23] and steered wirelessly, without the friction, hysteresis, and backlash that limit tendon-driven, concentric tube, and fluidic catheters at small diameters [24], [25], [26], [27]. Because the field is generated outside the patient, no steering mechanism occupies the catheter, and robotic control of the field allows closed-loop execution of repeatable steps such as cannulation. However, most mSCRs have been demonstrated as navigation devices in the neurovascular [28], [29], [30], pulmonary [31], and reproductive [32] systems, and functions such as laser ablation [31], in situ biosensing [33], printing [34], or microrobot deployment [35], [36] have been added one at a time. At the millimeter scale imposed by the anatomy, function and steerability compete for the same cross-section: each cable, fiber, lumen, or shell displaces magnetic material and adds bending stiffness, and a module added as a separate non-magnetic segment shortens the magnetized length. Tools that must exert force at the tip also lose much of their reaction force through a compliant body, and the conventional remedy, a stiffer body or sheath, removes the compliance that justified a soft catheter. A 0.95-mm fiberscopic robot has shown that imaging and tools can share one magnetic body ex vivo [37], but its imaging and light-guide fibers stiffen the body, and fiber-bundle imaging gives pixelated images that may be poorly suited to fluid-filled ducts. A tendon–magnetic coactuated concentric robot enlarged the reachable workspace in bile and pancreatic duct phantoms, but addressed steering rather than action at the target [38]. Reducing the diameter alone is not sufficient: each function must be added at a measured and limited cost to steerability (Suppl. 1).

Here, we present a modular mSCR with a nominal outer diameter of 1.47 mm for access and intervention in the pancreatobiliary tree, and we measured the bending cost of each function by adding the modules to the magnetic body in sequence. Onboard imaging, fiber Bragg grating (FBG) shape sensing, a tendon-driven compliant gripper, magnetic anchoring balloons, and an ultrasound-triggered release shell were designed following five rules. First, functions are actuated wirelessly where possible: steering by the external magnetic field and payload release by ultrasound. Second, a module that occupies length along the body is made from the magnetized material itself: the anchoring balloons are molded and magnetized with the catheter body. Third, other modules use space that is already occupied or add only a surface layer: the gripper is bonded to the camera housing, and the release shell is a micrometer-thick coating. Fourth, stiffness is added only at the target and only when the task requires it: the magnetic balloon grounds the distal catheter on demand. Fifth, the actuation of each module is kept apart from the others in force or frequency: the gripper flexures minimize the tendon tension that would displace the tip, and the release trigger is placed away from the imaging frequencies and from the natural frequencies of the gripper blades. Autonomy is restricted to papilla cannulation, the step with the clearest clinical rationale, and is implemented as task autonomy [39], [40]: an operator-initiated subroutine performs papilla alignment and insertion, an independent FBG shape sensor monitors agreement between the measured shape and the magnetic model, and the operator retains override authority. Navigation within the ducts was teleoperated from the onboard camera view.

We validated the platform in live porcine models along the steps of the procedure (movie S1). The mSCR performed task-autonomous cannulation of the native papilla without sphincterotomy, teleoperated navigation 75 mm into the pancreatic duct, a region in which about 15–20% of pancreatic adenocarcinomas arise [41], and, to our knowledge, the first guidewire-free, magnetically steered transpapillary access to the gallbladder, with onboard visualization of its lumen. Anchoring, grasping, and payload release were demonstrated in vivo as separate steps.

## RESULTS

### Procedural requirements set the functionalization budget

We derived the design of the mSCR from the steps of the pancreatobiliary procedure (Fig. 1). The catheter must be deployed through the working channel of a standard endoscope, selectively cannulate the papilla, and then navigate ducts whose diameter decreases from the common bile duct and pancreatic head toward the pancreatic body and tail and the cystic duct, typically about 3.3, 2.4, and 1.5 mm in the head, body, and tail of the main pancreatic duct and about 2.5 mm in the cystic duct (Fig. 1F). Within these ducts, the operator needs to see the lumen, choose between branches, stabilize the catheter at the target, and act on the tissue or deliver a payload. These needs translate into six requirements (Table S1): (R1) an outer diameter compatible with endoscope working channels and with the target ducts; (R2) sufficient magnetic steerability to align with the papilla and select branches under the fields achievable at the patient's bedside; (R3) compliance during navigation, so that contact with the papilla and the duct wall remains gentle; (R4) onboard visualization to complement blind, fluoroscopy-dependent advancement; (R5) distal stabilization and the ability to perform a therapeutic or diagnostic action at the target; and (R6) compatibility with standard endoscopes and with the bedside workflow. These requirements conflict: imaging, sensing, and tools (R4, R5) consume the cross-section and stiffness budget left by R1 and R2, and the force needed at the target (R5) conflicts with the compliance needed on the way (R3). For each conflict, Table 1 lists the conventional solution, its cost, and the design decision taken here.

Each module serves one or more requirements, and its design is constrained by the others (Fig. 1, C to E). Building upon prior work on mSCR design and control [31], [42], a scalable low-pressure injection molding process (Materials and Methods) produces a monolithic magneto-elastic body that integrates a 100-μm channel for flushing, aspiration, and insufflation, a balloon lumen, and guides for a miniature camera with LED illumination, a multicore FBG, and an 80-μm nylon tendon, all within the 1.47-mm outer diameter (Fig. 2, A and B, movie S2). The camera housing carries a compliant gripper (R5), the body incorporates magnetizable anchoring balloons (R3, R5), and a patterned polylactic acid (PLA) shell loaded with the payload is bonded around the catheter (R5). Colored markers near the tip support visual tracking during cannulation (R2). The configuration can be adapted to the task: for access to the gallbladder, for example, the gripper was omitted in some trials to maximize the camera field of view.

Magnetic actuation was achieved by a large external permanent magnet (EPM) mounted onto a 7-DoF robotic manipulator positioned at the patient's bedside (Fig. 1A). The N52 EPM (101.6 mm in diameter and length) gives 15.7 mT on its axis 18 cm from its face and 7.8 mT at 24 cm. Its face was kept 18 to 24 cm from the catheter, about 10 cm or more above the abdominal wall for ducts about 8 cm deep. Software-defined motion limits keep the EPM out of the patient space (Fig. 2C and Fig. S1); because the field is generated outside the patient, the endoscopic workflow is unchanged (R6).

To implement these limits, the reachable workspace of the robotic manipulator, defined as the set of positions in space that can be reached by its end-effector, is constrained using a magnetic-feasible workspace, adapted from the magnetic workspace of electromagnetic navigation systems [43] and defined as the set of positions where a desired set of magnetic tasks is feasible for a given end-effector position. The typical desired task was $D = \{\boldsymbol{B} \in \mathbb{R}^3 | 8 \text{ mT} \leq |\boldsymbol{B}| \leq 16 \text{ mT}\}$, which can be achieved within a workspace of 0.05 m³ with field strength up to $|\boldsymbol{B}| = 16$ mT on the mSCR, sufficient to apply a magnetic torque at the base of the mSCR of up to $\tau = 3.4 \times 10^{-4}$ N·m (Fig. 2C).

Within this workspace, the pose of the EPM sets the shape of the catheter (Fig. 2D). With the EPM beside the catheter and its dipole pointing at the body, the nearly uniform field bends the catheter into a C-shape, and moving the EPM up or down changes the curvature; turning the dipole along the body reverses the field direction between base and tip, which can produce an S-shape with an inflection point set by the EPM height (Fig. 2D, Fig. S2 and S3). With a remanence of 0.31 T and the component stiffnesses of Table S2, the model of the mSCR with gripper and magnetic balloon (configuration 6a) gives tip angles of 66° at 8 mT and 80° at 16 mT for a field perpendicular to its base. In the model, rotating the field to 150° raises the tip angle to 104° and 131° and the lateral reach to 50 and 51 mm (Fig. 2E). Beyond about 90°, the modeled catheter has two equilibrium shapes, so these angles depend on the path of the field. The measured effect of each component is given in the next section.

A single joystick coordinates all functions (Fig. 2F): the stick steers the catheter in either endoscope-view or mSCR-camera teleoperation, and two buttons drive the automated advancing unit for feed and retraction, the gripper, and the balloon through a computer-controlled insufflation system. All subsystems run under Robot Operating System (ROS), enabling modular control and synchronized data streaming. The workflow was validated on five endoscopes across two distinct magnetic actuation setups (Fig. S2, S3 and Table S3).

## Adding function at a measured cost to steering

At this scale, the non-magnetic components (a camera cable of about 0.4 mm, an LED fiber of 0.25 mm, a multicore FBG of 0.2 mm, the 80-µm tendon, the 100-µm channel, and a balloon lumen of about 0.2 mm; Fig. 2B) compete with the magnetic material for the 1.7-mm² cross-section: the camera cable, LED fiber, and FBG alone occupy 12% of it but carry 66% of the bending stiffness of the composite (Table 1, Suppl. 2). To measure the cost of each function, we added the components to the magnetic body in sequence and measured the tip angle in uniform fields from 8 to 25 mT (Table 1; n = 3, N = 15 per configuration and field). Because the catheter approaches alignment with the field at 16 mT, the cost is compared at 8 mT. The magnetic body alone bent by 86.0° at 8 mT. Imaging and sensing accounted for most of the loss: the camera cable and LED fiber reduced the tip angle to 80.1°, and the FBG to 69.7°; together, imaging and sensing account for 75% of the total loss at 8 mT. Because the bare body is close to alignment with the field even at 8 mT, the field needed for a 60° tip angle, computed with the model (Suppl. 2), gives a clearer measure: it rose from 1.8 mT for the magnetic body to 3.1 mT with camera and LED and 5.4 mT with the FBG, and reached 6.3 mT for configuration 6a. This loss is the price of onboard vision (R4) and of the independent shape measurement used to monitor task-autonomous cannulation.

Each module of the final design added after imaging and sensing reduced the tip angle at 8 mT by 4.6° or less (Table 1). First, the anchoring balloons are molded and magnetized with the catheter body, so that, when deflated, they keep the magnetization continuous and remain compatible with shape forming [44]. In the single-axis coil (Fig. 3E), at 25 mT, the mSCR bent by 86° without a balloon, by 84° with the magnetic balloon, and by 58° with an otherwise identical non-magnetic balloon (n = 3, N = 12). The magnetic balloon therefore preserved 98% of the bending, whereas a non-magnetic balloon, which adds stiffness without adding torque, reduced it by 32%. At 8 mT, the magnetic balloon reduced the tip angle from 66.4° to 61.8°, and the non-magnetic balloon to 25.3° (Table 1). The magnetic balloon also remained compatible with shape forming on dual permanent-magnet and electromagnetic platforms (Fig. S2, S3). Second, the gripper is bonded co-axially to the camera housing, so that it occupies space already used by the camera, and is closed by the 80-µm tendon, which changed the tip angle by 0.1°; the rigid housing and the gripper reduced it by a further 3.2°. Third, the release shell is a 10-µm PLA coating bonded around the body, which occupies no internal cross-section; with the shell, the tip angle at 8 mT was 60.1°. In total, the tendon, gripper, magnetic balloon, and release shell reduced the tip angle at 8 mT from 61.8° to 60.1°, so that the fully functionalized mSCR (configuration 7) retained 92% of the bending left after imaging and sensing, and bent by 64.2° at 8 mT and 78.4° at 16 mT, the range delivered by the EPM at the bedside.

**Task-autonomous cannulation: subroutine and phantom comparison**

All later steps depend on cannulation of the papilla (R2, R6). Following established classifications of autonomy in medical robotics [39], [40], the system operates at two levels (Suppl. 3). Within the ducts, the operator steers the mSCR by teleoperation from the onboard camera view, within software-enforced workspace limits. For papilla cannulation, the system provides task autonomy: once the operator has chosen the target and started the subroutine, it aligns the catheter tip with the papilla and inserts the catheter without continuous operator input, while the operator supervises and can override at any time. We automated this step because it is critical and difficult, and follows the same sequence each time: aligning a catheter with a small, mobile papilla through a side-viewing endoscope is demanding, repeated attempts lead to adverse events such as post-ERCP pancreatitis [3], [5], and competence is acquired over hundreds of procedures [8].

The task-autonomous subroutine combines a YOLOv8 detector, a magnetic steering model, and an FBG shape sensor (Suppl. 3). The detector identifies the papilla, its entry point, and two mSCR tip markers from endoscopic video and estimates the tip orientation (Fig. 4, B and D). It achieved precision = 0.81, recall = 0.76, and F1 = 0.78 on a dataset combining phantom, ex vivo, and in vivo images (N = 3200 images; Suppl. 4). The angular error relative to the desired entry line is mapped to a desired deformation, which a dipole–dipole magnetic model converts into an EPM pose. Once the deformation brings the mSCR tip to point toward the papilla, the advancing unit deploys the mSCR. Because depth remains unknown, the system pauses and realigns before resuming whenever misalignment recurs.

The FBG, the largest single contributor to the estimated bending stiffness of the catheter (43%, Table 1), is retained because it provides an independent measurement of the catheter configuration, which is compared with the orientation predicted by the model. The model–sensor error stayed consistently below 2° (Fig. 4E). In Fig. 4E, the bending angle is the FBG-measured distal orientation relative to the endoscope channel exit in the nonuniform EPM field; it therefore differs from the uniform-field tip angle of Table 1. Errors were largest at small angles, where environmental changes and the nonlinearity of the field made small deviations difficult to correct, and at large angles, where the FBG could not follow sharp bends. A discrepancy exceeding 3° for four consecutive control cycles (~150 ms) triggers a rollback to the last consistent EPM configuration, and loss of any required visual detection halts advancement and returns the EPM to the last configuration with reliable detections, where it is held until tracking recovers (Materials and Methods). Operator interrupts trigger safe holding, from which the system can resume after validation, switch to teleoperation, or move the magnet out of the operating field.

We first compared operator-controlled and task-autonomous insertion in a realistic phantom (Fig. 4, A, B, and H; n = 1, N = 30, 15 per mode). In the phantom, the operator was faster (170 ± 9 s versus 180 ± 5 s) but more variable, consistent with learning effects and case-dependent use of tissue contact: in certain scenarios, the operator leveraged support from the surrounding tissue to assist insertion.

## Task-autonomous papilla cannulation in vivo

A flexible endoscope was advanced to the duodenum to visualize the major and minor papillae and was then secured on a bedside support. Pigs were used because their pancreatobiliary ducts are close in size to those of humans [45]; because the porcine pancreatic duct drains separately through the minor papilla [46], it was cannulated there. In vivo, we compared operator-controlled and task-autonomous cannulation in two animals (Fig. 4, C and D; n = 2, N = 22, 11 per mode). Mean completion time was 223 ± 10 s for task-autonomous cannulation and 253 ± 23 s for operator-controlled cannulation, corresponding to a 30-s, approximately 13%, longer mean time under operator control (Fig. 4H). Navigation strategies relying on tissue contact, which had helped the operator in the phantom, were less transferable in vivo, and operator control exhibited more erratic EPM motion (Fig. 4, F and G, Fig. S4).

The approach also worked with both duodenoscopes and gastroscopes (Fig. 4H, movie S3), and both the operator and the task-autonomous subroutine completed the procedure with each. Because current ERCP catheters require a dedicated side-viewing duodenoscope, which makes advancement toward the papilla more demanding, particularly in surgically altered anatomy [47], [48], compatibility with forward-viewing endoscopes could simplify access to the papilla. No sphincterotomy was performed: in all in vivo procedures, the 1.47-mm catheter entered the native papilla directly, whereas the insertion of cholangioscopes of about 3.5 mm is often facilitated by one [9]. Throughout all trials, the papilla showed no visible bleeding, tearing, perforation, or gross mucosal disruption on endoscopic inspection, although progressive dilation was observed after repeated insertions. These observations were macroscopic and acute; no histopathological, inflammatory, or survival assessment was performed.

## Visualized navigation to the pancreatic duct and the gallbladder

### *Deep pancreatic duct navigation*

Navigating beyond the papilla requires a diameter that fits the narrowing ducts (R1) and a view of the lumen (R4): once the papilla is cannulated, ERCP typically relies solely on external imaging because conventional guidewires lack onboard visualization. By integrating the camera and LED fiber within the mSCR, at the bending cost quantified above, we enabled first-person visual navigation, while the device remained radiopaque for fluoroscopic confirmation when required (Fig. S5). Using joystick-controlled magnetic manipulation, the mSCR entered the pancreatic duct through the minor papilla and reached depths of up to 75 mm (Fig. 5, A and D). The relative reconstruction places the segmented cross-sections along the 7.5-cm path, from the narrow proximal duct to the deepest point (Fig. 5E). To reduce the risk of debris-related visual obstruction, the 100-µm channel, based on the integrated microchannels of [49], was used to flush the duct with water and subsequently aspirate fluids, following the same preventive cleaning principle used in current ERCP practice with the SpyScope.

***Transpapillary access to the gallbladder***

Because the porcine pancreatic and biliary ducts drain separately [46], [50], branch selection could not be demonstrated at a pancreatobiliary junction; the junction where the common hepatic duct (CHD) and the cystic duct join to form the common bile duct provides this demonstration within the target anatomy. In the smaller animal (33 kg), the gallbladder was reached in three of four attempts; it was not specifically attempted in the second animal. The first access used the mSCR with all modules, including the gripper (Fig. S6). Owing to the modular architecture, the two subsequent accesses used a configuration without the gripper, which widens the camera field of view.

Fig. 5C follows one of these accesses from the onboard camera (movie S4), from the major papilla and the common bile duct (steps 1 and 2) to the gallbladder. The catheter first entered the CHD; withdrawing from it (step 3), it reached the junction, where the CHD and the cystic duct were visible in the same frame (step 4). The operator re-entered the CHD for 2.24 cm, returned to the junction, and then steered into the cystic duct (steps 5 to 7, Fig. S7), which led to the gallbladder neck 5.1 cm further (steps 8 and 9). The gallbladder body and fundus were then inspected from the inside, showing bile-stained, granular mucosa (steps 10 and 11). To maintain visibility, fluid was aspirated and air was insufflated alternately through the 100-μm channel. The lumen was visible for most of the final approach: in the last 116 s before the neck, automatic segmentation found the lumen in 97% of the video frames analyzed (one frame every 0.5 s; 233 frames). The gallbladder was visually identified by the operator using the onboard camera. The distance from the papilla to the gallbladder neck was measured at 79 mm and successfully traversed using a 90 mm mSCR, so that only the 1.47-mm diameter section of the device passed through the papilla during gallbladder access.

To show the path taken, we reconstructed the lumen geometry from the onboard videos (Fig. 5B). In each frame, the dark lumen region was segmented, and its cross-section was placed in 3D along the catheter path, which was estimated from the optical flow and scaled to the measured distances (Materials and Methods). The reconstruction shows the common bile duct, the junction where the two lumens appear in the same frame, the short excursion into the CHD, and the cystic duct with its local bends up to the gallbladder neck; the pancreatic duct (Fig. 5E) is drawn at the same scale. It is a relative reconstruction: the absolute cross-section size depends on an assumed viewing distance, the angle between the CHD and the cystic duct was not measured, and the cystic duct was not reconstructed where the lumen was hidden by fluid.

In the clinical cases reported to date, cholangioscopes of 2.3 to 3.5 mm reached the gallbladder over a guidewire already placed in it, in some cases after sphincterotomy [15], [16], [17]. Here, the mSCR selected the cystic duct at its junction with the CHD and was steered through it by magnetic actuation without sphincterotomy or guidewire assistance, providing, to our knowledge, the first magnetically steered transpapillary access to the gallbladder and the first view of its lumen from a sub-2-mm robotic catheter.

## On-demand anchoring provides tip force without stiffening the catheter

Once at the target, the compliance that allowed safe navigation (R3) limits what the catheter can do (R5): forces applied at the tip are transmitted back through the soft body, and the catheter can be displaced by magnetic perturbations or tissue motion during intervention. We instead added stiffness only at the target, with an anchoring balloon that grounds the distal catheter on demand (Fig. 3), so that a gripper or a release shell can act against a stable base. These steps were demonstrated in vivo separately from gallbladder access: anchoring in the minor papilla, grasping in the small intestine near the papilla, and release in the pancreatic duct (Fig. 3A, Fig. 6, E to G, and Fig. 7E).

To compare the two strategies, we measured, in a separate set of catheters clamped at a 60-mm free length, the gripping force with the balloon deflated or inflated, and that of non-anchored catheters whose silicone matrix was two and three times stiffer ($n = 3$, $N = 15$ per condition; Materials and Methods). Anchoring raised the gripping force from 0.43 ± 0.05 N to 1.28 ± 0.20 N, about threefold, as in the gripper characterization (Fig. 6C, $n = 2$, $N = 60$); without anchoring, matrices two and three times stiffer gave 0.80 ± 0.13 N and 1.35 ± 0.12 N. Because the integrated components carry 66% of the bending stiffness, tripling the matrix stiffness raises that of the catheter by 67%: the tip angle of configuration 5 at 8 mT falls from 66° to 52°, and the model predicts that 13.4 mT would be needed to recover it. At 16 mT, the difference is smaller (80° versus 71°) because the catheter approaches alignment with the field. Anchoring therefore provides the tip force of a stiffer catheter only at the target, without stiffening the catheter during navigation.

Inflatable structures are widely used in soft robotics for actuation, sensing, and navigation [51], [52], [53], [54], but a soft tube under pressure and axial load can bifurcate into prismatic, axisymmetric (bulging), or asymmetric (bending) modes [55], which would compromise anchoring. We therefore designed the anchoring balloon within a buckling-aware framework (Materials and Methods). Considering the large strains observed (>300%), we employed a Yeoh strain-energy density function to link internal pressure, external tissue pressure, axial loads from gripper actuation, and radial deformation (Fig. 3B and Suppl. 5). The model showed its reliability particularly in the higher-pressure range (RMSE = 0.4 mm below 27 kPa and 0.3 mm above; $n = 3$, $N = 30$, Fig. 3B). The burst pressure (64 ± 11 kPa) was repeatable and defined the safe actuation range of the inflation unit ($n = 3$, $N = 30$, Fig. 3C). With the balloon pressed against the force sensor, the inflation pressure reached 57 kPa at 2.6 mm of radial deformation, below the burst pressure ($n = 3$, $N = 30$, Fig. 3D), and the holding force reached $F_{anchor} = 3.3 \pm 0.2$ N ($n = 3$, $N = 30$, Fig. S8), sufficient to stabilize the catheter within the duct during gripper actuation. In vivo, the balloon was inflated and anchored the catheter in the minor papilla, where inflation could be followed from the endoscope (Fig. 3A, movie S5). Distal anchoring also allows the tip to be steered independently of the rest of the body. With the balloon inflated in a 4-mm synthetic vessel and the tip driven along four prescribed trajectories by an electromagnetic navigation system, the tip-tracking RMSE fell from 0.60–1.16 mm without anchoring to 0.37–0.45 mm with anchoring, and the fraction of tip positions within 0.5 mm of the path rose from 31–55% to 75–82% (Fig. 3F, Suppl. 6).

## Grasping with a low-tension compliant gripper

With the catheter grounded, a compliant gripper can apply forces at the target (R5; Fig. 6). The gripper is a single 3D-printed part without pins or joints, which would add bulk at this scale, and is inspired by magnetically actuated compliant milli-grippers [56] and by the flexure-based L-arm mechanism [57]; here, a central shuttle translates relative to a fixed base to close the jaws. In our embodiment, the base is bonded to the camera housing for co-axial alignment, and actuation is provided by pulling the 80-µm nylon tendon, which runs in a molded guide close to the neutral axis of the catheter. We extended the gripper model from localized [57] to distributed compliance and optimized the flexure geometry to minimize the tendon tension needed for closure. With the optimized flexures, full closure required 0.48 ± 0.01 N of tendon tension and displaced the catheter tip by 0.22 ± 0.03 mm with the balloon deflated and by 0.07 ± 0.02 mm with the balloon inflated ($n = 3$, $N = 15$), whereas a non-optimized flexure, which required three times this tension (1.43 ± 0.04 N), displaced it by 0.60 ± 0.03 mm and 0.22 ± 0.02 mm, respectively. To avoid interference with ultrasound-triggered release, the jaw blades incorporate a twisted design that shifts their nearest natural frequencies to 0.62 and 1.48 MHz, away from the 1-MHz trigger (Fig. 6, A and B); a gripper exposed to the trigger for 10 min showed no jaw motion or change in closing force. Model predictions of jaw deflection and triggering force agreed with measurements within 7% (n = 2, N = 60; Fig. 6C).

Peak gripping force without anchoring was 0.43 ± 0.06 N (n = 2, N = 60); when combined with the anchoring balloon, the effective gripping force increased to 1.28 ± 0.16 N (n = 2, N = 60), a roughly threefold increase (Fig. 6C). Because the gripper was characterized once fully integrated in the mSCR, the reported tip force includes transmission losses along the catheter. Without anchoring, the tip force is about one third of the anchored force; with anchoring, it approaches that of the gripper on a rigid base (1.42 ± 0.10 N). Across bench experiments, we demonstrated gripping of materials with different mechanical properties and at different moving speeds (Fig. 6D, S9). Repeated-gripping tests showed that all samples remained operational over 250 cycles (Suppl. 7). The local contact pressure and its consequences on tissue were not characterized and will be an important next step for translational validation.

In vivo, grasping was performed in the small intestine near the papilla, where it could also be followed from the endoscope; the onboard camera provided a direct view of the gripper–tissue interface, and the gripper grasped and removed an object and manipulated soft tissue (Fig. 6, E to G, movie S6).

**Wireless payload release compatible with ultrasound imaging**

The second action at the target (R5) is local delivery, which could concentrate a therapeutic agent where it is needed. Release needs no onboard actuator: it is triggered wirelessly, as steering is, and the payload is held by a micropatterned coating that reduced the tip angle at 8 mT by only 1.8° (Table 1). Unfocused ultrasound (US) serves both for real-time imaging (typically 7–28 MHz) and for mechanically triggered release (Fig. 7). Rather than the electrospun fabrics used previously for ultrasound-triggered release [58], the shell is a micrometer-thick PLA coating around the mSCR with a designed micropattern (squares, 20 µm edge length, 5 µm depth, 10 µm spacing; Fig. 7B) whose tiles rock on the interconnecting film and the compliant adhesive and silicone layers at about 1 MHz, whereas the first mode of a free tile lies near 14 MHz. We chose 1 MHz to avoid overlap with imaging US and to leverage deeper tissue penetration. The pattern on the molded shell had an area of 402 ± 24 µm², validating the molding approach (400 µm² expected; Fig. S10), and finite-element modal analysis of the tiles on their supporting layers confirmed the 1-MHz natural frequency (Fig. S11). The other modules are either insensitive to 1-MHz ultrasound or, as for the gripper blades, designed to remain so.

In vitro, we demonstrated time-controlled release: samples were continuously stimulated at 7.2 MHz to simulate imaging, and at pre-specified times (20, 25, 30 min), a 1-MHz stimulus was applied and release was visually confirmed (Fig. 7A, movie S7). Release experiments showed a payload release of 80% within t = 104 ± 22 s after 1-MHz insonation (n = 4, N = 12), with no measured leakage during baseline imaging. Neither an unpatterned shell insonated at 1 MHz nor the patterned shell insonated at 2.5 MHz at matched MI released more than 10% of the payload in 10 min. Phantom and ex vivo studies demonstrated optical confirmation of release using narrow-band endoscopic imaging at 415 nm (Fig. 7, C and D). In the absence of surrounding fluid, direct tissue contact is required to couple ultrasound energy into the shell. The thermal and mechanical indices (Materials and Methods) remained within the diagnostic safety framework, and thermal characterization in a tissue-mimicking medium showed only a transient local temperature increase of ~0.8°C (Suppl. 8). The model payload was selected for visualization, and compatibility with a clinically relevant therapeutic payload was verified separately (Suppl. 9).

In vivo, ultrasound was used intermittently during navigation to visualize the surrounding anatomy, and the acoustic artifact of the catheter gave the operator a confirmatory cue of its position (Fig. 7F, Fig. S12); the image did not resolve a sharp outline of the mSCR and was not used by the magnetic controller. The release site was selected on the ultrasound image, and the catheter was then navigated to it using the onboard camera; once the catheter was in position, the anchoring balloon was inflated, ultrasound imaging was paused, and a separate 1-MHz transducer was activated to trigger payload release while the catheter remained mechanically stabilized. The released payload was visible at the papilla as the catheter was retracted (Fig. 7E).

## DISCUSSION

This work shows that a magnetically steered soft continuum catheter with a 1.47-mm outer diameter can combine onboard visualization, steerability, distal stabilization, and action at the target within the size limits of the pancreatobiliary ducts. Because the catheter enters the native papilla, none of the procedures required a sphincterotomy, an incision associated with bleeding and perforation [9], and cannulation was completed with forward-viewing gastroscopes as well as with duodenoscopes. The contribution lies less in the diameter itself than in how the functions were fitted within it and in measuring what each one costs. At this scale, imaging and shape sensing occupy 12% of the cross-section but 66% of the bending stiffness, and they reduced the tip angle at 8 mT from 86° to 70°. The remaining modules were therefore made magnetic, placed on existing footprints, or reduced to a surface layer, and together cost 8% of the remaining bending. The stiffness needed at the target was not built into the body but supplied on demand by the anchoring balloon, which increased the gripping force about threefold (Fig. 6C) without stiffening the catheter during navigation, whereas a body stiff enough to provide the same force would need 1.7 times the field to bend as far. Because the balloons are molded and magnetized with the body, several interconnected balloons can be placed along the catheter (Materials and Methods), which could provide anchoring at multiple points in future procedure-specific designs. The same competition between function, compliance, and magnetic response applies to other soft robots that must carry function through narrow anatomy, and measuring the bending cost configuration by configuration, as in Table 1, gives a direct way to compare design options before in vivo testing.

Access to the gallbladder illustrates what this combination could enable. Because its lumen cannot currently be inspected or sampled endoscopically, uncertain lesions are managed by surveillance or cholecystectomy, which removes many benign gallbladders while some cancers are found only after surgery [10], [11], [12]. A catheter that reaches the lumen through the papilla under direct vision, and can then be anchored, could support visual assessment of polypoid lesions and, with a sampling tool, tissue diagnosis before a decision on surgery; it could also deliver local therapy in patients unfit for surgery. Drainage is a nearer application: during access, the catheter already aspirated fluid and insufflated the lumen through its own channel to restore visibility, and a catheter that crosses the cystic duct under direct vision could guide the guidewire that drainage currently passes blindly. These applications remain prospective: no lesion was inspected or sampled here, porcine and human pancreatobiliary anatomy differ [45], and gallbladder access was attempted in one animal only.

The workflow combines joystick teleoperation with a task-autonomous subroutine for papilla alignment and insertion. Endoscopic video supplies the target and catheter-marker detections used by the magnetic controller, whereas FBG reconstruction independently monitors agreement between the predicted and measured catheter configurations. We deliberately restricted autonomy to cannulation, a well-defined step whose difficulty is linked to adverse events [3], [5], and the in vivo comparison should be read as evidence that this subroutine can perform the step reliably rather than as a demonstration of autonomous navigation. During cannulation, the endoscope was held on a bedside support, which simplified the visual task compared with a hand-held duodenoscope. The subroutine lacks depth estimation; localizing the tip would reduce readjustments and allow tracking inside the lumen. Tip localization could come from miniaturized trackers [59], [60], at the cost of extra cabling, or from the onboard camera, as in camera-based servoing of magnetic endoscopes [61], which in fluid-filled ducts would need a lens-cleaning lumen.

Several module-level limitations remain. Despite geometric optimization, transmitting force from the base to the tip still introduces some loss of precision, and distal actuation, for example with biohybrid actuators at the catheter tip [62], [63], could in the longer term improve accuracy without compromising compliance. The separation of the gripper blade modes from the 1-MHz trigger was established by finite-element modal analysis and verified on the bench. Once triggered, the shell delivers its entire payload, which is appropriate when a single delivery is expected; multiple deliveries would require variable patterns, other matrices with different natural frequencies, or other triggering mechanisms [64], [65], [66]. Using a single EPM to control a soft catheter also involves greater nonlinearity than coil- or dual permanent-magnet platforms [31], [36], which offer finer control but face space and access limitations in the operating theatre. The tendon–magnetic coactuated concentric robot steered by a coil tracked 3D paths with 0.87–1.10 mm error within an 80 × 80 × 30 mm workspace [38]; the bedside EPM trades some linearity for a 0.05-$m^3$ magnetic-feasible workspace around the patient. Anchoring partly offsets this nonlinearity at the target: with the balloon inflated, planar tip-tracking errors fell to 0.37–0.45 mm (Fig. 3F), although the two setups are not directly comparable.

All in vivo procedures were performed by one operator, a consultant pancreaticobiliary and general surgeon, who was familiarized with the platform on a phantom immediately before the experiments. Because only one operator was involved, the study does not establish training requirements or a learning curve, and multi-operator studies are needed to assess usability and performance variability. The reported timing and trajectory differences are descriptive observations within the tested conditions, not estimates of population-level performance or clinical efficacy, and the study was not powered to assess safety or superiority over conventional cannulation. Future work should include prespecified multi-animal and multi-operator comparisons against conventional guidewires or catheters along matched anatomical paths.

The present study provides acute functional-feasibility data and short-term biological compatibility rather than a tissue-level safety assessment. The papilla showed no visible bleeding, tearing, perforation, or gross mucosal disruption, although progressive dilation was observed after repeated insertions, and no histopathological, inflammatory-response, or survival assessment was performed. This is particularly relevant in ERCP, where current catheterization approaches can lead to complications such as acute pancreatitis and cholangitis [5], [7], potentially delaying treatment or upfront surgery. The small diameter and soft construction of the mSCR may help reduce procedure-related tissue trauma, but this must be established in dedicated studies with predefined controls, tissue sampling, and survival follow-up before first-in-human use.

By measuring the bending cost of each function, adding function through the magnetic body, existing footprints, and surface layers, supplying stiffness only at the target, and scoping autonomy to a single clinically defined task, the platform demonstrates a route toward procedure-specific multifunctional soft robotic catheters for regions that current endoscopic tools cannot readily reach.

## MATERIALS AND METHODS

### Robotic platform

A 7-DoF kinematically redundant serial manipulator (LBR Med 14 R820, KUKA) carried a cylindrical N52 external permanent magnet (EPM; K&J Magnetics, USA) with a remanence of approximately 1.45 T, 101.6 mm in diameter and 101.6 mm long (dipole moment about 950 A·$m^2$). By adjusting the pose of the EPM relative to the patient, magnetic forces and torques were applied to the mSCR to achieve the desired shape changes (section "Task-autonomous cannulation: subroutine and phantom comparison" and Fig. S2). Five different endoscopes were used successively during the trials, a duodenoscope, a gastroscope, a bronchoscope, and two different colonoscopes (Table S3). Endoscope video and the onboard camera feed were captured via a framegrabber and streamed to the control computer for use in the control loop.
A joystick (Nunchuk Controller, Nintendo, Japan) was used to issue EPM pose commands and to control catheter deployment and retraction as well as functional module actuation. A feed unit advanced the catheter, and a dedicated inflation unit controlled the balloon pressure. The setup during the in vivo trials is presented in Fig. S1.
The platform is based on ROS, with networked nodes for: camera synchronization, feature detection and orientation estimation, manipulator control, motor drivers, FBG acquisition and shape reconstruction, user interface, and data storage.
To mitigate magnetic-interaction risks, the procedure is performed only after confirming that no ferromagnetic instruments or objects are present within 2 m of the robotic actuation area, with nearby tools restricted to non-ferromagnetic materials such as titanium-based alloys or polymers. Active implanted electronic devices are treated as contraindications, and passive metallic implants are reviewed on a case-by-case basis before use.
The system builds on our previous magnetic robotic platform [67], parts of which have already been translated to human studies [68], as well as on magnetic surgical systems already used in the operating room (Levita Magnetics) [69]. To support safe operation, the system includes configurable workspace limits, torque-based stop conditions during autonomous insertion, tracking-consistency monitoring based on the agreement between model prediction and sensed catheter shape, and a user-accessible emergency stop. Full details of the control loop, software architecture, and safety implementation are provided in Suppl. 3.

**mSCR actuation unit**

The advancing unit comprised a five-phase hybrid stepper motor (MS17HV, MOONS') driven by a digital driver (5SRZ02H-S, MOONS') under an Arduino UNO R4 Minima. The motor was mounted on a lockable articulating base stage (SL20/M, Thorlabs) using a 3D printed adapter positioned at the endoscope inlet. The inflation unit used a lead screw linear module (drylin SAW 1080, igus) to actuate a 10 mL syringe through the same driver and microcontroller. A three-way stopcock connected the syringe to a pressure sensor (VSP1130, Amphenol) and the silicone inflation line to enable closed loop control. All electronics were enclosed in a safety housing with switching supplies (MPS KO-D120524V, Dynamic). Arduinos were operated as ROS nodes via rosserial over UART protocol: subscribers received joystick commands; publishers streamed pressure for closed-loop inflation to prevent over-pressurization and bursting, as well as the mSCR length deployed.

**mSCR manufacturing process**

Soft robotic catheter bodies were produced by low-pressure injection molding using high-resolution molds (microArch S140, Boston Micro Fabrication; HTL resin). Two-part shells incorporated integrated cores to define the 100-µm channel and balloon lumen, together with guides for the camera, fiber, and tendon. The magneto-elastic matrix consisted of a 3.7:1 mass mixture (34.5% NdFeB by volume) of neodymium-iron-boron powder with 5 µm particles (MQFP-B+, Magnequench) and DragonSkin 30 silicone (Smooth-On, USA) that was mixed and degassed in a planetary mixer (Thinky ARE250, Intertronics, UK) at 2000 rpm for 1.5 minutes, loaded into a syringe, and injected while vacuum venting removed trapped air. After curing, parts were released in an acetone bath; targeted acetone injection cleared sacrificial cores and confirmed lumen patency. The monocular NanEyeM camera module (320 × 320 pixels, ams OSRAM, USA), LED fiber (Thorlabs, USA), and multicore FBG (FBGs International, Belgium) were verified prior to integration. The compliant gripper was bonded coaxially to the camera housing and coupled to a nylon tendon. The magnetized composite had a remanence of 0.31 T and an elastic modulus of 3.04

MPa (from the Yeoh constant C1). Finished catheters measured 1.45 ± 0.02 mm outer diameter and 90 mm length; the 1-mm marker bands were recessed in molded grooves, so that the outer diameter stayed below 1.5 mm, and the camera cable was integrated without its PTFE jacket. The 100-µm channel was connected to a soft rubber tubing (0.3 mm ID and 0.64 mm OD, McMaster-Carr, IL, USA), then all the cables, fibers and tube were routed inside a medical grade PTFE tube (1.2 mm ID and 1.6 mm OD, McMaster-Carr, IL, USA). All the components can be quickly connected to their respective actuation unit before use.

Depending on the task, the mSCR was assembled with different module combinations from the same molded body. For the gallbladder trials, the first access used the configuration with all modules, and the two subsequent accesses used a configuration in which the gripper was omitted to widen the camera field of view, with the same outer diameter.

**mSCR magnetic manipulation**

To successfully manipulate the magnetic soft robot, an understanding of the coupling between finite deformation and magnetic fields is required to model hard-magnetic soft materials. We consider the magnetoactive soft material as a homogenized continuum body with a constitutive law given by a Helmholtz free energy as a function of the Cauchy–Green tensor and a Lagrangian magnetic field variable as independent variables [70]. To facilitate physical interpretation and separation of the magneto-mechanical constitutive laws, as demonstrated in [71], [72], we divide the nominal Helmholtz free-energy density into an elastic part $\widetilde{W}_{\mathrm{el}}(\boldsymbol{F})$ as a function of the deformation gradient $\boldsymbol{F} \in \mathbb{R}^{3\times3}$, relating to both the reference and current configurations of the mSCR, and a magnetic part $\widetilde{W}_{\mathrm{mag}}(\boldsymbol{F}, \widetilde{\boldsymbol{B}})$ as a function of the deformation gradient $\boldsymbol{F}$ and the nominal magnetic flux density $\widetilde{\boldsymbol{B}} \in \mathbb{R}^3$. This separation ensures that the elastic part can be linked to commonly used constitutive models for soft materials such as Neo-Hookean, Mooney–Rivlin, and Ogden. In bending, the surface strain of the mSCR stays below 10% up to 16 mT (up to 12% for the magnetic body alone at 25 mT), and the integrated cables limit axial stretch, so that a Neo-Hookean form is sufficient; the Yeoh model is used only for the balloon, whose strains exceed 300%. Of the principal invariants $(I_1, I_2)$, the Neo-Hookean form retains only the first. The strain-energy density is then given by

$$W_{el} = C_1\left(I_1 J^{-2/3} - 3\right) + D_1(J-1)^2$$

Where $C_1$ and $D_1$ are material constants which, by consistency with linear elasticity, can be expressed respectively as half the shear and bulk moduli, and J is the determinant of $\boldsymbol{F}$.

We then consider a locally uniform magnetic field $\boldsymbol{B}_{EPM} \in \mathbb{R}^3$ generated by the EPM, or electromagnetic coils (Fig. S2, S3), and a mSCR of length L and radius r, with a Young's modulus E and a uniform remanence $\tilde{B}_r$, so that the orthogonal magnetic torque at position x on the mSCR can be expressed as

$$\tau(x) = \mu_0^{-1}\left|\widetilde{\boldsymbol{B}}_r\right|\left|\boldsymbol{B}_{EPM}\right|\pi r^2(L-x)$$

Where x denotes the curvilinear coordinate along the mSCR measured from its proximal end, and the maximal deflection $\delta_{max}$ can then be derived as

$$\delta_{max} = \frac{4\left|\widetilde{\boldsymbol{B}}_r\right|\left|\boldsymbol{B}_{EPM}\right|L^3}{3E\mu_0 r^2}$$

Where $\mu_0$ is the magnetic permeability of free space ($4\pi \times 10^{-7}$ N·A$^{-2}$). While the model covers an axially magnetized mSCR, it can be extended to piecewise magnetization profiles by discretizing the mSCR into multiple segments. The integrated components are non-ferromagnetic and are therefore treated as mechanical, rather than magnetic, perturbations through the effective composite properties of the mSCR. Moreover, the applied field (≤16 mT in vivo and ≤25 mT on the bench) is only a few percent of the intrinsic coercivity of the NdFeB particles (of order 1 T), which sets the field required to alter the remanence ($\tilde{B}_r$). Any demagnetization caused by this field is therefore negligible. The values of the different constants are summarized in Table S4.

The robotic end-effector was controlled in Cartesian space by means of an incremental algorithm. The magnetic interaction between the EPM and the mSCR is modeled using a magnetic dipole expansion and builds on previous work, with $B_{EPM}$ related to the pose as follows:

$$\boldsymbol{B}_{EPM} = \frac{\mu_0|\mathbf{m}|}{4\pi|\mathbf{p}|^3}(3\widehat{\boldsymbol{p}}\widehat{\boldsymbol{p}}^T - I_3)\widehat{\boldsymbol{m}}$$

where $\boldsymbol{p} \in \mathbb{R}^3$ represents the position of the EPM, $\boldsymbol{m} \in \mathbb{R}^3$ is the magnetic dipole moment, and $|.|$ denotes the Euclidean norm, with $\hat{\cdot} = \frac{\cdot}{|\cdot|}$. The pose of the EPM, for the desired field, derived from the optimization, is then given by:

$$|\mathbf{p}| = \left(\frac{\mu_0|\mathbf{m}| \parallel (3\widehat{\mathbf{p}}\widehat{\mathbf{p}}^T - I_3)\widehat{\mathbf{m}} \parallel}{4\pi|\mathbf{B}_{EPM}|}\right)^{1/3}$$

$$\widehat{\mathbf{m}} = \frac{(3\widehat{\mathbf{p}}\widehat{\mathbf{p}}^T - I_3)^{-1}\widehat{\mathbf{B}}_{EPM}}{\parallel (3\widehat{\mathbf{p}}\widehat{\mathbf{p}}^T - I_3)^{-1}\widehat{\mathbf{B}}_{EPM} \parallel}, \quad (3\widehat{\mathbf{p}}\widehat{\mathbf{p}}^T - I_3)^{-1} = 3/2\, \widehat{\mathbf{p}}\widehat{\mathbf{p}}^T - I_3$$

At every time-step, the position variation ($\delta$p) and heading increment ($\delta\widehat{\mathrm{m}}$) with respect to the previous step are computed and sent to the low-level controller.

**Bending cost of functionalization and anchoring versus stiffening**

To measure the bending cost of each component, mSCRs were assembled in cumulative configurations (Table 1), starting from the magnetic body with its 100-µm channel and balloon lumen and adding, in turn, the camera cable and LED fiber, the FBG, the tendon, the rigid camera housing with the gripper, the magnetic balloon or, as an alternative, an otherwise identical non-magnetic balloon, and the release shell. Each catheter was clamped at its base and exposed to uniform fields of 8, 10, 12, 16, 20, and 25 mT generated by the single-axis coil, perpendicular to its axis, and the tip angle was measured from images of the catheter (three catheters per configuration clamped at a 60-mm free length, five repeats per field). The share of each component in the composite bending stiffness was computed from the composite modulus (3.04 MPa) and from component data (Suppl. 2), and the measured tip angles were compared with the magneto-elastic model using the measured remanence (Suppl. 2). All catheters, including those used in vivo, had a 90-mm soft segment. For the comparison between anchoring and stiffening, the gripping force was measured as described in the section "Compliant gripper", and the tip angle at 8 and 16 mT was measured on non-anchored catheters in which the DragonSkin 30 matrix was replaced by Smooth-Sil 945 (about 2.3 MPa) or Smooth-Sil 960 (about 3.4 MPa) (Smooth-On, USA), two and three times stiffer than the DragonSkin 30 matrix (about 1.1 MPa), respectively (three catheters per condition, five repeats).

**Buckling-aware design of a magneto-sensitive hyperelastic balloon**

When deflated, the balloon should maintain continuous magnetization within the catheter, remaining compatible with previously reported shape-forming techniques and complex magnetization [44]. To ensure this property, a thick-walled cylinder is required (diameter-to-thickness ratio < 10); the balloon section has a wall thickness of 160 µm. This allows the compressive hoop stress to be accurately estimated [73] and reduces the risk of kinking, a localized form of buckling, while enabling shape-forming for navigation.

Once inflated, the relationship between pressure, force, and deformation can be expressed using the Yeoh strain-energy density function,

$$W = \sum_{i=1}^{N} C_i(\lambda_r^2 + \lambda_\theta^2 + \lambda_z^2 - 3)^i$$

where N is the polynomial order and $C_i$ are material constants describing the nonlinear elastic response of the elastomer, with $C_1$ related to the small-strain shear response, $C_2$ to the intermediate-strain behavior, and $C_3$ to strain hardening at larger deformations. The Yeoh model was preferred over the Ogden model in this scenario due to the lack of a straightforward analytic model arising from irrational numbers in the material constants [74], leading to non-integrability of the force–pressure–deformation expression. For a third-order Yeoh model (N = 3), the analytical solutions for pressure and axial force were computed using Maple as functions of $C_1$, $C_2$, and $C_3$, under the assumption of no axial stretch ($\lambda z = 1$), as:

$$\Delta p = \tilde{p}(\lambda_{in}, 1) - \tilde{p}(\lambda_{ex}, 1) \text{ and } f_{ex} = \tilde{f}(\lambda_{in}, 1) - \tilde{f}(\lambda_{ex}, 1)$$

For clarity, the full expressions are provided in Suppl. 5.

A recurrent issue noted in the literature for soft balloons is buckling during inflation, often linked to the absence of a suitable model or to the use of strain-energy functions that exclude certain bifurcation features (Neo-Hookean, Mooney–Rivlin). In our case, the anchoring section is subjected to internal and external pressure, while potentially experiencing axial loading during gripper actuation; these phenomena had to be modeled to derive geometric parameters that avoid instability. Following Haughton and Ogden [55], three types of bifurcation were analyzed: prismatic bifurcation, in which the cross-section becomes non-circular uniformly along the length; axisymmetric bifurcation (bulging), in which the cross-section remains circular but the radius varies along the length; and asymmetric bifurcation, in which the balloon axis bends. For a balloon held at fixed axial stretch, these correspond respectively to the exact criterion for prismatic bifurcation (mode n = 1) [55], the onset of bulging in an infinitely long tube, which gives the lowest stretch at which bulging can begin for any balloon length [75], and the necessary condition for bending [55]:

$$\begin{cases} \displaystyle\int_{\lambda_{r_{out}}}^{\lambda_{r_{in}}} \frac{\frac{\partial^2 \widehat{W}}{\partial \lambda^2} \lambda d\lambda}{(\lambda^2 \lambda_z - 1)\left(\frac{\partial \widehat{W}}{\partial \lambda}\right)^2} = 0 \\ J = \dfrac{\partial P}{\partial \lambda_{r_{in}}} \dfrac{\partial F}{\partial \lambda_z} - \dfrac{\partial P}{\partial \lambda_z} \dfrac{\partial F}{\partial \lambda_{r_{in}}} = 0 \\ D = \displaystyle\int_{\lambda_{r_{out}}}^{\lambda_{r_{in}}} \frac{\left(2\lambda_z \frac{\partial \widehat{W}}{\partial \lambda_z} - \lambda \frac{\partial \widehat{W}}{\partial \lambda}\right) \lambda d\lambda}{(\lambda^2 \lambda_z - 1)^2} \leq 0 \end{cases}$$

where $\lambda$ is the azimuthal stretch r/R, $\lambda_{r,in}$ and $\lambda_{r,out}$ are its values at the inner and outer walls, $\lambda_z$ is the axial extension ratio, which in our case is equal to 1 because the cables prevent axial deformation, and $\hat{W}(\lambda, \lambda_z)$ is the strain-energy density function. P is the internal pressure, and $F = N - \pi r_{in}^2 P$ is the axial force excluding the pressure acting on the closed ends, where N is the resultant axial force [55]:

$$P = \int_{\lambda_{r_{out}}}^{\lambda_{r_{in}}} \frac{\frac{\partial \widehat{W}}{\partial \lambda}}{(\lambda^2 \lambda_z - 1)} d\lambda, \quad \frac{N}{\pi R_{in}^2} = \left(\lambda_{r_{in}}^2 \lambda_z - 1\right) D + P \lambda_{r_{in}}^2$$

The inner and outer radii are linked by incompressibility:

$$r_{out}^2 = \frac{R_{out}^2 - R_{in}^2}{\lambda_z} + r_{in}^2$$

Under internal pressure, $\lambda^2 \lambda_z > 1$ across the wall, so the first criterion cannot be met while $\partial^2 \hat{W}/\partial \lambda^2 > 0$. For the Yeoh form with $C_1 > 0$ and $C_2, C_3 \geq 0$ (Table S4), this inequality holds for all $\lambda$, so prismatic bifurcation cannot occur. The bulging condition J = 0 was evaluated from the closed-form expressions for P and N (Suppl. 5); bulging cannot begin below $\lambda_{r,in} = 3.2$ (outer radial deformation of 1.2 mm), whatever the balloon length [75]. Bending of the balloon axis requires $D \leq 0$, which is met for an inflated tube held at $\lambda_z = 1$; it is resisted by the catheter core and the cables that run through the balloon, which we assume prevent it.

When interconnecting two or more balloons, additional instability must be considered. For two interconnected balloons [76] (Fig. S13), with radii $R_a$ and $R_b$, and defining $\lambda_i = R_i/R_0$, the snap-through instability can drive fluid flow from the smaller-volume balloon to the larger one. For an incompressible spherical membrane of initial radius $R_0$ and initial wall thickness $h_0$, inflated to a stretch $\lambda$ and thickness $h_0/\lambda^2$, the Mooney–Rivlin pressure–stretch relation is

$$p = \frac{4h_0 C_1}{R_0}\left(\frac{1}{\lambda} - \frac{1}{\lambda^7}\right)\left(1 + \frac{C_2}{C_1}\lambda^2\right)$$

Under the same internal pressure, the equilibrium can be written explicitly as:

$$\left(\frac{1}{\lambda_a} - \frac{1}{\lambda_a^7}\right)\left(1 + \frac{1}{K}\lambda_a^2\right) = \left(\frac{1}{\lambda_b} - \frac{1}{\lambda_b^7}\right)\left(1 + \frac{1}{K}\lambda_b^2\right)$$

where $C_1$ and $C_2$ are the Mooney–Rivlin constants (distinct from the Yeoh constants above) and $K = C_1/C_2$. The bifurcation branches off from the symmetric solution at the point $\lambda = \lambda_{max}$ which corresponds to the maximum of the pressure–radius relation. Beyond this point, either both balloons proceed symmetrically downwards in pressure, or one balloon grows while the other falls back in pressure and radius. As pressure increases with the growing balloon radius up to a critical value $p_{cr}$, snap-through occurs and the balloon suddenly expands to a larger radius due to the highly non-linear elasticity of the rubber.

## Tip-trajectory tracking with and without anchoring

The mSCR tip was driven along four prescribed trajectories (circle, square, spiral, and butterfly) by an electromagnetic navigation system (MFG-100, MagnebotiX, Switzerland), with the anchoring balloon either inflated or deflated inside a synthetic vessel with a 4-mm inner diameter (SynTissue, SynDaver, USA). The tip, illuminated by the integrated LED fiber at 450 nm, was tracked with a camera (Basler, Germany). The planned path was registered to the workspace from the commanded trajectory of the navigation system, without fitting to either run, and the same registration was used for both runs. The RMSE was computed from the distance between each tracked tip position and the nearest point of the planned path, and path adherence as the percentage of tip positions within 0.5 mm of it (Suppl. 6).

## Compliant gripper

To model elastic deformation of the flexible element, we approximate it as a beam in which bending dominates over shear [77], [78]. Under the assumption of a pure end-load $f = [f_x \ f_y \ M]^T$ generating a variable tangent angle $\phi(S)$ with the distance along the neutral axis $S \in [0, L]$, where L is the beam length, the curvature of the beam $\phi'$ can be integrated using a smooth curvature model. Previous research has shown that shifted Legendre polynomials are a good choice, allowing low-order but high-accuracy prediction of large deformation [78]:

$$\phi'(S, \alpha) = \frac{\alpha_0}{L} + \frac{\alpha_1}{L}\left(\frac{2S}{L} - 1\right) + \frac{\alpha_2}{L}\left(\frac{6S^2}{L^2} - \frac{6S}{L} + 1\right)$$

The coefficients $\alpha = [\alpha_0 \ \alpha_1 \ \alpha_2]^T$ serve as generalized coordinates that describe the flexure configuration. This generic formulation is applied to determine the deformation of the gripper under the pulling force required to close it, as well as to calculate the mSCR tip coordinates under the loading generated by gripper actuation. Material and geometric differences between the gripper and the mSCR arise when deriving the energy function under the same smooth-curvature basis. We then established a simple system of equations linking the gripper actuation force to the resulting mSCR deformation. As the force required to actuate the gripper depends on geometric parameters, we ran an optimization over the gripper geometry to achieve the smallest acceptable mSCR deformation.
The final part of the gripper parametrization concerned the mechanical ultrasound drug delivery. To avoid matching the natural frequencies of the gripper beams with that of the drug coating (and to ensure structural integrity), we pre-twisted the beams. Pre-twisting increases stiffness and raises the natural frequency [79], [80], whose dimensionless expression can be calculated as:

$$\overline{w} = w\left(\frac{\rho A L^4}{\sqrt{EI_{xx}EI_{yy}}}\right)^{\frac{1}{2}}$$

Once designed, the grippers were then manufactured using an ultra-high resolution 3D printer (microArch® S140, Boston Micro Fabrication, MA, USA) using the HTL resin.
We validated that the actuation-force minimization did not alter the soft-body shape by driving the catheter with a single-axis coil and comparing the tip position with the gripper open and closed, with the balloon deflated and inflated, while recording the tendon tension (deviation at full closure, 0.22 ± 0.03 mm deflated and 0.07 ± 0.02 mm inflated; n = 3, N = 15). We also validated the actuation-force model by connecting the actuation wire to a load cell (F/T Sensor Nano17 Titanium, ATI, MA, USA) and a linear translation stage (NRT150, Thorlabs, NJ, USA) (Fig. S9). By inverting this set-up, we characterized the gripping force. Tests were performed with and without balloon stabilization, demonstrating increased gripping force when using the balloon (Fig. 6C).

**Ultrasound Triggered Drug Release Shell design and fabrication**
As the shell is a vibrationally driven system, that requires both deep-tissue penetration and spectral separation from imaging ultrasound, it was designed to resonate at 1 MHz. Starting from a simple square pattern, we first performed mechanical-impedance analysis (Z = F/v, with v the vibration velocity) of the patterned shell on its adhesive and silicone support to select a set of geometric parameters (edge length, depth). We then ran finite-element modal analysis (ANSYS 2023, Ansys Inc., USA) to extract the resonant modes of the patterned tiles. The model included the interconnecting PLA film and the underlying adhesive and silicone layers. This confirmed that a micro-pattern comprising interconnected PLA square tiles with 20 μm edge length and 5 μm depth, separated by 10 μm spacing, resonates near 1 MHz in a rocking mode of the tiles on this compliant support (Fig. S11). A negative mold with the optimized geometry was printed on an ultra-high-resolution 3D printer (microArch® S140).

The model payload used in the results demonstrations was selected primarily for visualization, whereas separate experiments with clinical drug were performed to verify compatibility of the platform with a clinically relevant therapeutic payload (Suppl. 9). Povidone-iodine (PVP-I; $(C_6H_9NO)_n \cdot (I_2)_m$) was used as a model payload for its enzyme reactivity in the pancreatobiliary tract and tincture visibility on tissue. A gelatin (150 Bloom, Scientific Laboratory Supplies, UK) / PVP-I (Videne, Ecolab, MN, USA) / deionized cold-water pre-mix (mass ratio 1:0.5:5) was prepared by blooming for 30 s, then mixing at 60 °C and 800 rpm on a hot-plate magnetic stirrer (Heidolph MR-Hei Standard, Germany). The mixture was quenched at −18 °C to prevent sedimentation, milled, incorporated into a PLA matrix, and spin-coated (WS-650-23, Laurell) onto the printed mold to form a thin, drug-loaded shell. The film's thinness allowed conformal shaping over the mSCR, and it was bonded using a water-activated polyurethane adhesive (Gorilla Glue, USA).

Bench validation was performed in a water bath held at physiological temperature. To emulate acoustic attenuation/scattering and ensure consistent coupling, samples were isolated in 3D-printed tubes (Elastic 50A Resin V2, Formlabs, USA), a resin with a stable ultrasound response and a sound speed close to that of soft tissue [81]. Continuous imaging was simulated with a Philips L9-3 probe at 7.2 MHz on an iU22 system. At 20, 25, and 30 min, sonomechanical activation was applied using a 1 MHz collimated transducer (5 $cm^2$) driven by an SP300 sonoporation system (SONIDEL) in unfocused mode. A 400 nm LED enabled visualization of release, compatible with narrow-band imaging at 415 nm on the Evis X1-CV1500 (Olympus, Japan). Control samples (unpatterned shell at 1 MHz; patterned shell at 2.5 MHz) were insonated at matched MI. Drug diffusion was quantified by video tracking (Blackmagic Cinema Camera 6K, Blackmagic Design, Australia) using Python 3.12.

During the in vivo release procedure, the ultrasound probe was positioned over the epigastric region, approximately 2 cm below the xiphoid process, initially in a transverse orientation (Fig. S12). The first landmark was the superior mesenteric artery, visible in cross-section as an anechoic structure surrounded by an echogenic wall. Just anterior to this, the comma-shaped portal–splenic confluence was identified, and the splenic vein was followed along the posterior surface of the pancreas. Sweeping the probe along the pancreatic body showed the pancreatic duct as a thin anechoic line within the gland, in which the acoustic artifact of the catheter was sought.

The operator then intermittently applied the imaging transducer to identify the relevant anatomy and verify the catheter's position within the duct from its acoustic artifact. Once the catheter had reached the selected region, the anchoring balloon was inflated using the insufflation trigger to secure the device intraluminally. Imaging was then discontinued, and payload release was initiated with a separate 1-MHz transducer while the catheter remained anchored. Drug delivery was also demonstrated in different gastrointestinal environments (esophagus, stomach, and colon) to test release across a range of pH conditions (Fig. S14).

During imaging, the thermal index (TI) and mechanical index (MI) reported by the iU22 scanner were 1.1 and 0.4. The SP300 release system reports neither index; for the 1-MHz transducer, TI and MI were derived from hydrophone measurements in water, derated by 0.3 $dB \cdot cm^{-1} \cdot MHz^{-1}$, which gave a peak negative pressure of 0.7 MPa (MI = 0.7) and TI = 1.9 at a duty cycle of 2%. Each release exposure lasted less than 2 min, within the exposure-time limits recommended for TI below 2.

Additional thermal characterization during ultrasound-triggered release in a tissue-mimicking medium showed only a transient local temperature increase of ~0.8°C, returning to baseline within 12 seconds after sonication ceased (Suppl. 8); together with the TI and MI values reported above, this indicates that the applied ultrasound exposures remained within the diagnostic safety framework used in this study.

### Task-autonomous cannulation control

Task-autonomous cannulation was enabled by an orientation estimation pipeline that couples a convolutional neural network detector (YOLOv8), trained to detect mSCR markers and the anatomical landmark (the papilla), with an algorithm that computes the geometric angle between these markers. The markers present on the mSCR (Fig. 4, B and D) consist of 1-mm sections of soft heat-shrink tubing, positioned 10 mm apart. Each marker has a distinct color, allowing it to be distinguished from the others and from the environment.

Because labeled in vivo datasets are scarce, we assembled a training set from phantom and fresh carcass trials, as well as from a prior exploratory in vivo procedure, spanning five endoscope models and three illumination settings, fluid interference, and partial obstruction, and applied data augmentation. We then trained the model to identify classes corresponding to the different markers so that, for each frame, among all detections, the one with the highest confidence score for each class of interest was retained. On a held-out evaluation set, the detector achieved a precision of 0.81, a recall of 0.76, and an F1 score of 0.78 (Suppl. 4).

The task-autonomous cannulation controller operated on the external endoscope video stream at 27 Hz. For each frame containing valid detections of the papilla and both colored catheter markers, the centroid of each detection was calculated. The distal catheter axis was defined by the vector connecting the two catheter-marker centroids, whereas the desired entry line was defined by the vector connecting the proximal marker to the papilla centroid. The signed image-plane angular error between these two vectors was used as the feedback variable for catheter alignment.

The angular error was mapped through the experimentally calibrated catheter deformation–magnetic-field relationship to a desired catheter deformation and corresponding magnetic command. The dipole-based magnetic model was then used to determine the target pose of the external permanent magnet. The 7-DoF robotic manipulator realized the resulting incremental EPM-pose command using a damped Jacobian pseudoinverse. A modified saturation-in-the-null-space term was applied to prevent the robotic manipulator from approaching its joint limits while preserving the commanded end-effector motion.

Catheter advancement was enabled only when the papilla and both catheter markers were reliably detected and the angular error satisfied the predefined alignment condition (< 3°). Advancement was set to zero whenever the alignment condition was violated or any required visual detection was missing. The controller then commanded the EPM to return to the last configuration in which the required visual features had been reliably detected before restarting advancement. Once the distal marker entered the papilla, the set of required detections changed to the proximal marker and the papilla: the EPM pose was held at the last two-marker alignment, with FBG monitoring of the distal orientation, while advancement continued. The task ended when the proximal marker reached the orifice, after about 10 mm, and control returned to the operator.

For safety, the commanded orientation change was compared with the FBG shape reconstruction. At each control cycle, the absolute difference between the model-predicted and the FBG-reconstructed distal orientation was evaluated. A model–sensor angular discrepancy exceeding 3° for four consecutive control cycles, corresponding to approximately 150 ms, triggered a rollback to the last consistent EPM configuration. If model–sensor consistency or reliable visual tracking could not be re-established, the current EPM pose was held. Similarly, if any required visual detection was unavailable or failed the detector-reliability criterion (confidence score of at least 0.65 for the papilla and its entry point and 0.45 for the catheter markers; Fig. 4, B and D), translational advancement was set to zero and the EPM returned to the last configuration with reliable detections, where it was held until tracking recovered. The alignment error was recomputed only after all required detections recovered. FBG monitoring remained active during visual dropout but was used only for catheter-configuration consistency monitoring. Task-autonomous cannulation could resume only after the required detections and consistency conditions had been restored. The robot also stopped if the external joint torque exceeded 5 N·m. Operator override and the emergency-stop function remained available throughout the subroutine.

**In vivo procedures**

In vivo experiments were conducted on two Yorkshire–Landrace gilts weighing 33 kg and 35 kg under terminal general anesthesia at the Large Animal Experimental Facility, University of Leeds. All procedures complied with the UK Animal (Scientific Procedures) Act 1986 under Home Office license PPL PP3093153 and followed NC3Rs and ARRIVE guidelines.

In line with the protocol described in [82], a high-energy liquid diet was used to produce a clean gastrointestinal tract without causing detectable physiological or behavioral abnormalities. Both gilts were placed in the supine position. The endoscope was advanced 90–100 cm through the upper gastrointestinal tract to the minor papilla and the mSCR was advanced into the pancreatic duct, with insufflation and suction or irrigation used as required.

Gallbladder access was attempted four times in the 33-kg animal and succeeded three times, by teleoperated magnetic manipulation under onboard visualization; it was not specifically attempted in the second animal. After cannulation of the papilla, the mSCR was advanced along the common bile duct to the junction of the CHD and the cystic duct, where the cystic duct was selected and followed to the gallbladder neck; the catheter was then advanced into the gallbladder lumen. Fluid was aspirated and air insufflated alternately through the 100-μm channel to maintain visibility, and the gallbladder was identified visually by the operator on the onboard camera view. Insertion distances were obtained from the advancing unit and the FBG shape sensor.

In a separate procedure, navigation was also demonstrated in the porcine lung with the bronchoscope, down to seventh-generation subsegmental branches and with sequential steering into two distinct airway branches (Fig. S15).

Species-specific anatomical differences, including the more distal minor papilla, were accounted for during procedures. However, despite this variation, pigs have long been employed as models for human gastrointestinal research [45], [46], [83].

**Sterilization and preliminary in vitro cytocompatibility assessment**

To be as close as possible to a clinical scenario, prototypes used during the in vivo trial were processed according to ISO 17664 (processing information) and ISO 17665 (moist-heat sterilization), and cytotoxicity was assessed with a protocol adapted from ISO 10993-5 [84].

On the morning before the trials, catheters were cleaned in 500 mL of hospital-grade disinfectant solution (Rely+On Virkon), rinsed with water, sealed in sterilization pouches, and autoclaved at 121°C for 30 min followed by 30 min of drying. To verify that this process did not impair device function, we performed pre- and post-sterilization characterization of the main functional modules. Tip angles after autoclaving were within 7% of those before; gripper lifetime and actuation force remained comparable to non-sterilized controls (273 ± 13 versus 261 ± 10 cycles; 0.477 ± 0.1 N versus 0.475 ± 0.1 N respectively); and balloon burst pressure was unchanged, although radial deformation decreased by approximately 10% after sterilization. Drug payloads were loaded, and the release shell bonded, only after sterilization under aseptic conditions, so they were not affected by the autoclave process; the shell itself was therefore not sterilized, and sterilization of the loaded shell for clinical use remains to be validated. Camera image and FBG signal were checked after autoclaving, before each trial. Full experimental methods and results are provided in Suppl. 7, 10. The mSCRs were kept sealed until use and disposed of afterward according to infectious healthcare waste protocols.

To assess the cytotoxicity of the silicone (DragonSkin 30) and of the DragonSkin 30–NdFeB composite (3.7:1 by mass) used to fabricate the mSCR, the viability of human fetal foreskin fibroblast 2 (HFFF2) cells exposed to extracts of, and in direct contact with, these materials was measured over 24 hours with a protocol adapted from ISO 10993-5 [84] (Suppl. 10). Untreated cells served as the reference, and NdFeB particles alone (diluted 1:10) served as the positive (cytotoxic) control. Cell viability was quantified using the CellTiter-Glo® 2D Cell Viability Assay (Promega), which measures intracellular ATP as an indicator of metabolically active cells. Viability relative to untreated cells was 86–97% for all test materials, above the 70% threshold of ISO 10993-5, and 92% for the NdFeB particles of the positive control.

**Onboard video analysis and lumen reconstruction**

Onboard videos (24 frames per second) were analyzed in Python with OpenCV; frames shown in figures are unprocessed apart from cropping and are the sharpest frame within ±0.5 s of the stated time. In each frame, the lumen was segmented as the largest connected region darker than 0.45 times the median luminance (the two largest in the junction frame), and papilla openings as the darkest region around the orifice after illumination correction. Camera rotation and forward or backward motion were estimated from dense optical flow with an affine model, after excluding specular, dark, and textureless pixels, and were validated against insertion-unit and FBG data. For the relative reconstructions (Fig. 5, B and E), the video was combined with the insertion distances obtained from the advancing unit and the FBG shape sensor: each segment of the path was scaled to the measured insertion distance (common bile duct, 2.8 cm; CHD, 2.24 cm; cystic duct, 5.1 cm; pancreatic duct, 7.5 cm), and its local direction followed the camera rotation and the position of the lumen in the image. Cross-sections were back-projected at an assumed viewing distance of 3.2 mm, the CHD and cystic duct were drawn with an assumed 25° angle, and the junction was lofted from the two lumens segmented in the same frame.

### Statistical reporting

Throughout the manuscript, n denotes the number of independent biological or experimental samples, whereas N denotes the total number of trials or measurements. Results are reported as mean ± SD unless otherwise specified.

**Acknowledgments:** We thank E. Murphy, J. Bilton for their support in organizing and performing the animal trial, S. Wilson and A. Alexander for their logistical assistance, T. Dean for assistance with $^{1}$H NMR measurements, and J. R. McLaughlan for his help with the thermo-acoustic analysis.

**Funding:** This work was supported by the Advanced Research and Invention Agency (ARIA) under grant SMRB-SE01-P01, the European Research Council (ERC) through the European Union's Horizon 2020 Research and Innovation Programme under Grant 818045, the Engineering and Physical Sciences Research Council (EPSRC) under grant EP/V009818/1, and J. Chandler was supported by the Leverhulme Trust and the Royal Academy of Engineering under a RAEng/Leverhulme Trust Research Fellowship (LTRF-2425-21-154). Any opinions, findings and conclusions, or recommendations expressed in this article are those of the authors and do not necessarily reflect the views of the ARIA, the ERC, the EPSRC, or the RAEng/LTRF.

**Author contributions:** Conceptualization: B.C. Methodology: B.C., A.M., N.J.G. Investigation: B.C., A.M., D.K. Visualization: B.C. Funding acquisition: B.C., P.V. Project administration: B.C., J.H.C., P.V. Supervision: P.V., V.P. Writing – original draft: B.C., N.J.G., G.L., J.H.C. Writing – review & editing: B.C., N.J.G., A.M., A.B., P.L., J.H.C., V.P., P.V.

**Competing interests:** Authors declare that they have no competing interests.

**Data and materials availability:** All data needed to support the conclusions of this manuscript are included in the main text or Supplementary Materials, including the deposit in the Dryad database: 10.5061/dryad.dbrv15fhh

## Figure captions

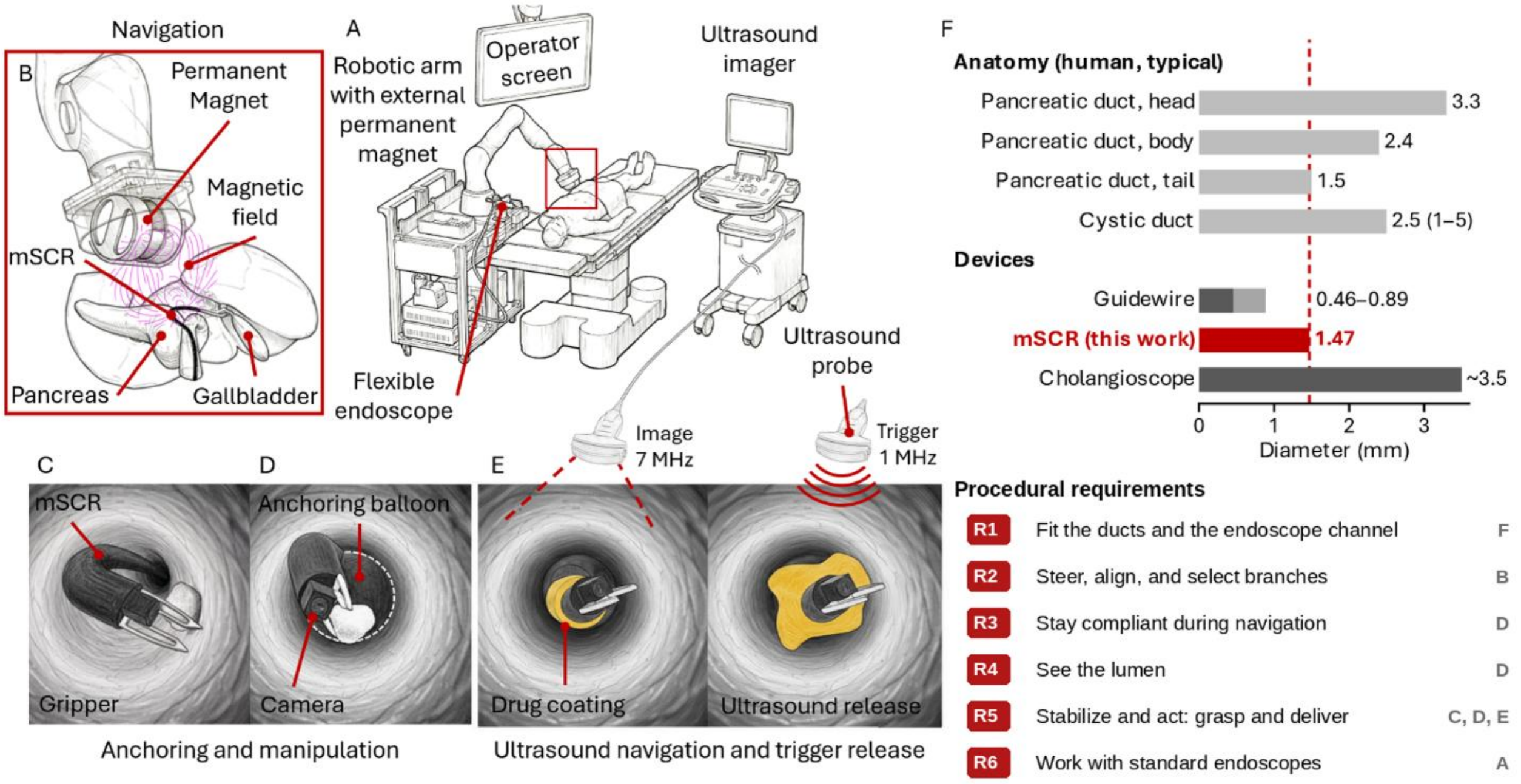

*Figure 1* ***Procedural requirements and modular mSCR for pancreatobiliary access.*** *(**A**) Robotic platform during the trials: robotic arm with an external permanent magnet, flexible endoscope, operator screen, and ultrasound imager. (**B**) Magnetic navigation of the mSCR in the pancreatobiliary anatomy. (**C** to **E**) Functional modules by clinical role: the mSCR with its compliant gripper (C); the anchoring balloon and the onboard camera (D); and the drug coating, imaged by ultrasound at 7 MHz and released by a 1-MHz trigger (E). (**F**) Size constraint: typical human duct diameters compared with guidewires (0.46–0.89 mm), the mSCR (1.47 mm), and a single-operator cholangioscope (about 3.5 mm), and the procedural requirements R1 to R6 (Table S1) with the panels that address them.*

*Table 1. Cost of each functionalization on cross-section, bending stiffness, and magnetic bending. Configurations are cumulative. Tip angle measured in a uniform field generated by the single-axis coil (n = 3 catheters, N = 15 per configuration and field). Cross-section relative to 1.7 mm²; shares of the composite bending stiffness (EI) computed from the composite modulus (3.04 MPa) and component data (Suppl. 2).*

| Configuration | Conventional implementation and its cost | Implementation in the mSCR | Cross-section (%) | Share of EI (%) | Tip angle at 8 mT (°) | Tip angle at 16 mT (°) |
|---|---|---|---|---|---|---|
| 1 Magnetic body with 100-µm channel and balloon lumen | Internal steering (tendons, tubes): friction, backlash, additional lumens | Remote magnetic actuation of a molded NdFeB–silicone body | 2.4 (lumens) | 33.7 (body) | 86.0 | 89.2 |
| 2 + camera cable and LED fiber | Separate scope or channel: larger diameter | Guides molded in the body | 10.3 | 23.4 | 80.1 | 87.1 |
| 3 + multicore FBG | External tracking or additional sensor cabling | Fiber in a molded guide | 1.9 | 42.8 | 69.7 | 82.0 |
| 4 + 80-µm tendon | Large closure force that bends the body | Nylon tendon; flexures optimized for minimum tension | 0.3 | 0.2 | 69.6 | 81.9 |
| 5 + camera housing and gripper | Tool through a working channel or sleeve: larger diameter | Gripper bonded to the camera housing | on camera footprint | not estimated | 66.4 | 80.1 |
| 6a + magnetic balloon (5 mm) | Non-magnetic balloon segment (6b) | Balloon molded and magnetized with the body | lumen in 1 | not estimated | 61.8 | 77.2 |
| 6b + non-magnetic balloon (5 mm) | – | – | lumen in 1 | not estimated | 25.3 | 44.4 |
| 7 + PLA release shell (10 µm, 8 mm) | Onboard actuator or hinged compartment | Patterned coating released by 1-MHz ultrasound | external layer | not estimated | 60.1 | 75.1 |

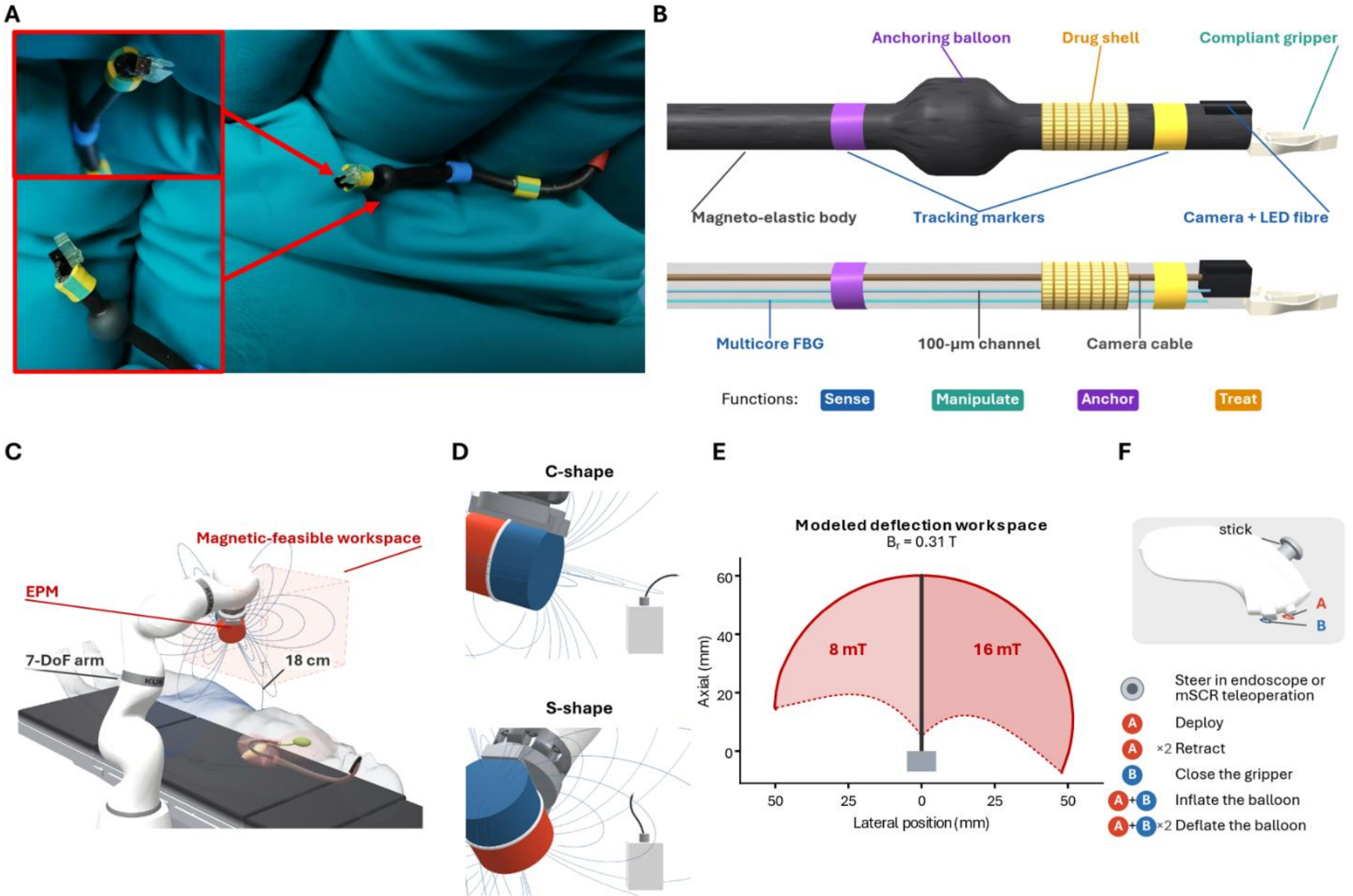

*Figure 2* ***Modular mSCR, magnetic actuation, and operator interface.*** *(**A**) The modular mSCR with its gripper, anchoring balloon, and tracking markers. (**B**) Rendering of the 1.47-mm tip, exterior (top) and cut-away (bottom), with modules colored by function. The magneto-elastic body is DragonSkin 30 loaded with NdFeB (3.7:1 by mass). The camera (NanEye, 320 × 320 pixels) is paired with an LED fiber, and the tracking markers are 1-mm bands 10 mm apart. The anchoring balloon is molded in the body, the drug shell is ultrasound-triggered PLA, and the compliant gripper is actuated by an 80-µm tendon. A multicore FBG for shape sensing and a 100-µm channel for flushing and aspiration run inside the body. (**C**) Robotic platform: a 7-DoF arm (KUKA LBR Med 14) carries the N52 EPM (101.6 × 101.6 mm; K&J Magnetics). The magnetic-feasible workspace (0.05 m³) contains the EPM poses that give 8–16 mT at the mSCR, with the magnet face 18 to 24 cm from the catheter; blue lines show the EPM field. (**D**) Shape control by the EPM pose, illustrated with the catheter clamped at its base (60 mm): with the EPM beside the body and its dipole pointing at it, the catheter forms a C-shape; turning the dipole along the body gives an S-shape. (E) Modeled deflection workspace of the mSCR (60-mm free length) with gripper and magnetic balloon (configuration 6a; remanence 0.31 T, component stiffnesses of Table S2) in a uniform 8 mT (left) or 16 mT (right) field rotated from 0° to 150°. Solid line, tip positions; dashed line, catheter shape at 150°; tip angle 66° and 80° for a perpendicular field and up to 104° and 131° at 150°. (**F**) Joystick mapping. The stick steers in either teleoperation mode: endoscope view (the EPM moves to bend the mSCR) or mSCR camera view (the tip is steered, with the EPM pose solved automatically). A deploys and a double click on A retracts (advancing unit); B closes the gripper; A + B inflates the balloon, and a double click deflates it.*

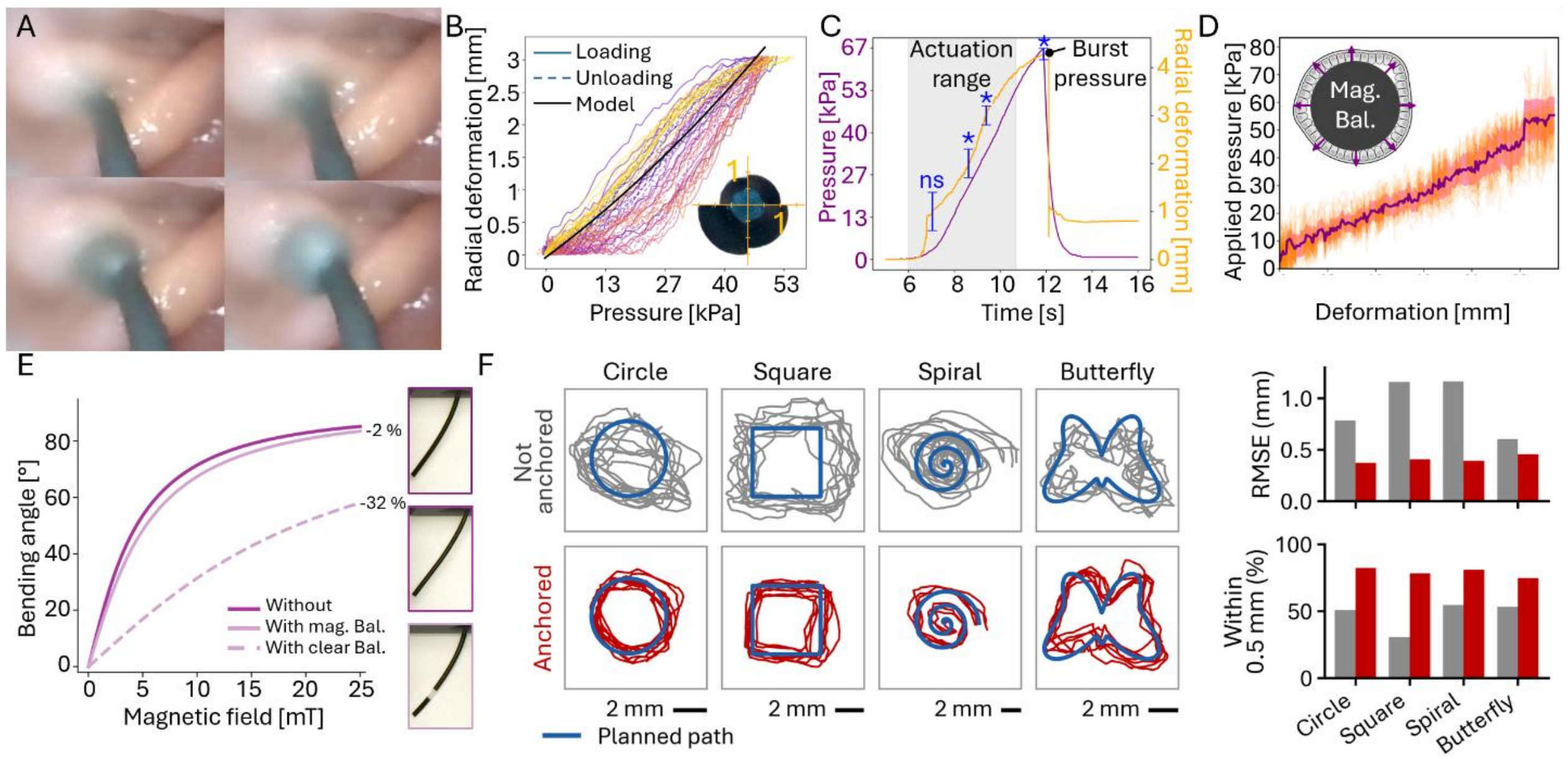

*Figure 3* ***Distal grounding through a magnetic anchoring balloon.*** *(**A**) Inflation of the anchoring balloon in the minor papilla in vivo, seen from the endoscope. (**B**) Model-predicted and measured large-strain radial deformation of the magnetic balloon during loading and unloading (n = 3, N = 30); inset, balloon cross-section. (**C**) Pressure and radial deformation during inflation up to bursting at 64 kPa, defining the safe actuation range of the inflation unit (n = 3, N = 30); ns, not significant. (**D**) Inflation pressure of the magnetic balloon pressed against the force sensor (Fig. S8C) as a function of radial deformation (n = 3, N = 30). (**E**) Bending angle as a function of the applied magnetic field without a balloon, with the magnetic balloon, and with an otherwise identical non-magnetic (clear) balloon; insets, catheter at 25 mT in each configuration; single-axis coil, 0–25 mT in 5-mT steps (n = 3, N = 12). (**F**) Tip trajectories under an electromagnetic navigation system without anchoring (top, gray) and with the balloon inflated (bottom, red) for four planned paths (blue); scale bars, 2 mm (each shape at its own scale, the same in both rows). Bars give the tip-tracking RMSE and the percentage of tip positions within 0.5 mm of the path. The planned path was registered from the commanded trajectory, without fitting to either run (Suppl. 6); one run per condition and path.*

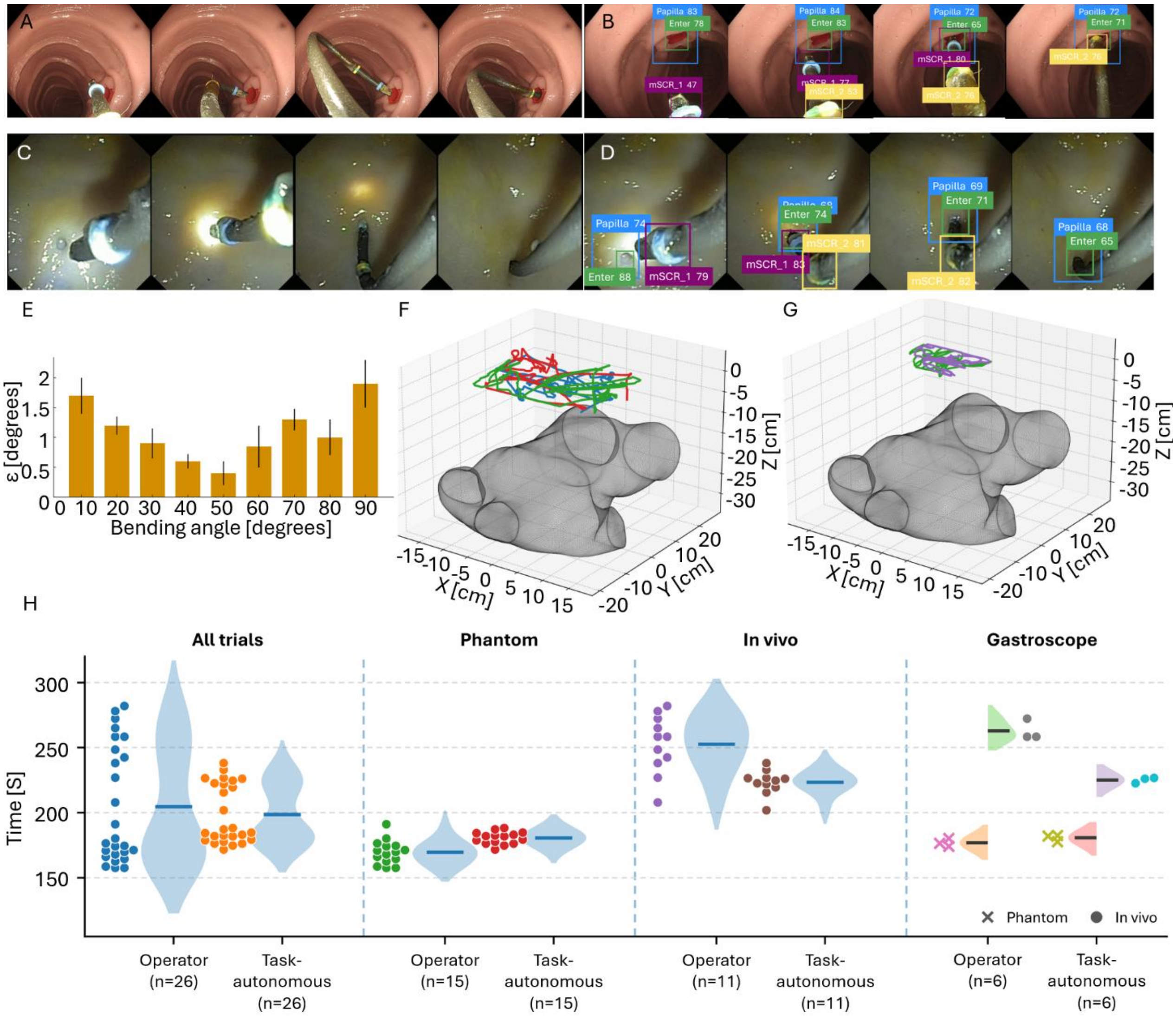

*Figure 4* ***Teleoperated and task-autonomous papilla cannulation.*** *(**A**) Operator-controlled cannulation of a realistic phantom papilla, seen from the endoscope. (**B**) Task-autonomous cannulation of the same phantom, with detections of the papilla, its entry point, and the two catheter markers (mSCR_1 and mSCR_2) and their confidence scores. (**C**) Operator-controlled cannulation in vivo (porcine model). (**D**) Task-autonomous cannulation in vivo, with detections. (**E**) Error ε between the bending angle estimated by the magnetic model and that measured by the FBG, as a function of the bending angle (distal orientation relative to the endoscope channel exit). (**F** and **G**) EPM trajectories during operator-controlled (**F**) and task-autonomous (**G**) cannulation; the meshed volume indicates the pig's body. (**H**) Procedure time under operator control and task autonomy, overall and for phantom trials (n = 1, N = 30, 15 per mode), in vivo cannulation (n = 2, N = 22, 11 per mode), and the subset of these procedures performed with a gastroscope instead of a duodenoscope (crosses, phantom; dots, in vivo); each point is one trial and lines mark the means.*

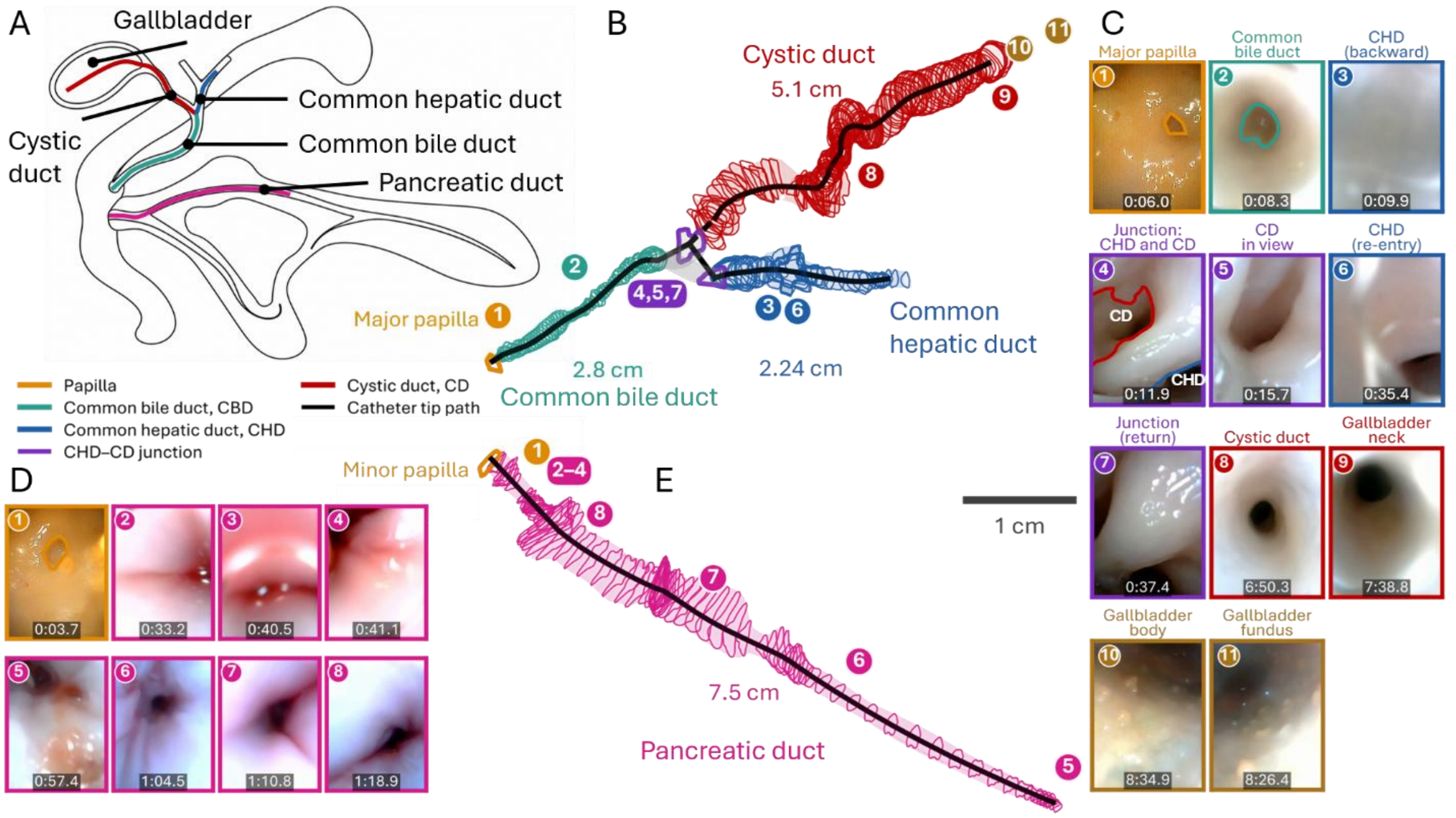

*Figure 5* ***Visualized deep access in the pancreatobiliary tree (porcine model).*** *(**A**) Pancreatobiliary anatomy with the ducts reached by the mSCR. (**B**) Relative 3D reconstruction of the route from the major papilla to the gallbladder: common bile duct (2.8 cm), common hepatic duct (CHD; 2.24 cm), and cystic duct (CD; 5.1 cm). (C) Transpapillary access to the gallbladder seen from the onboard camera (times in min:s from the start of each clip; frames were taken from separate clips of the same access): major papilla (1), common bile duct (2), withdrawal in the CHD (3), junction of the CHD and the CD, with both lumens segmented (4), CD in view (5), re-entry into the CHD (6), return to the junction (7), navigation in the CD (8), gallbladder neck (9), body (10), and fundus (11). (**D**) Navigation in the pancreatic duct seen from the onboard camera: minor papilla (1), narrow proximal duct (2 and 3), entry (4), debris in the duct (5), and the duct lumen (6 to 8), acquired during withdrawal (Fig. S16). (**E**) Relative 3D reconstruction of the pancreatic duct along the 7.5-cm path. In (B) and (E), segmented lumen cross-sections are placed along the catheter path estimated from the optical flow and scaled to the insertion distances; numbers refer to the frames in (C) and (D), both reconstructions share one scale bar (1 cm), and cross-section size depends on an assumed viewing distance (Materials and Methods). Colors: papilla (orange), common bile duct (teal), CHD (blue), CHD–CD junction (purple), CD (red), gallbladder (brown), and pancreatic duct (magenta).*

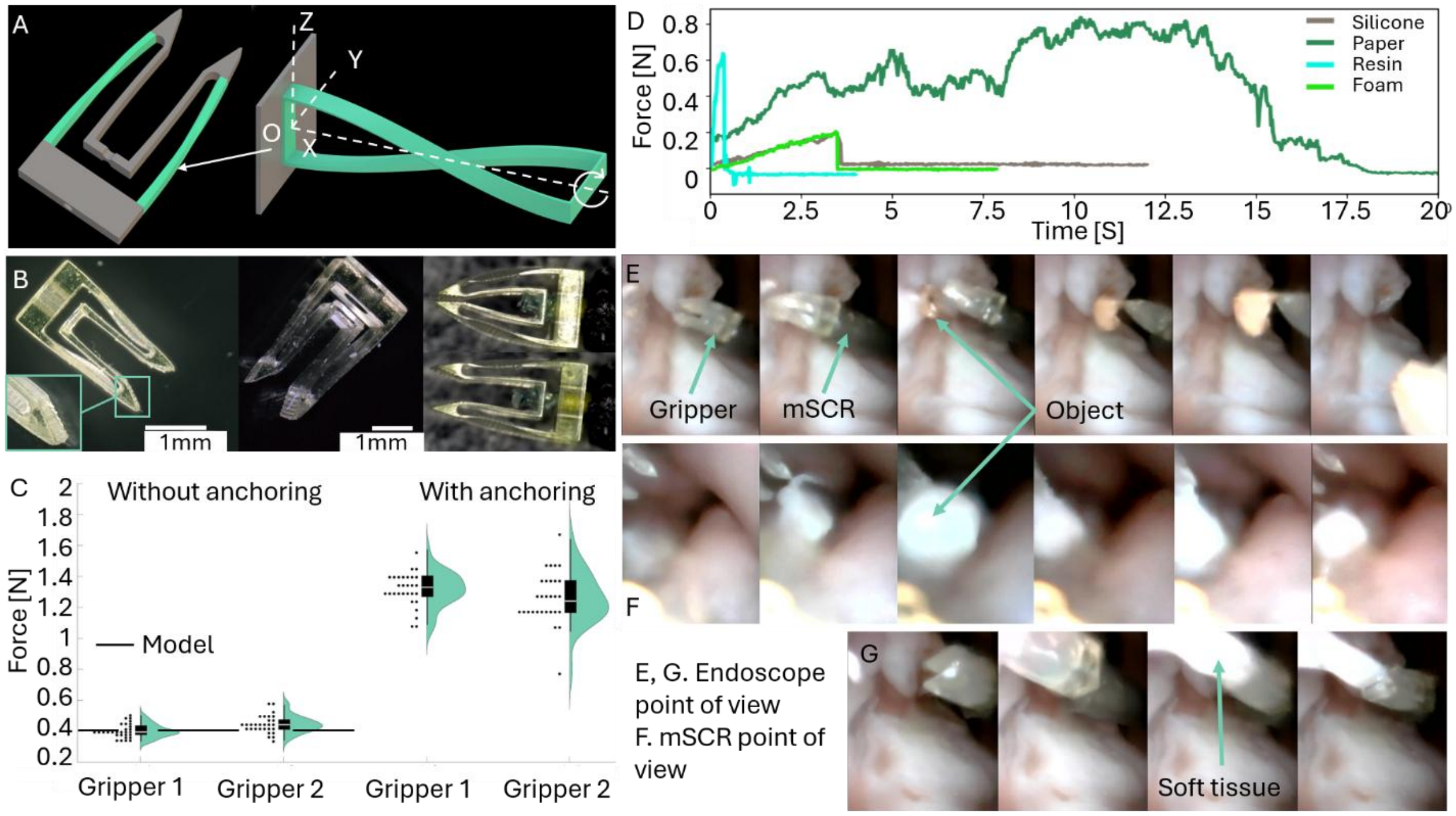

*Figure 6* ***Grasping and manipulation with a compliant gripper.*** *(**A**) Flexure-based L-arm gripper (left) and its blade with a 45° twist (right; O–XYZ, blade frame). (**B**) Microscopic images of the 3D-printed grippers: jaw pattern (inset) and open and closed*

*configurations; scale bars, 1 mm. (**C**) Gripping force of gripper 1 (straight blades) and gripper 2 (45°-twisted blades) without and with balloon anchoring, with the model prediction (n = 2, N = 60). (**D**) Gripping force over time on objects of different materials (silicone, paper, resin, and foam). (**E** to **G**) Manipulation in vivo: grasping and removing an object seen from the endoscope (E) and from the mSCR camera (F), and manipulation of soft tissue seen from the endoscope (G).*

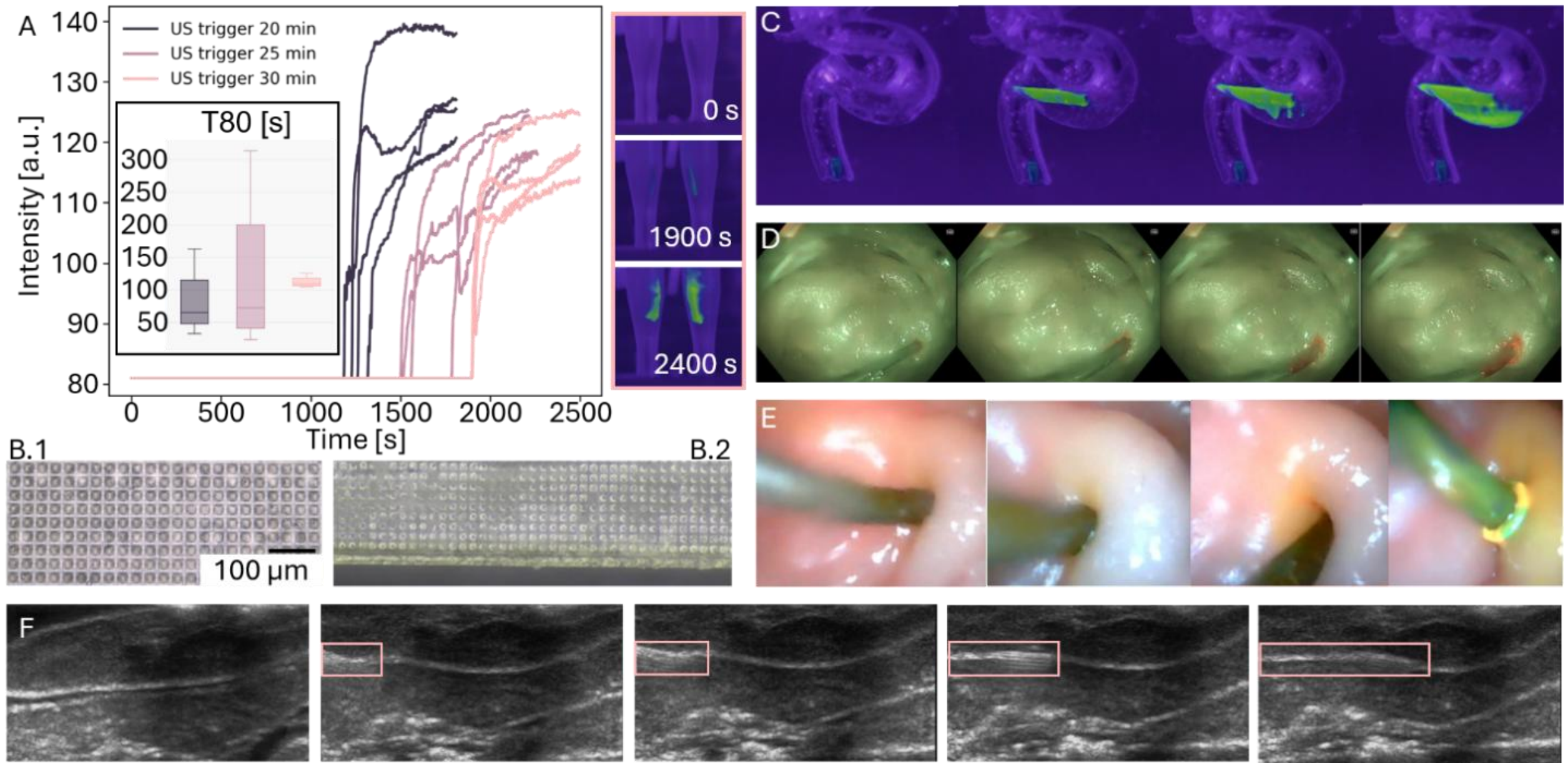

*Figure 7* ***On-demand local therapy with an ultrasound-triggered release shell.*** *(**A**) Time-controlled release in vitro: release intensity after a 1-MHz trigger applied at 20, 25, or 30 min of continuous 7.2-MHz insonation, with the 80% release time (T80) of each batch (inset; n = 4, N = 12) and representative frames at 0, 1900, and 2400 s. (**B**) The mold was 3D-printed with the micropattern (B.1; scale bar, 100 µm) and coated with payload-doped PLA by spin coating (B.2) to fabricate the shell that is then assembled onto the mSCR. (**C**) Controlled release in an aneurysm phantom. (**D**) Ex vivo assessment of release using narrow-band imaging. (**E**) Released payload visible at the papilla in vivo as the catheter is retracted, seen from the endoscope. (**F**) The catheter visible on in vivo ultrasound images (boxes).*

# Supplementary Materials for

## **Modular magnetic soft robotic catheter enables visualized intervention in the pancreatobiliary tree**

Benjamin Calmé *et al.*

Corresponding author: Benjamin Calmé, b.p.calme@leeds.ac.uk

The document includes:

Other Supplementary Material for this manuscript includes Movies S1 to S7:

- Movie S1: Overview of the modular mSCR along the steps of the procedure
- Movie S2: Modular magnetic soft robot
- Movie S3: Magnetic manipulation and papilla cannulation
- Movie S4: Transpapillary access to the gallbladder from the onboard camera
- Movie S5: Anchoring balloon
- Movie S6: Compliant gripper
- Movie S7: Drug delivery

**Suppl. 1 - Comparison of the mSCR with previously reported miniaturized functional catheters and soft robotic interventional devices**

| Ref. | Application | Functionalization | Dimension | Material | Actuation | Control | Validation |
|---|---|---|---|---|---|---|---|
| [85] | Endovascular | Hydrogel coating; micro-object manipulation | 200 µm | PDMS + NdFeB | Magnetic | Manual + Open-loop | Bench (microfluidic channels) |
| [27] | Endovascular | Optical fiber / laser | 500 µm | Ferromagnetic soft polymer body with hydrogel skin | Magnetic | Manual | Bench (cerebrovascular phantom) |
| [86] | Endovascular | Rotative tip | 0.8 mm | PDMS + permanent magnet | Magnetic actuation | Manual + Open-loop | In vivo (rabbit) + ex vivo (human placenta) |
| [35] | Bile ducts / Endovascular | Artificial cilia, drug delivery | 1 mm | Ecoflex 30 + $Fe_3O_4$, PDMS | Magnetic | Manual | In vivo (rabbit) |
| [37] | Lung | Fiber optics; Drug delivery; Laser ablation | 0.95 mm | Ecoflex 30 + iron, tip in resin | Magnetic | Teleoperation | Ex vivo (porcine lung) |
| [28] | Endovascular | Helical rotating tip | 0.8 mm | Pebax + NdFeB | Magnetic | Teleoperation | In vivo (pig) |
| [87] | - | Soft gripper | 3 mm | PDMS + NdFeB | Magnetic | Open-loop | Bench |
| [88] | Kidney, Lung, GI | FBG triaxial force sensing + tool channel | 5 mm (tip) | Ecoflex 00-30 + NdFeB magnets | Magnetic | Teleoperation | Bench + ex-vivo |
| [66] | Kidney | Payload delivery and retrieval | 1.5 mm | Dragon Skin 30 + NdFeB | Magnetic | Teleoperation | Bench + ex vivo (porcine kidney) |
| [31] | Lung | Laser fiber + FBG shape sensing | 2.4 mm | Dragon Skin 10 + NdFeB | Magnetic | Open-loop | Phantom + human cadaveric lungs |
| [34] | Liver | Bioprinting | 2 mm | PDMS + NdFeB | Magnetic | Open-loop | In vivo (rat) |
| [89] | Articular cartilage | Camera, hydrogel injection, optical fiber | 4 mm | Silicone + permanent magnet | Magnetic | Closed-loop (visual servoing) | Ex vivo |
| [90] | Lung | Camera, tool channel | 5 mm | TPE film + carbonyl-iron fluid | Hydraulic eversion + magnetic steering | Teleoperation | Bench |
| [91] | Cardiac | Variable stiffness | 2.3 mm | Shape-memory polymer + permanent magnet | Magnetic + electrothermal | Open-loop | Bench |
| [30] | Endovascular | Contrast-agent and embolic-agent infusion | Reaches 180-µm arteries | Polyimide, permanent magnet | Magnetic steering + blood-flow-assisted navigation | Teleoperation | In vivo (pig) |
| [92] | Cardiac | Stabilization + tool guidance | 8 mm | TPU, Kapton | Hydraulic + cable | Teleoperation | Ex vivo |
| [93] | Endovascular | Coil deployment | 0.9 mm | Dragon Skin 10 | Microhydraulic | Manual | In vivo (pig) |
| [94] | Cardiac | Force sensing | 2.6–3.6 mm | Dragon Skin 30, PDMS, kevlar | Pneumatic | Closed-loop force control | Bench |
| [95] | Endovascular | Tool channel | 2.6 mm | Dragon Skin 30 | Hydraulic | Open-loop | In vivo (pig) |
| [52] | Lung | Tip steering + stabilization + biopsy | 3.5 mm | Ecoflex 00-30, DragonSkin 10, TPE, PTFE | Fluidic | Manual | Bench |
| [96] | Eye (mock-up) | CMOS camera | 1.66 mm | PDMS | Pneumatic | Open loop | Bench |
| [97] | Lung | Camera, biopsy tool | 2.4 mm | Dragon Skin 10 | Pneumatic/fluidic | Teleoperation + vision-assisted alignment | Ex vivo |
| [98] | Lung | Camera, tool channel | 4.5 mm | Silicone, NiTi rod | Push/pull NiTi rods | closed-loop | Bench |
| [99] | Eye | gripper | 1 mm | LMPA | Electrothermal / magnet | closed-loop | Bench |
| [100] | Liver | Fiber | 1.65 mm | PEEK | Thermal | closed-loop | In vivo (pig) |
| **Proposed approach** | **Endoluminal, pancreatic and bile duct, lung** | **Anchoring, manipulation, sensing, targeted drug delivery. (+ task-autonomous control)** | **1.47 mm OD** | **Dragon Skin 30 + NdFeB** | **Magnetic actuation** | Task-autonomous (teleoperation + task-autonomous cannulation) | **In vivo (pig)** |

This supplementary table shows that the main contribution of the mSCR lies not in being the smallest catheter reported, as several sub-millimeter systems remain better suited to the narrowest lumens, but that it combines capabilities that are rarely integrated in a single platform. In particular, modularity is itself uncommon among miniaturized soft interventional catheters, and the addition of embedded sensing and task-autonomous cannulation is even less common. The proposed mSCR brings these elements together in a 1.47 mm outer-diameter soft catheter that integrates anchoring, manipulation, sensing, and targeted drug delivery within one modular architecture. It was also tested in vivo in the pancreatobiliary tree.

## Suppl. 2 - Stiffness budget and bending model of the functionalized mSCR

The bending stiffness of the magneto-elastic body was computed from the composite modulus ($E = 6C1 = 3.04$ MPa, Table S4) over the 1.47-mm section minus the lumens, and that of each component from its diameter and supplier modulus (Table S2). The components were assumed to slide freely in their guides, so that their stiffnesses add. On this basis, the camera cable, LED fiber, and FBG, which occupy 12% of the cross-section, carry 66% of the bending stiffness, and the catheter behaves as a homogeneous rod with an equivalent modulus of 9.0 MPa.

The measured tip angles of Table 1 were compared with a magneto-elastic Kirchhoff rod model clamped at the endoscope exit (free length 60 mm) and loaded by the distributed magnetic torque $m'B \sin(\alpha - \theta)$, where $m' = BrA/\mu 0 = 0.357$ A·m is the magnetic moment per unit length for $Br = 0.31$ T, $A \approx 1.45$ mm² being the net magnetic cross-section (85.2% of the 1.7-mm² section; Table S2), B and α are the magnitude and direction of the uniform field, and θ is the local tangent angle. The 4-mm rigid tip and the balloon section were modeled as separate segments; the magnetic balloon was given the properties of the body, the non-magnetic balloon those of a non-magnetized segment, and the release shell an additional bending stiffness.

Because the magnetic body approaches alignment with the field at 8 mT, the field that the model requires for a tip angle of 60° separates the configurations more clearly: 1.8 mT for the magnetic body, 3.1 mT with the camera and LED, 5.4 mT with the FBG, and 6.3 mT for configuration 6a. Up to 16 mT, the modeled surface strain stays below 10% (9.4% for the magnetic body alone, 5.5% for the fully functionalized mSCR).

*Table S2 Bending-stiffness budget of the mSCR (components assumed to slide freely in their guides).*

| **Component** | **Diameter (mm)** | **Cross-section (%)** | **EI (N·m²)** | **Share of EI (%)** | **Source** |
|---|---|---|---|---|---|
| Magneto-elastic body (E = 3.04 MPa) | 1.47 | 85.2 | $6.92 \times 10^{-7}$ | 33.7 | Computed from C1 (Table S4) |
| Camera cable (without PTFE jacket) | 0.40 | 7.4 | $1.00 \times 10^{-7}$ | 4.9 | supplier |
| LED fiber | 0.25 | 2.9 | $3.80 \times 10^{-7}$ | 18.5 | supplier |
| FBG fiber (125-µm silica, coated) | 0.20 | 1.9 | $8.80 \times 10^{-7}$ | 42.8 | supplier |
| Nylon tendon | 0.08 | 0.3 | $5.0 \times 10^{-9}$ | 0.2 | Computed |
| Channel and balloon lumens | – | 2.4 | 0 | 0 | – |
| Total | 1.47 | 100 | $2.06 \times 10^{-6}$ | 100 | Sum ($E_{eff}$ = 9.0 MPa) |

## Suppl. 3 - Robotic control architecture and safety framework

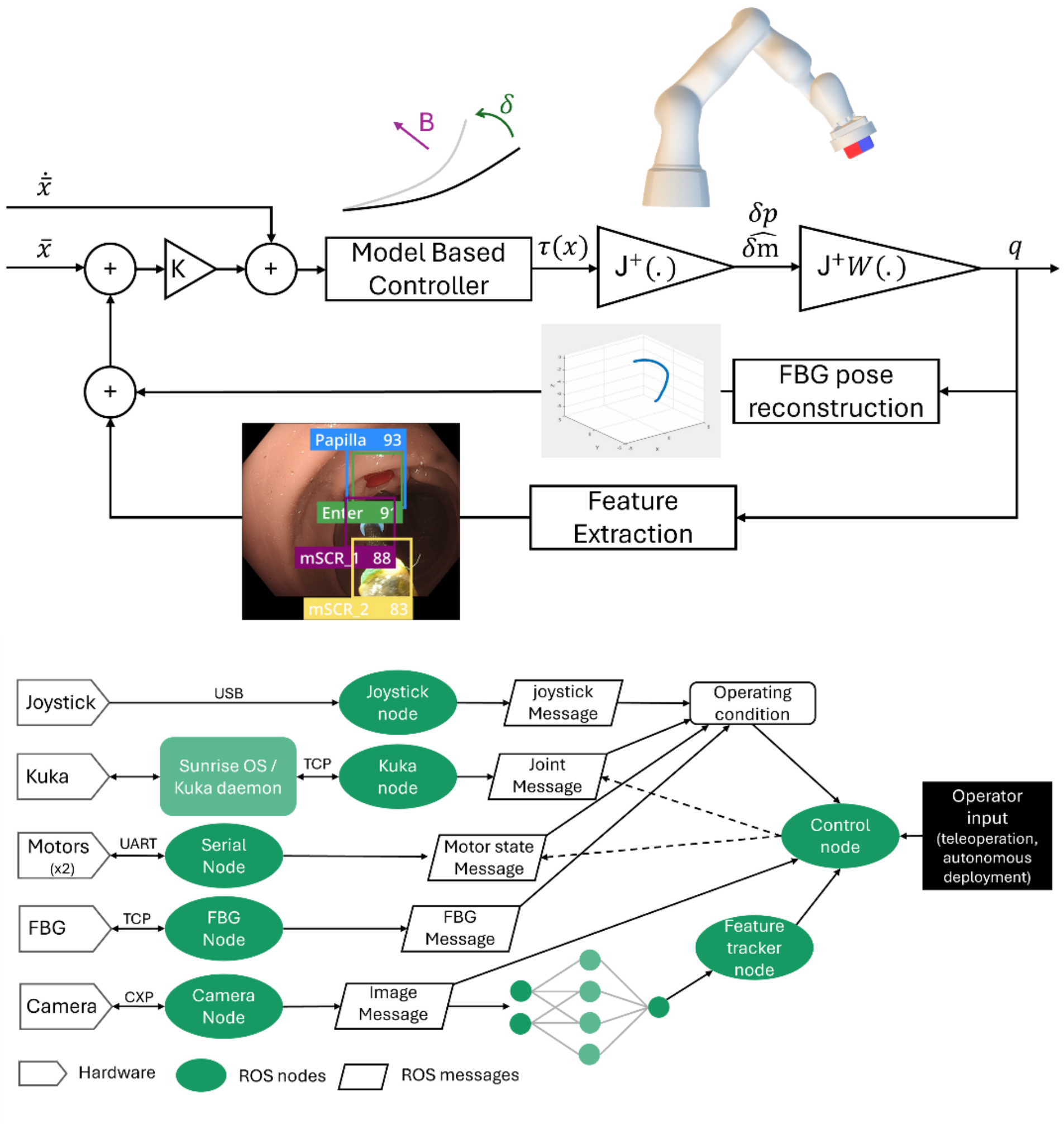

*Figure S17* ***Comprehensive diagram of the robotic platform control framework and ROS architecture.***

Fig. S17 illustrates both the conceptual control loop and the ROS-based software architecture of the mSCR platform. In the upper panel, the controller combines model-based magnetic actuation with multimodal feedback from endoscopic image features and FBG-based pose reconstruction. This feedback is used to estimate the current catheter state, compare it with the desired target state, and generate the magnetic command sent to the robotic manipulator carrying the external permanent magnet. In the lower panel, the ROS architecture is shown as a set of interconnected hardware and software nodes. The joystick, KUKA robot, motor drivers, FBG interrogator, and camera each communicate through dedicated interface nodes, while the feature-tracking node processes the image and the control node generates the actuation commands according to the current operating mode. This architecture enables coordinated teleoperation, task-autonomous cannulation, and system-level safety supervision.

***Direct robot teleoperation.*** This control layer provides the core functionality for teleoperating the external permanent magnet (EPM) in order to indirectly actuate the mSCR. The end effector of the robotic arm is operated via a joystick interface connected to a low-level controller. Operator commands provided through the joystick define the desired motion of the robot end effector. These inputs, consisting of angular displacements $\delta\hat{m} \in \mathbb{R}^3$ and linear displacements $\delta p \in \mathbb{R}^3$, are converted into joint-angle variations $\delta q \in \mathbb{R}^7$ using the differential kinematic relation, where $J^+ \in \mathbb{R}^{7\times 6}$ is the pseudoinverse of the robot Jacobian. To avoid joint limits, a modified saturation-in-the-null-space algorithm is used, introducing an additional corrective term $\delta q_0 \in \mathbb{R}^7$ in the null space of the Jacobian. This term is scaled by a gain related to the robot velocity, which was manually tuned.

***Task-autonomous cannulation.*** Task-autonomous steering of the mSCR relies on real-time tracking of the catheter tip position. From the tip Cartesian pose observed in the endoscopic image and the desired target position, defined here as the center of the papilla, the required catheter deformation is computed. A model-informed controller then determines the magnetic field and corresponding magnetic torque $\delta\tau \in \mathbb{R}^3$ required to achieve this deformation, yielding the associated $\delta\hat{m}$ and $\delta p$. These commands are subsequently applied using the same robot-level control strategy as in teleoperated mode. The resulting deformation is validated through image tracking and compared with the FBG-based position reconstruction. Any discrepancy between the predicted and measured catheter configuration is treated as a control error and compensated accordingly. When image-based tracking and FBG reconstruction remain aligned, deployment of the mSCR continues until a new misalignment is detected.

The velocity input governing translational advancement of the mSCR is directly proportional to the alignment between the catheter tip and the center of the papilla, and is set to zero when the alignment error exceeds 3° or a required detection is missing. Once the distal marker has entered the papilla, the required detections change to the proximal marker and the papilla; the EPM pose is then held at the last two-marker alignment, with FBG monitoring of the distal orientation, until the proximal marker reaches the orifice and control returns to the operator.

***Safety***. In teleoperated mode, collision avoidance with external instruments in the workspace is not automated and therefore relies on the operator to maintain a clear and safe trajectory. However, a configurable hard-coded height limit is imposed to prevent the robot from entering the patient space. This limit can be adjusted before the procedure according to the patient anatomy and setup.

In task-autonomous mode, safety is ensured through the same workspace height restriction together with a torque-limiting strategy. In the event of unintended contact with an object, the robotic arm detects the interaction through an external joint-torque threshold of 5 N·m and stops its motion. Although this does not constitute full collision avoidance, it provides a fail-safe response to unexpected contact during task-autonomous execution.

During task-autonomous catheter insertion, loss of tracking or decoupling immediately stops advancement and initiates rollback to the last configuration with reliable visual detections. If visual tracking is not recovered, the EPM pose is held. FBG reconstruction independently monitors model–sensor consistency. A distal-orientation discrepancy exceeding 3° for four consecutive control cycles triggers rollback to the last model–sensor-consistent EPM configuration. If consistency cannot be restored, the EPM pose is held and task-autonomous cannulation remains suspended. FBG sensing does not replace visual identification of the papilla or catheter markers.

As a final safety layer, an emergency-stop button is always positioned next to the operator and immediately overrides all commands, stopping the robot in its current configuration.

**Suppl. 4 - Dataset for the AI based detection model**

To train and validate the model, we used 3200 endoscopic images collected during various phantom, ex vivo, and in vivo experiments to fine-tune the base YOLOv8 model. As one of the strengths of the mSCR is its platform versatility, five different endoscopes (Table S3) were used, resulting in images with two different resolutions (720×480 and 1920×1080) and three different brightness levels.

We created the dataset by manually annotating one third of the images with bounding polygons using Roboflow; the remaining images were labeled with a model trained on this subset and corrected manually. A validation check was performed to ensure correct labeling of all images. The dataset was then partitioned into training, validation, and test sets with proportions (number of images) of 70% (2,240), 20% (640), and 10% (320), respectively.

Furthermore, to improve robustness, we generated augmented versions of each original training image using (i) horizontal and vertical flips with 50% probability, (ii) exposure adjustments between −9% and +9%, (iii) Gaussian blurring between 0 and 0.25 pixels, (iv) hue shifts between −25 and +25, and (v) cropping with up to 23% zoom. As a result, the number of images in the training set was quintupled to 11,200.

**Suppl. 5 - Modeling of a thick cylindrical anchoring balloon**

To model the balloon, we first define a reference (undeformed) configuration in which each material point is described by the coordinates (R,Θ,Z), and a deformed configuration described by (r,θ,z) . In this setting, torsional effects are neglected; thus, under an internal pressure $p_{in}$, an external pressure $p_{ex}$, and an external axial force $f_{ex}$, the magnetic balloon undergoes radial (R) and axial (Z) deformation. Therefore, in cylindrical basis, the deformation gradient **F** is given by:

$$\boldsymbol{F} = \begin{bmatrix} \frac{\partial r}{\partial R} & 0 & \frac{\partial r}{\partial Z} \\ 0 & \frac{r}{R} & 0 \\ 0 & 0 & \frac{\partial z}{\partial Z} \end{bmatrix}$$

Since the cylinder is closed at both ends, the radial deformation is not uniform along its longitudinal axis. The deformation increases progressively from each end and reaches a maximum; this maximum then remains constant over a central portion of the cylinder of length $\delta Z$. Therefore, the deformed radial coordinate can be written as:

$$r = f(R, Z), \frac{\partial f}{\partial z} = 0 \forall Z \in \delta Z$$

For large deformations, we restrict attention to the diagonal terms of the matrix. Moreover, because the cables and fibers prevent axial extension, we take the axial deformation to be zero. The principal stretches are then:

$$\lambda_r = \frac{\partial f}{\partial R}, \lambda_\theta = \frac{r}{R}, \lambda_z = 1$$

Moreover, we assume an isochoric deformation, implying that volume is conserved throughout the deformation. For instance, the volume enclosed between the inner radius $R_{in}$ and $R$ in the reference configuration remains unchanged after deformation. This leads to the following relations:

$$R = \sqrt{r^2 - r_{in}^2 + R_{in}^2}$$

$$\lambda_\theta = \sqrt{\frac{r^2}{r^2 - r_{in}^2 + R_{in}^2}}$$

This allows us to express the relationship between $\Delta p$ and $\lambda_\theta$and in terms of strain energy function

$$\Delta p = \int_{\lambda_{ex}}^{\lambda_{in}} \left(\lambda_\theta{}^2 - 1\right)^{-1} \frac{\partial W}{\partial \lambda_\theta} d\lambda_\theta$$

where $\lambda_{in} = \frac{r_{in}}{R_{in}}$ and respectively for $\lambda_{ex}$.

We now solve the axial force equilibrium resulting from the gripper actuation

$$f_{ex} = f_{in} - P_{in} \pi r_{in}^2$$

Where $P_{in}$ is the internal pressure, and $f_{in}$ is the internal force generated at the cylinder cross-section; it can be expressed as a function of $\lambda_\theta$

$$f_{in} = 2\pi \int_{\lambda_{ex}}^{\lambda_{in}} \sigma_z \, |_{\lambda_z = 1} \, \frac{r^2}{\lambda_\theta (\lambda_\theta^2 - 1)} d\lambda_\theta$$

Here, $\sigma_z$ is one of the principal Cauchy stresses, expressed in terms of the strain energy function $W(\lambda_r, \lambda_\theta, \lambda_z)$

$$\sigma_i = -p_{in} + \lambda_i \frac{\partial W}{\partial \lambda_i}, \forall i = r, \theta, z$$

Applying Cauchy's first law of motion, the divergence of the stress tensor vanishes. Together with the volume-conservation assumption, this allows $\sigma_z$ to be expressed as:

$$\sigma_z \mid_{\lambda_z=1} = -p_{ex} + \left.\frac{\partial \widehat{W}}{\partial \lambda_z}\right|_{\lambda_z=1} - \int_{\lambda_{ex}}^{\lambda_\theta} \frac{\left.\frac{\partial \widehat{W}}{\partial \lambda_\theta}\right|_{\lambda_z=1}}{\lambda_\theta^2 - 1} \, d\lambda_\theta.$$

Then, substituting this into the expression for $f_{in}$ and applying integration by parts, we obtain the following expression for $f_{ex}$

$$f_{ex} \;\; = \pi R_{in}^2(\lambda_{in}^2 - 1) \int_{\lambda_{ex}}^{\lambda_{in}} \frac{2\left.\frac{\partial \widehat{W}}{\partial \lambda_z}\right|_{\lambda_z=1} - \lambda_\theta \left.\frac{\partial \widehat{W}}{\partial \lambda_\theta}\right|_{\lambda_z=1}}{\left(\lambda_\theta^2 - 1\right)^2} \lambda_\theta \, d\lambda_\theta \; - \; p_{ex} \pi \lambda_{ex}^2 R_{ex}^2.$$

For the reasons mentioned above, we choose the Yeoh strain energy density function

$$W = \sum_{i=1}^{N} C_i(\lambda_r^2 + \lambda_\theta^2 + \lambda_z^2 - 3)^i$$

where N is the polynomial order and $C_i$ are the material constants. For a third-order Yeoh model (N = 3), the analytical solutions for the pressure and axial force are computed in Maple as functions of $C_1$, $C_2$, and $C_3$. They are given, respectively, by:

$$\Delta p = \tilde{p}(\lambda_{in}, 1) - \tilde{p}(\lambda_{ex}, 1)$$

$$f_{ex} = \tilde{f}(\lambda_{in}, 1) - \tilde{f}(\lambda_{ex}, 1)$$

Where

$$\tilde{p}(\lambda_\theta, 1) = \frac{-C_1 + 2C_2 - 10.5C_3 - 6C_3\lambda_\theta^2 - C_3\lambda_\theta^{-2}}{\lambda_\theta^2}$$

$$+2C_2\lambda_\theta^2 - 15C_3\lambda_\theta^2 - 1.5C_3\lambda_\theta^4 + \frac{6C_3\lambda_\theta^4 + 15C_3 - 2C_2}{\lambda_\theta^2} + \ln \lambda_\theta \,(2C_1 - 4C_2 + 12C_3)$$

And

$$f(\lambda_\theta, 1) = \pi(\lambda_{in}^2 - 1)(-6C_1 \ln(\lambda_\theta) - 2C_2\lambda_\theta^2 + 8C_2 \ln(\lambda_\theta) + \frac{16C_2}{\lambda_\theta^2} - \frac{2C_2}{\lambda_\theta^4}$$
$$-\frac{3}{2}C_3\lambda_\theta^4 + 12C_3\lambda_\theta^2 + 24C_3 \ln^2(\lambda_\theta^2 - 1) - 60C_3 \ln(\lambda_\theta^2 - 1) - \frac{30C_3}{\lambda_\theta^2} + \frac{39C_3}{2\lambda_\theta^4})$$

**Suppl. 6 - Effect of balloon anchoring on magnetic tip actuation**

We characterized the tip motion of the mSCR with and without balloon anchoring. Balloon anchoring modifies the boundary conditions of the catheter and improves the transmission of magnetic actuation to the distal end. Under the small-deformation assumption, the bending response of the anchored segment can be approximated from beam theory, in which the local curvature is given by

$$\theta = \int_0^L \frac{\tau(x)}{\sum_i E_i I_i} dx$$

where $E_i$ and $I_i$ are the Young's modulus and second moment of area of the $i$-th constituent, respectively: (1) silicone + NdFeB, (2) glass fiber, (3) LED fiber, (4) camera cable, and (5) nylon wire. Based on this, we derived four trajectories and expressed them as parametric equations.

$$\rho_{circle}(\theta) \; = \; r \cdot \sin(\theta), \theta \in [0, \pi]$$

$$\rho_{spiral}(\theta) \; = \; r \cdot 0.1\theta, \theta \in [0, 5\pi]$$

$$\rho_{square}(\theta) \; = \; r \cdot \frac{2}{|\cos(\theta) - \sin(\theta)| + |\cos(\theta) + \sin(\theta)|}, \theta \in [0, 2\pi]$$

$$\rho_{butterfly}(\theta) = r \cdot \frac{(2 - \sin(\theta) + \sin(3\theta) + 2\sin(5\theta) - \sin(7\theta) + 3\sin(2\theta) - 2\sin(4\theta))}{6}, \theta \in [0, 2\pi]$$

To quantify the anchoring effect experimentally of the present mSCR, we applied controlled magnetic fields using the MiniMag (MFG 100, MagnebotiX) [101] and tracked the resulting tip motion in both the deflated- and inflated-balloon configurations. For ease of video tracking, the integrated LED fiber was used at a visible wavelength of 450 nm, and the tip position was continuously recorded with a camera (Basler, Germany). The experiment was repeated in four representative actuation scenarios (Fig. S18), and the root-mean-square error (RMSE) with respect to the prescribed trajectory was calculated in both conditions (Table S5). The planned path was registered to the workspace from the commanded trajectory of the navigation system, without fitting to either run, and the same registration was used for both runs; the RMSE was computed from the distance of each tracked tip position to the nearest point of the planned path. One run was performed per condition and trajectory; the same data are shown in Fig. 3F of the main text. Because we did not have access to a tissue-mimicking pancreatic-duct model suitable for anchoring, the balloon was deployed inside a synthetic coronary artery segment (4 mm inner diameter; SynTissue, SynDaverX, USA), which provided a repeatable constrained geometry for benchtop evaluation.

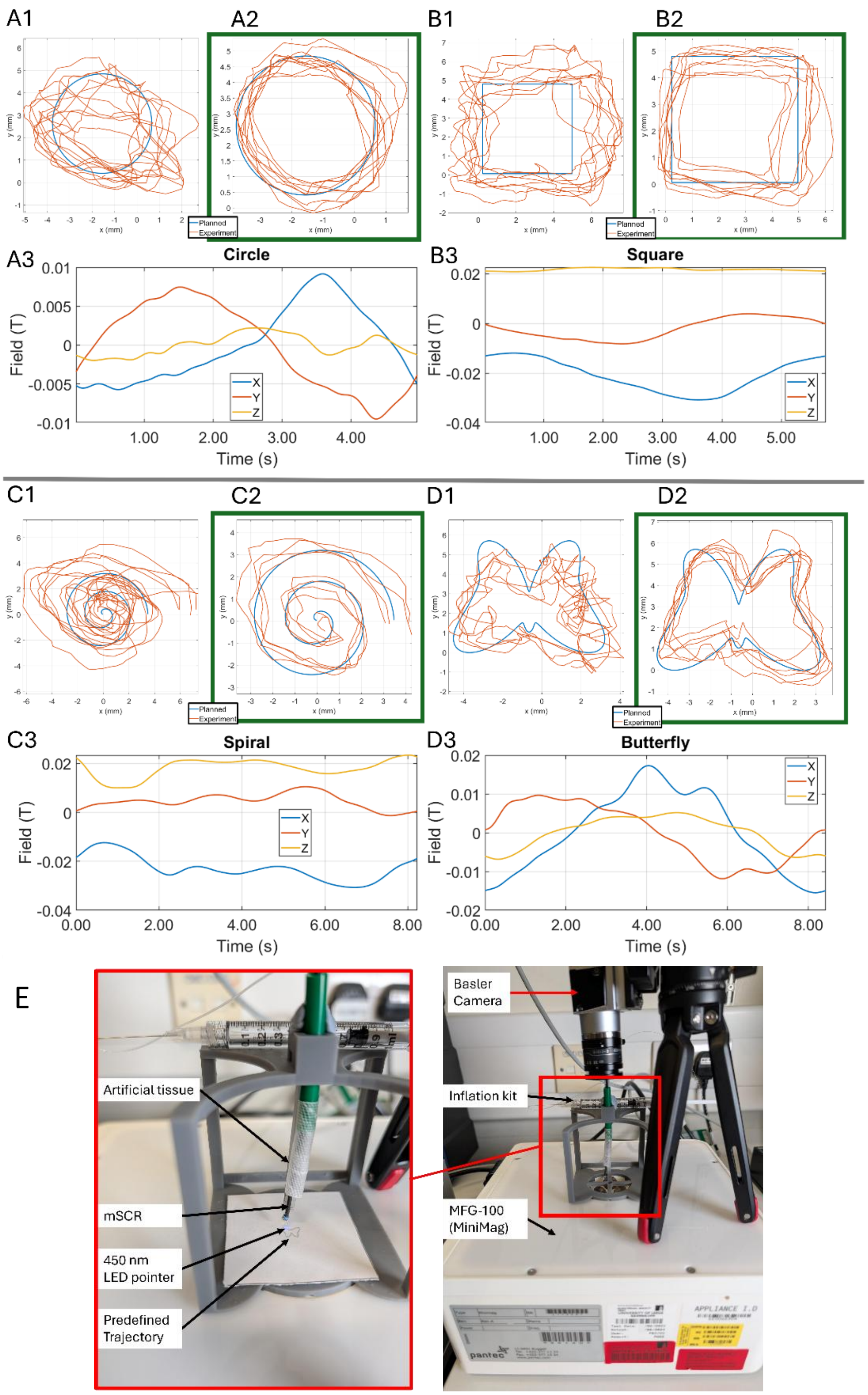

*Figure S18 Effect of the anchoring balloon on tip dexterity during the execution of different trajectories: (A) circle, (B) square, (C) spiral, and (D) butterfly. For each trajectory, _1 indicates the experiment without the balloon, _2 the experiment with the balloon, and _3 the applied magnetic field generated by the MiniMag system. (E) Experimental setup.*

| Trajectory | Without anchoring | With anchoring |
|---|---|---|
| Circle | 0.78 mm | 0.37 mm |
| Square | 1.16 mm | 0.41 mm |
| Spiral | 1.16 mm | 0.39 mm |
| Butterfly | 0.60 mm | 0.45 mm |

*Table S5 Tip-tracking RMSE between the planned trajectory and the tracked tip positions, without and with balloon anchoring (one run per condition and trajectory).*

## Suppl. 7 - Effect of autoclave sterilization on the mSCR functional performance

To evaluate the effect of sterilization on mSCR performance, we conducted additional experiments assessing gripper lifetime and actuation force, balloon deformation and burst pressure, and the magnetic responsiveness of the mSCR (Fig. S19).

For gripper performance, we fabricated 20 grippers and compared 10 sterilized devices with 10 non-sterilized controls using the same setup described in the “Compliant gripper” section (Fig. S9). Grippers in both groups remained functional for more than 250 cycles (273 ± 13 cycles before sterilization versus 261 ± 10 cycles after sterilization), corresponding to less than 5% difference in lifetime. No statistically significant difference in actuation force was observed between groups (0.477 ± 0.1 N before sterilization versus 0.475 ± 0.1 N after sterilization).

For balloon performance, we fabricated 40 balloons and divided them into four groups of 10 to evaluate deformation over repeated actuation and burst pressure before and after sterilization, using the setup described in Fig. S8. After sterilization, the balloons showed approximately 10% lower radial deformation over 300 actuation cycles (3.1 ± 0.2 mm before sterilization versus 2.8 ± 0.1 mm after sterilization), while burst pressure remained essentially unchanged (64 ± 11 kPa before sterilization versus 64 ± 13 kPa after sterilization).

For magnetic performance, we fabricated 20 magnetic soft robotic catheters, magnetized them, and quantified their deformation under applied magnetic fields of 10, 15, and 20 mT before and after sterilization. Tip angles of catheters with housing and gripper (configuration 5 of Table 1; 60-mm free length, single-axis coil) changed by less than 7% after autoclaving (71.5 ± 1.9° versus 76.3 ± 2.1° at 10 mT, 79.1 ± 6.7° versus 81.1 ± 6.7° at 15 mT, and 83.0 ± 1.5° versus 86.1 ± 1.1° at 20 mT).

Regarding the drug delivery module, the drug payload was loaded, and the release shell bonded, immediately after sterilization and just before the procedure; the shell itself was therefore not autoclaved. The release profile and trigger threshold of the shell were thus not affected by autoclaving.

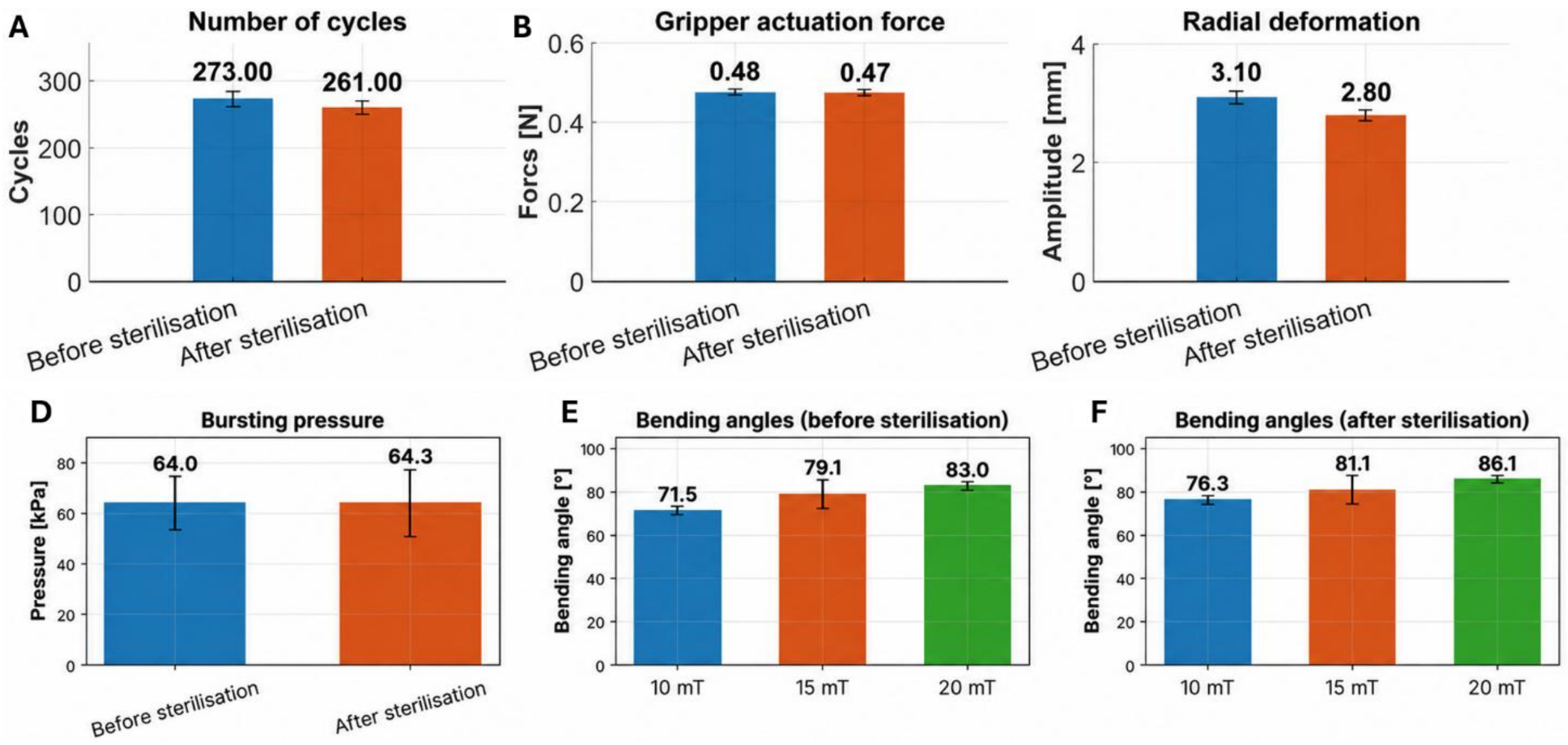

*Figure S19* ***Effect of sterilization on the performance of the fabricated components, including the gripper, balloon, and magnetic actuation. (A)*** *Gripper lifespan.* ***(B)*** *Gripper actuation force.* ***(C)*** *Balloon radial deformation.* ***(D)*** *Balloon burst pressure.* ***(E, F)*** *Bending angle under three different applied magnetic field strengths.*

## Suppl. 8 - Thermo-acoustic safety characterization of ultrasound-triggered drug release

We evaluated the thermal effect associated with ultrasound stimulation of the shell during drug release. To do so, we fabricated an anatomically realistic tissue-mimicking phantom based on polyvinyl alcohol gel, chosen for its tissue-like acoustic properties, including speed of sound and acoustic attenuation. The phantom was placed in a temperature-controlled bath of deionized water, and a thermal camera (thermoIMAGER TIM M-1, Micro-Epsilon) was positioned above the setup and calibrated before the experiment. The catheter was then deployed into the pancreatic-duct-mimicking channel of the phantom, and the system was allowed to reach thermal equilibrium for 2 h before ultrasound-triggered release was initiated.

To improve thermal measurement quality, infrared images were acquired as snapshots at 4 Hz rather than as continuous video. Each snapshot was exported as a CSV file and processed in MATLAB (MathWorks). The experiment consisted of initiating thermal acquisition, triggering ultrasound-mediated drug release, and monitoring the temperature change at the release location (Fig. S20). A transient local temperature increase of approximately 0.8 °C was observed at the release site, with the temperature returning to baseline within 12 s after ultrasound stimulation ceased.

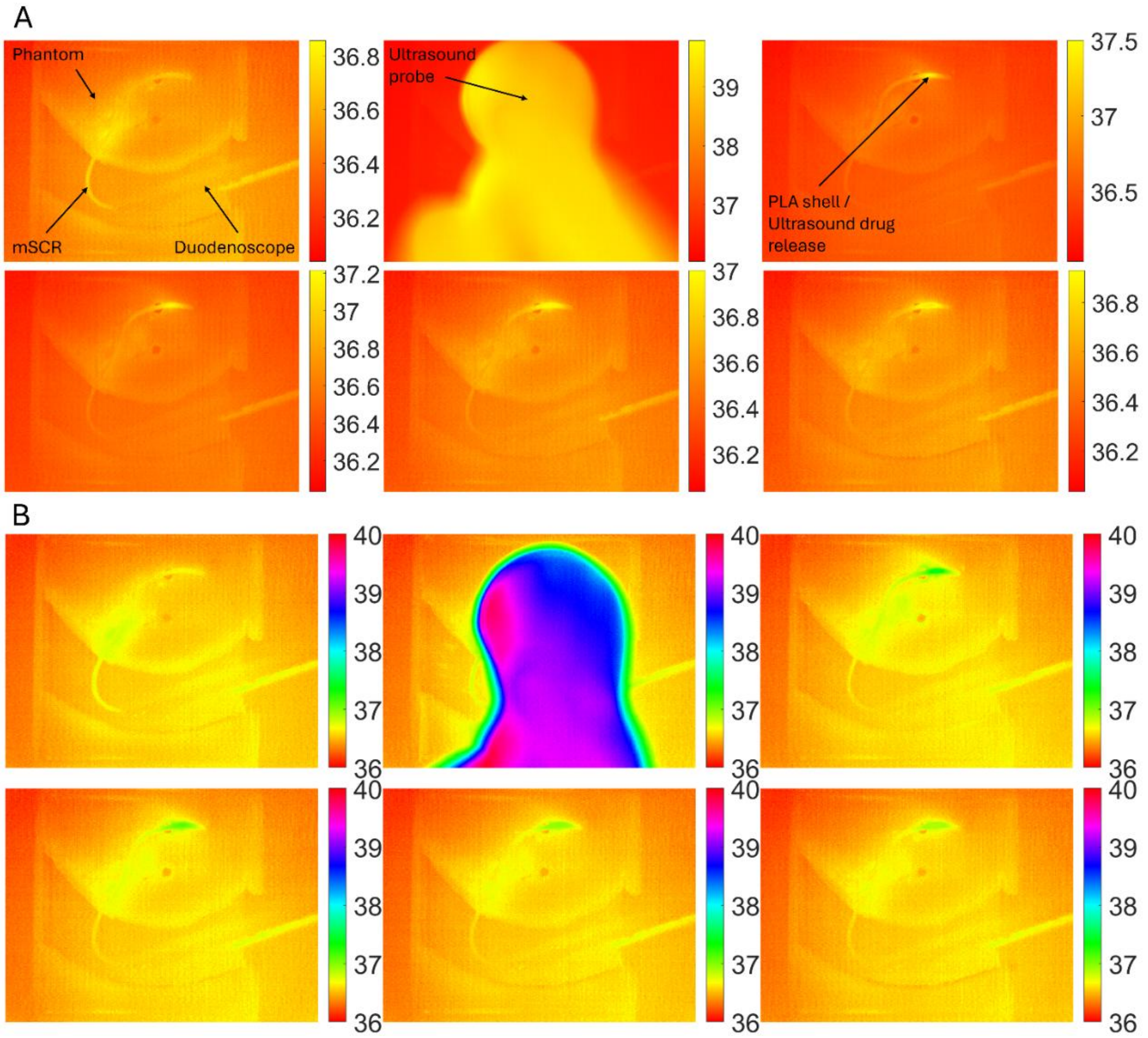

*Figure S20* ***Evaluation of the thermoacoustic effect induced by ultrasound-triggered drug release.*** *Identical snapshots are displayed using an adaptive color scale [unit: degree Celsius] (A) and a fixed color scale (B) to emphasize local temperature changes.*

## Suppl. 9 - Compatibility of the ultrasound-triggered release platform with a clinically relevant therapeutic payload

To evaluate the compatibility of the ultrasound-triggered release platform with a clinically relevant therapeutic, we performed additional experiments using paclitaxel–albumin (PTX-HSA), a drug formulation used in pancreatic cancer treatment. Because the aim of this study was to assess loading and triggered release compatibility rather than therapeutic efficacy, expired PTX-HSA that was no longer suitable for clinical use was used for these experiments. After integration into the coated shell, the samples were stored at 4°C and used within 24 h.

To assess premature release and ultrasound-triggered release, the drug-loaded shell was deployed in the previously described tissue-mimicking phantom submerged in deionized water, which served as the surrounding collection medium. A first sample was collected immediately after setup and used as the baseline reference. The system was then left undisturbed during thermal stabilization, with the shell maintained inside the phantom. A second sample was collected 2 h later, before ultrasound triggering, to evaluate possible early release. Finally, ultrasound stimulation was applied to trigger release, and a third sample was collected immediately afterward to assess payload release.

Because PTX-HSA does not provide a distinct visual signal in the surrounding medium, the collected samples were analyzed by proton nuclear magnetic resonance spectroscopy ($^{1}$H NMR;

AV3HD-400, Bruker). For analysis, each 2 mL sample was diluted 1:1 with deuterium oxide. As expected, all spectra showed a dominant water peak. Signals associated with PLA were detected in samples collected before and after stimulation, with only trace levels observed before triggering and an approximately sevenfold greater signal after ultrasound activation, consistent with limited background polymer release and substantially increased shell disruption after stimulation. Importantly, PTX-HSA-related signals were detected only in the post-stimulation sample, confirming on-demand release of a clinically relevant therapeutic payload (Fig. S21).

These results demonstrate that incorporation of PTX-HSA does not prevent shell loading or ultrasound-triggered release under the conditions tested. However, further studies will be required to evaluate long-term payload stability, tissue uptake, and therapeutic efficacy after release.

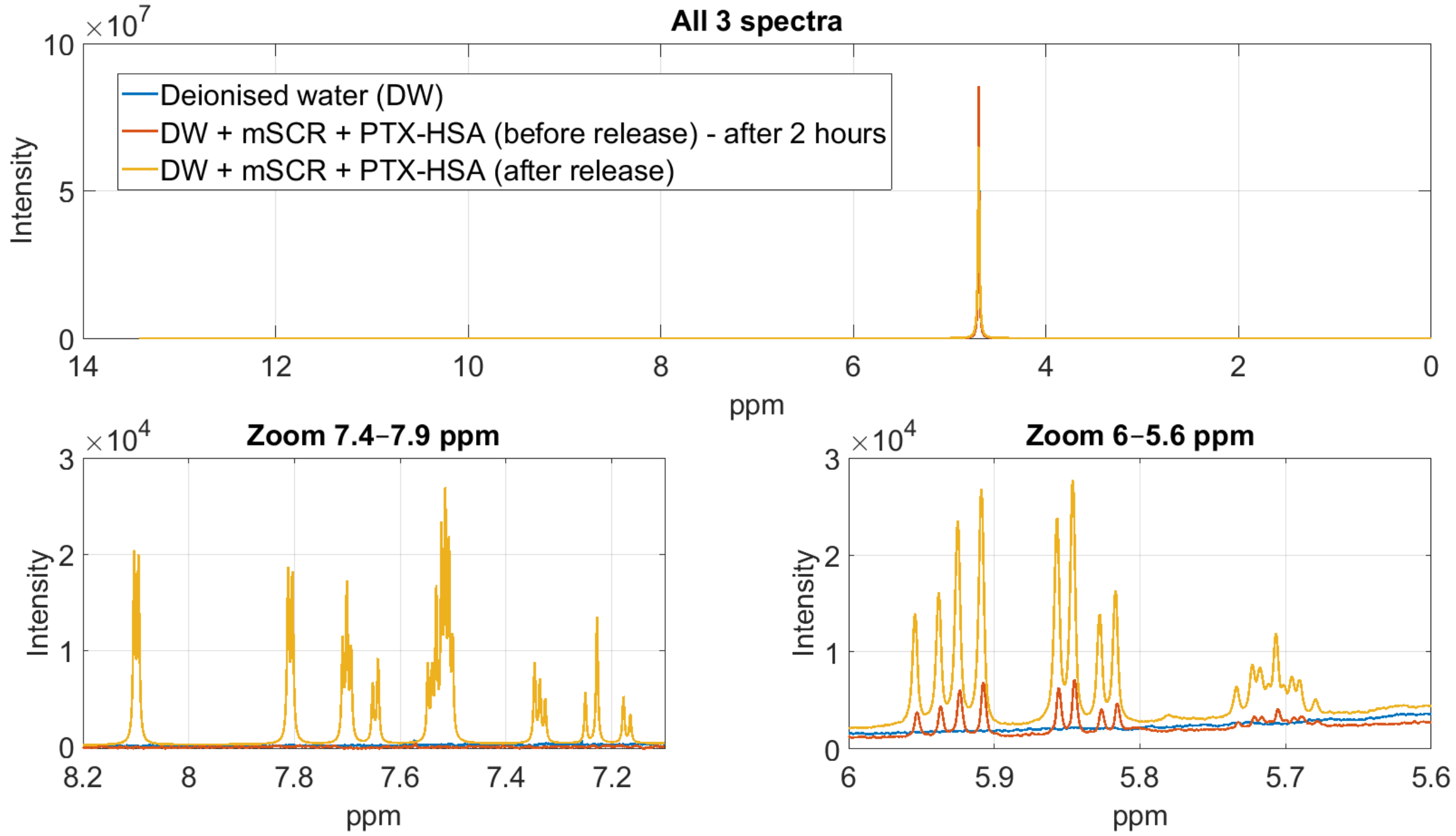

*Figure S21* ***Evaluation of clinically relevant ultrasound-triggered drug release.*** *Proton NMR spectra are shown for three samples: a reference, before drug release, and after drug release.*

**Suppl. 10 - Biocompatibility analysis**

To assess the biological response to the DS 30 and DS 30+NdFeB microparticles, both samples were first autoclaved at 121 °C for 30 min followed by 30 min of drying, as for the devices. Under sterile conditions, they were then subjected to an extract preparation process to prepare conditioned media for cell culture. The conditioned media were prepared by incubating 10 mm segments of autoclaved DS 30 or DS 30+NdFeB microparticles in 100 mL of Dulbecco's Modified Eagle Medium (DMEM; Thermo Fisher Scientific) supplemented with 10% fetal bovine serum (FBS; Sigma-Aldrich), 1% Penicillin-Streptomycin (P/S; Sigma-Aldrich), and 1% GlutaMAX (Thermo Fisher Scientific) in 75 $cm^2$ culture flasks at 37 °C for 48 hours with oscillation. As the positive (cytotoxic) control, NdFeB microparticles were diluted 1:10 in DMEM/10% FBS medium and incubated under identical conditions. Untreated cells served as the reference.

Human fetal foreskin fibroblast 2 (HFFF2) cells, provided by Dr Nicola Ingram (Molecular and Nanoscale Physics Research Group, School of Physics and Astronomy), were cultured in DMEM/10% FBS medium. Cells were seeded at a density of $5 \times 10^4$ cells per well in 24-well plates and maintained at 37 °C in a humidified (95-99%) incubator with 5% $CO_2$ until confluent.

Once confluent and with conditioned media prepared, the culture medium in each well was aspirated and replaced with 500 µL of DS 30 conditioned medium, DMEM/10% FBS culture medium with 10 mm autoclaved DS 30, DS 30+NdFeB conditioned medium, DMEM/10% FBS culture medium with 10 mm autoclaved DS 30+NdFeB microparticles, and DMEM/10% FBS culture medium with the prepared NdFeB particles at a dilution of 1:10. Cells were incubated for 24 h under standard cell culture conditions (37 °C, 5% $CO_2$, 95–99% humidity), and videos were acquired with IncuCyte® S3 Live-Cell Analysis System to observe cellular behavior under the different conditions.

Cell viability was quantified using the CellTiter-Glo® 2D Cell Viability Assay (Promega), which measures intracellular ATP as an indicator of metabolically active cells. A 1:1 volume of CellTiter-Glo reagent was added to each well, and the plates were placed on an orbital shaker for 2 min to induce cell lysis. Lysates were then transferred to opaque 96-well plates and incubated at room temperature for 10 min to stabilize the luminescent signal. Luminescence was measured using a Tecan Spark Microplate Reader (Low End, MIC9412). Viability relative to untreated cells was 86–97% for the silicone and composite conditions and 92% for the NdFeB particles of the positive control (Table S6, Fig. S22).

*Table S6 Cell viability of HFFF2 fibroblasts after 24 h with extracts of, or direct contact with, DragonSkin 30 (DS 30) and the DS 30–NdFeB composite, and with NdFeB particles alone (positive control), measured as intracellular ATP (CellTiter-Glo). Viability is given relative to untreated cells; SE, standard error.*

| **Condition** | **ATP luminescence (RLU)** | **SE** | **Viability (% of untreated)** |
|---|---|---|---|
| Untreated cells (reference) | 1365683 | 100749 | 100 |
| Cells + DS 30 extract | 1311364 | 100373 | 96.0 |
| Cells + DS 30, 1 cm segment | 1183931 | 115584 | 86.7 |
| Cells + DS 30–NdFeB extract | 1175753 | 52447 | 86.1 |
| Cells + DS 30–NdFeB, 1 cm segment | 1321327 | 62805 | 96.8 |
| Cells + NdFeB particles (1:10; positive control) | 1257250 | 67711 | 92.1 |

*Figure S22 ATP-based metabolic activity of HFFF2 fibroblasts after 24 h under each condition of Table S6 (mean ± SE).*

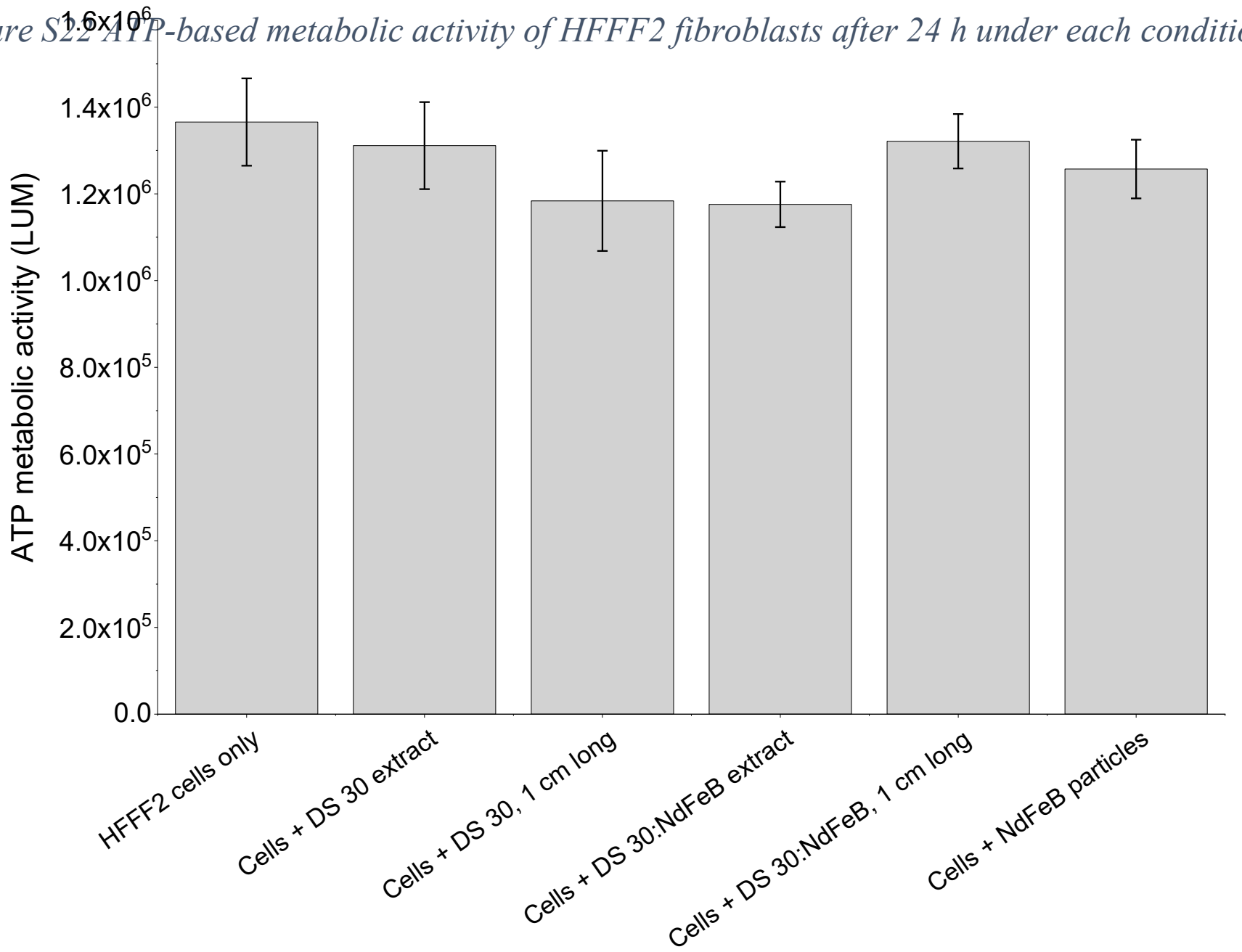

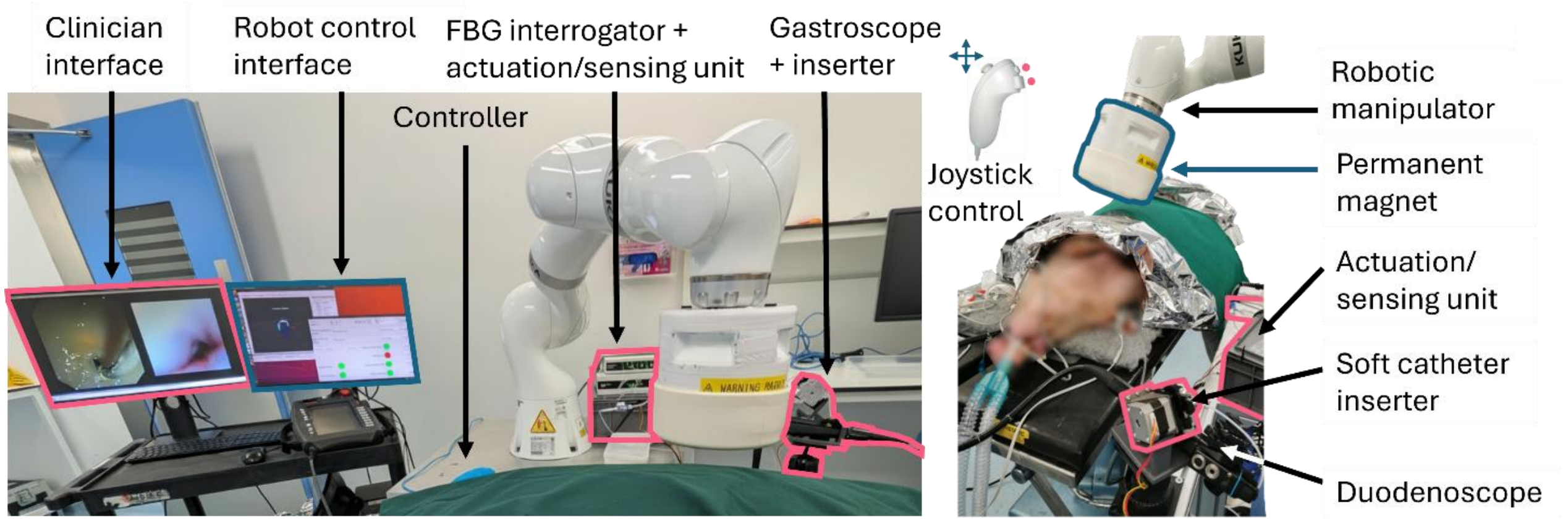

*Figure S1: In vivo experimental setup. (Left) Control station: clinician interface showing the endoscope view alongside the mSCR camera view, robot control interface, controller, FBG interrogator with the actuation and sensing unit, and gastroscope with inserter. (Right) Robotic manipulator and permanent magnet beside the pig, with the actuation and sensing unit, soft catheter inserter, and duodenoscope, and the joystick used to control the robotic manipulator and the mSCR.*

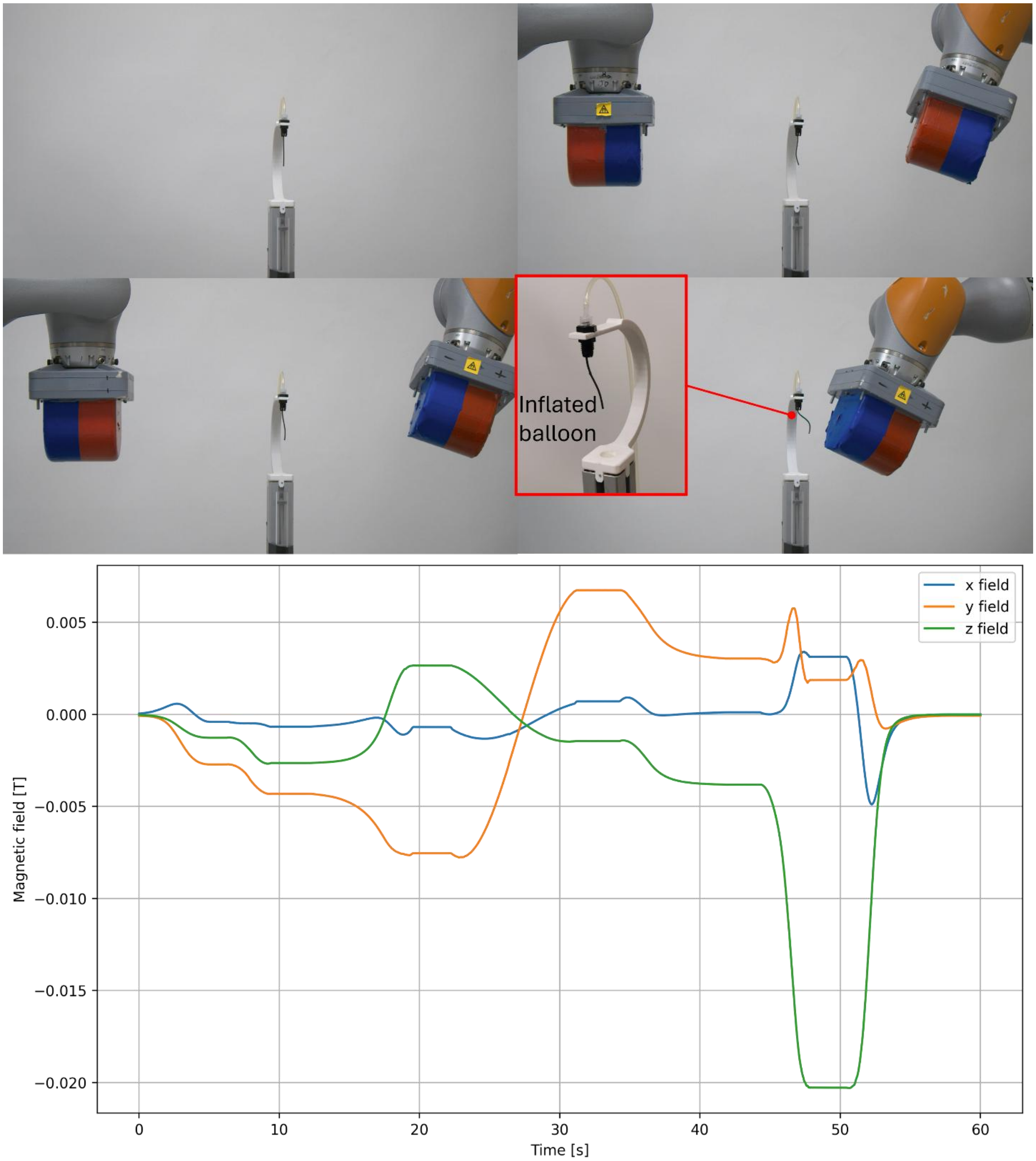

*Figure S2 Demonstration of the conservation of the ability to perform shape forming using the dual permanent magnet platform [31], with the magnetic balloon integrated.*

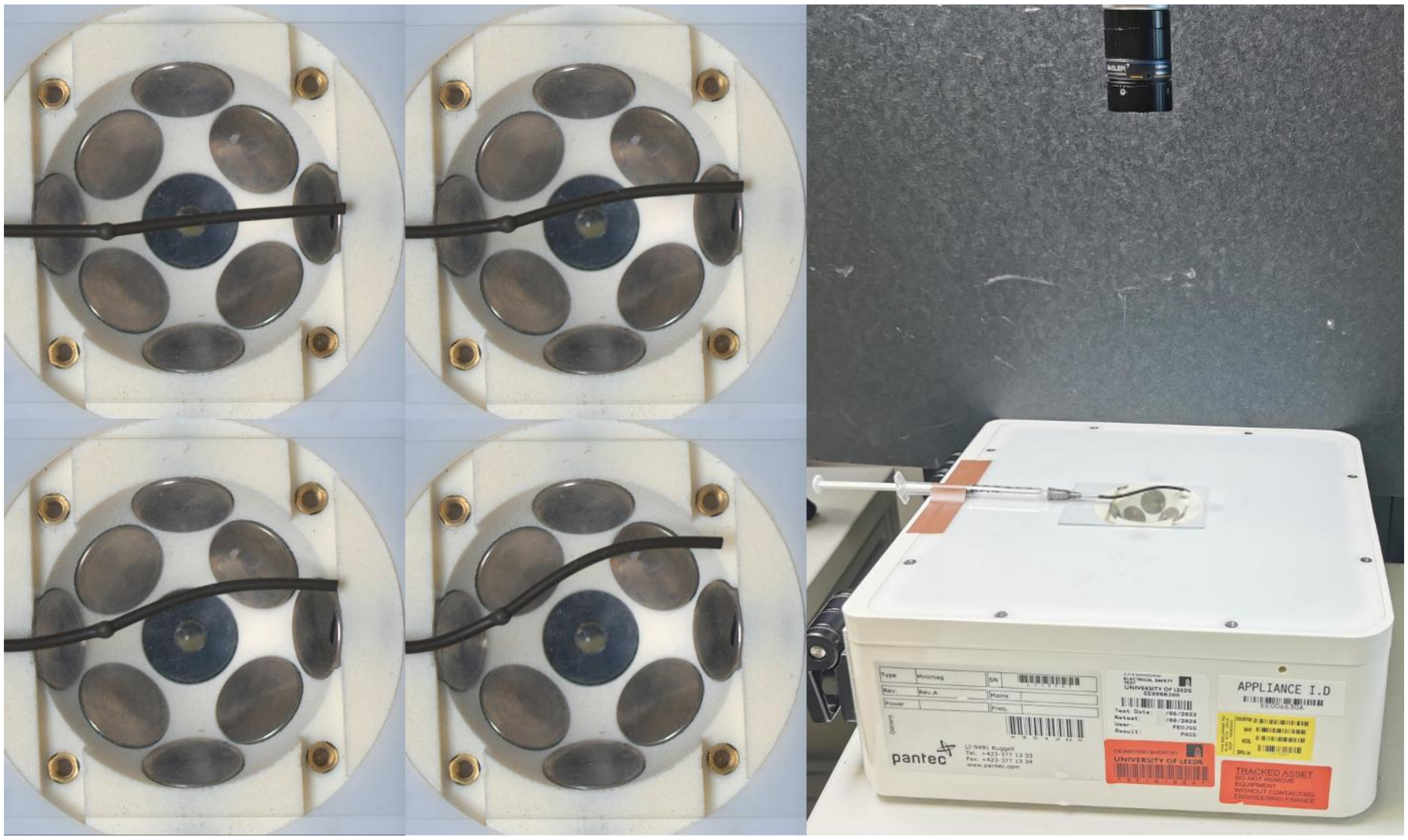

*Figure S3 Demonstration of the conservation of the ability to perform shape forming using electromagnet platform (MiniMag, MagnebotiX [101]), with the magnetic balloon integrated.*

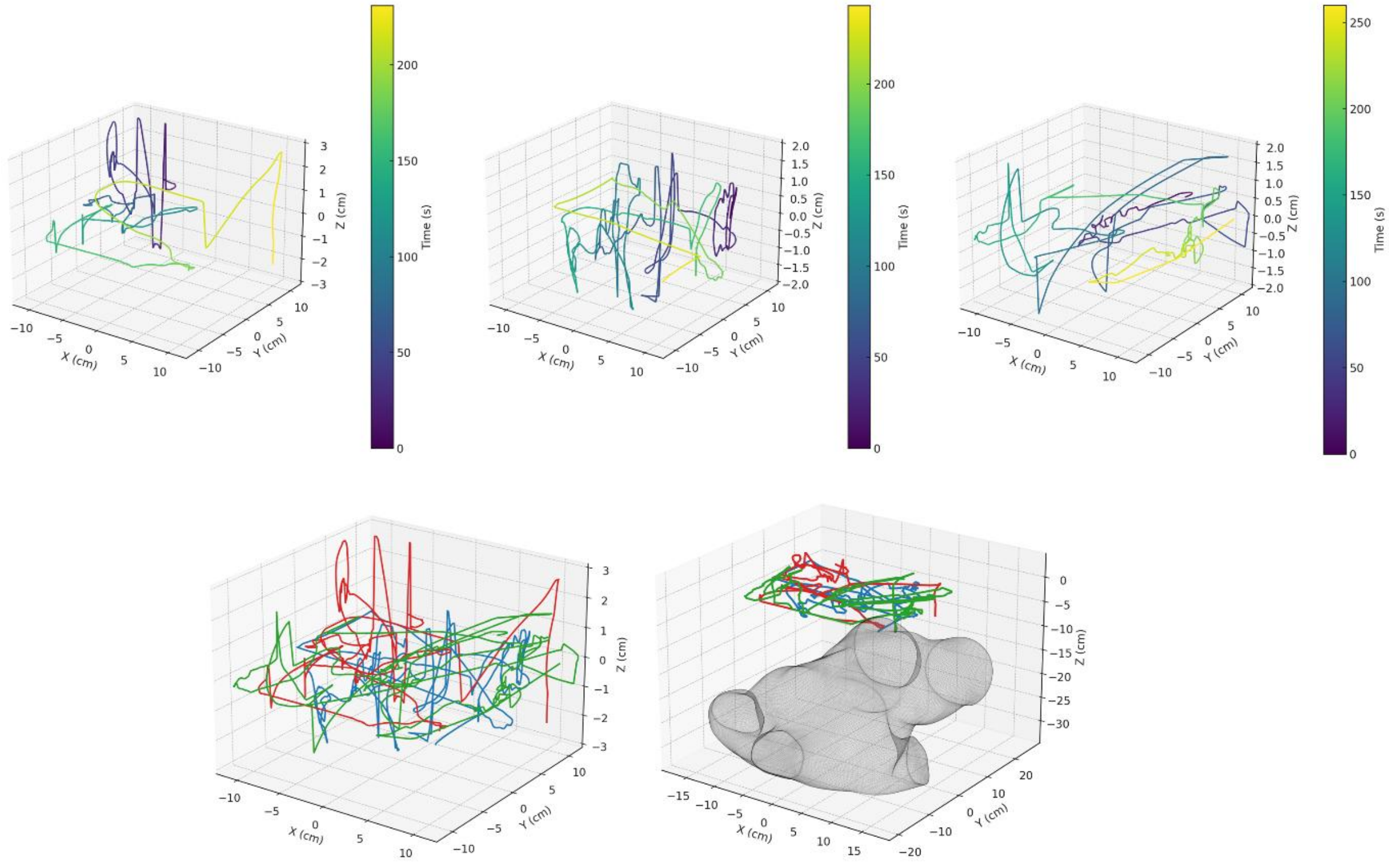

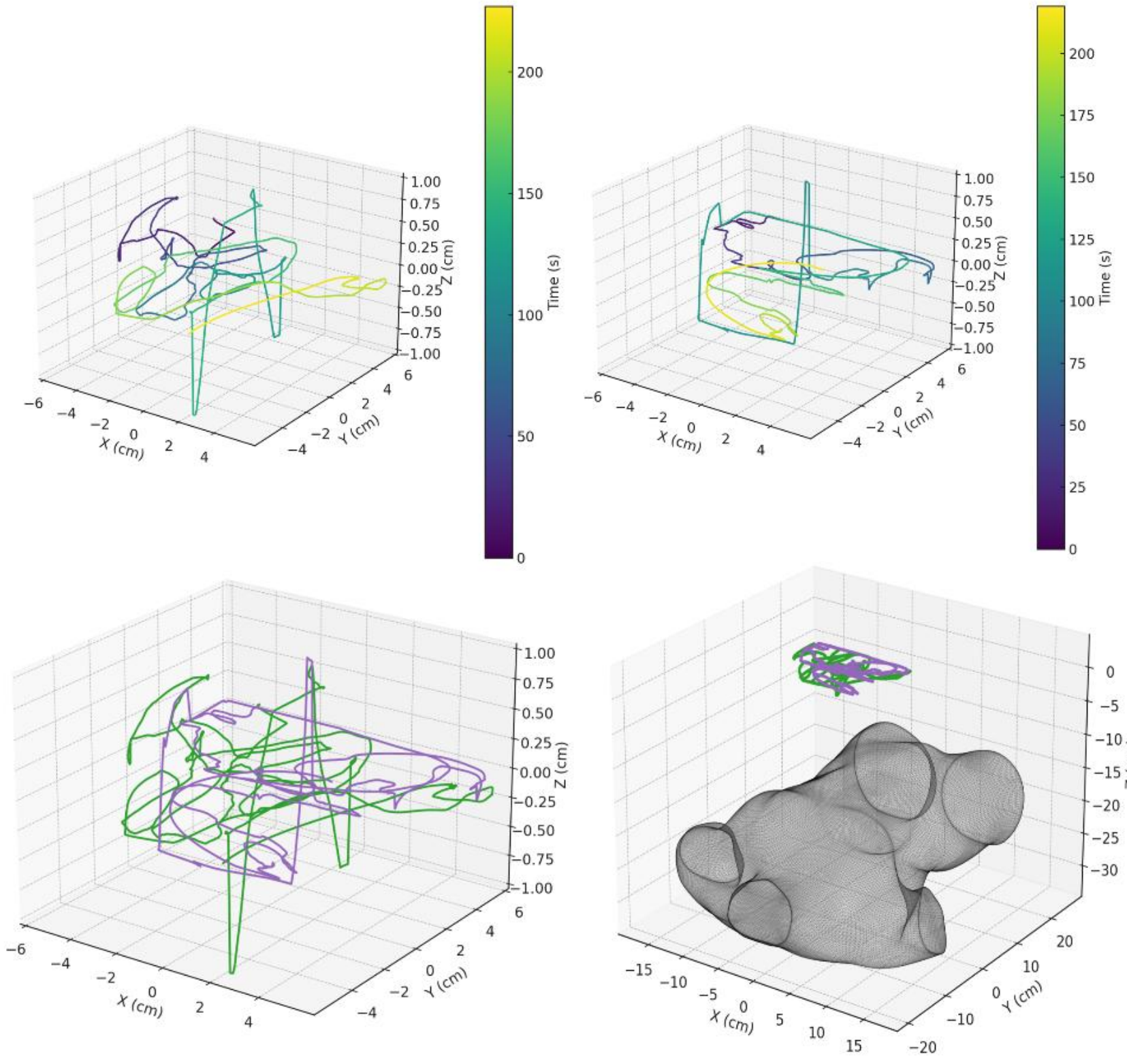

*Figure S4 Details of the EPM trajectory illustrated in Fig. 4.*

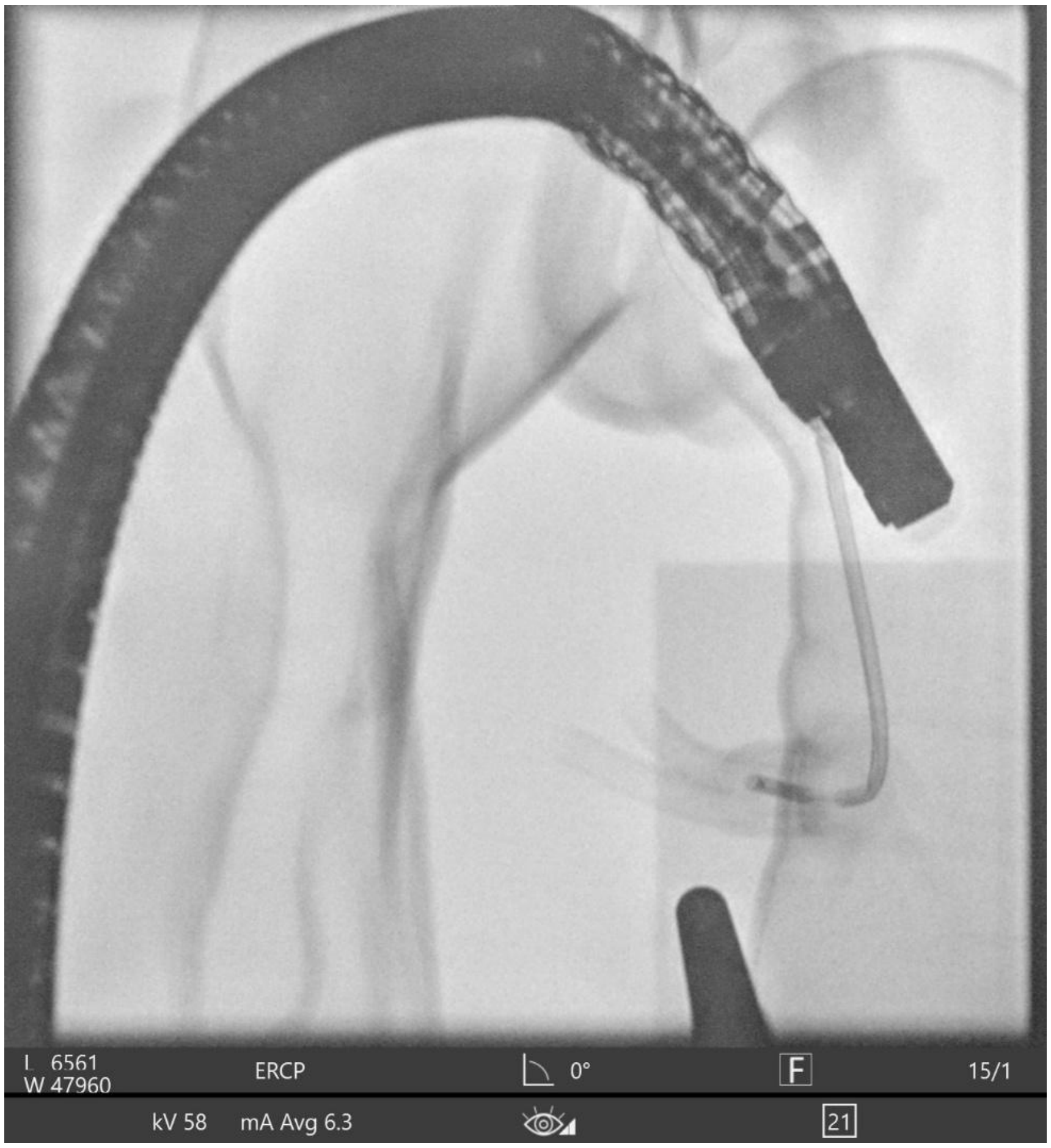

*Figure S5 mSCR navigation visualized under C-arm fluoroscopic imaging (Omniscope Dream, Stefanix).*

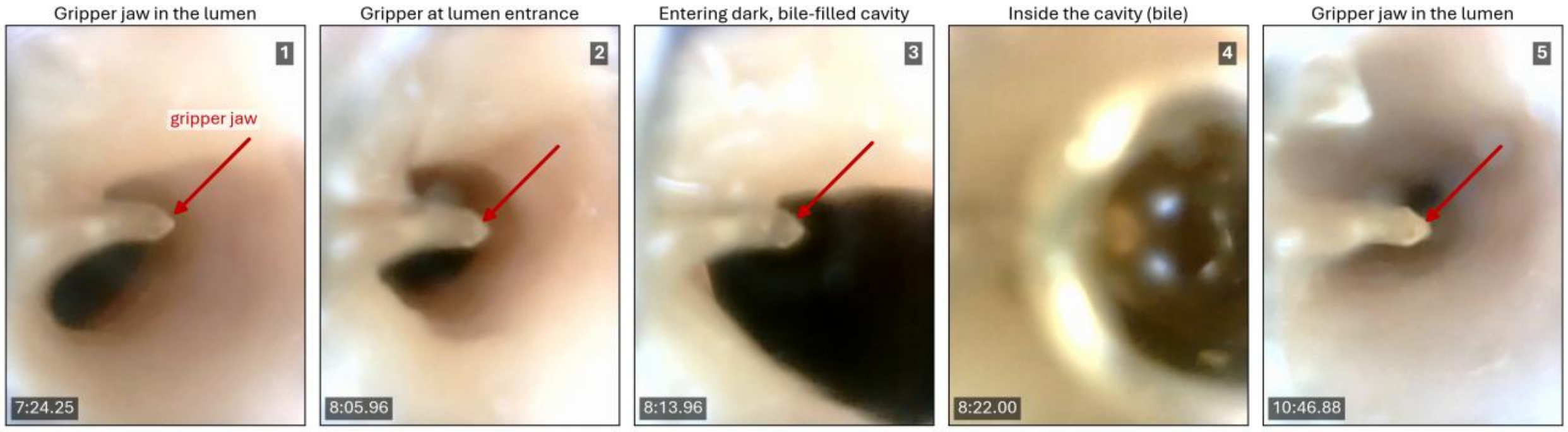

***Figure S6*** *Transpapillary access to the gallbladder with the fully functionalized mSCR, including the gripper (first access; onboard camera; times in min:s). (1, 2) Gripper jaw (arrow) in the duct lumen. (3) Entry into a dark, bile-filled cavity. (4) Inside the cavity. (5) Gripper jaw in the duct lumen at a later time point. The gripper was omitted in the subsequent accesses (Fig. 5, B and C) to widen the camera field of view.*

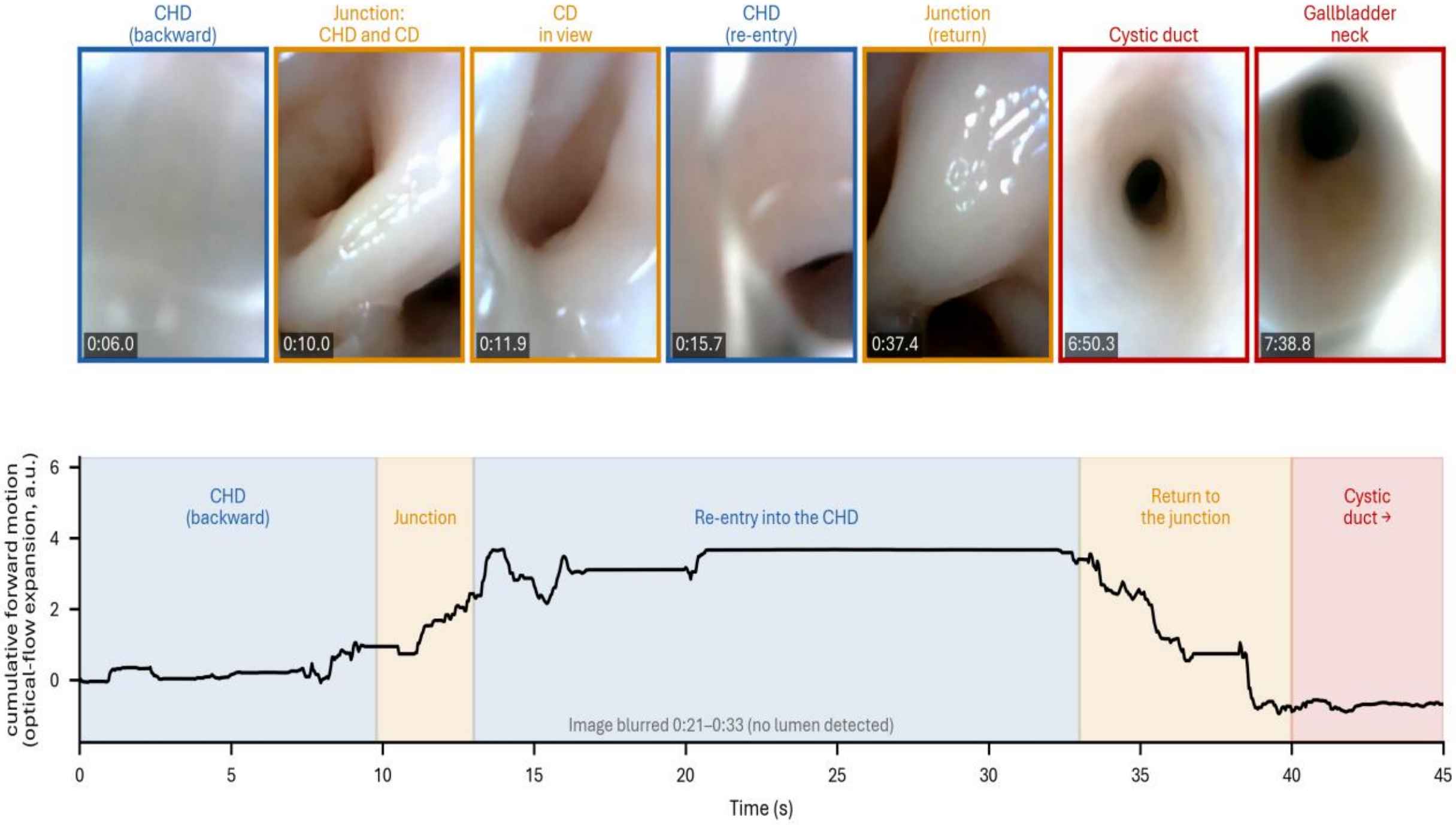

***Figure S7*** *Branch selection at the junction of the common hepatic duct (CHD) and the cystic duct (CD), which join to form the common bile duct (same video as Fig. 5C). Top: onboard frames of each step, with frame colors indicating the CHD (blue), the junction (orange), and the CD (red). Bottom: cumulative forward motion estimated from the expansion of the optical flow (arbitrary units) over the first 45 s, showing the re-entry into the CHD (0:12.5 to 0:14), the pause with a blurred image (0:21 to 0:33), and the return to the junction (0:33 to 0:40) before the catheter entered the CD.*

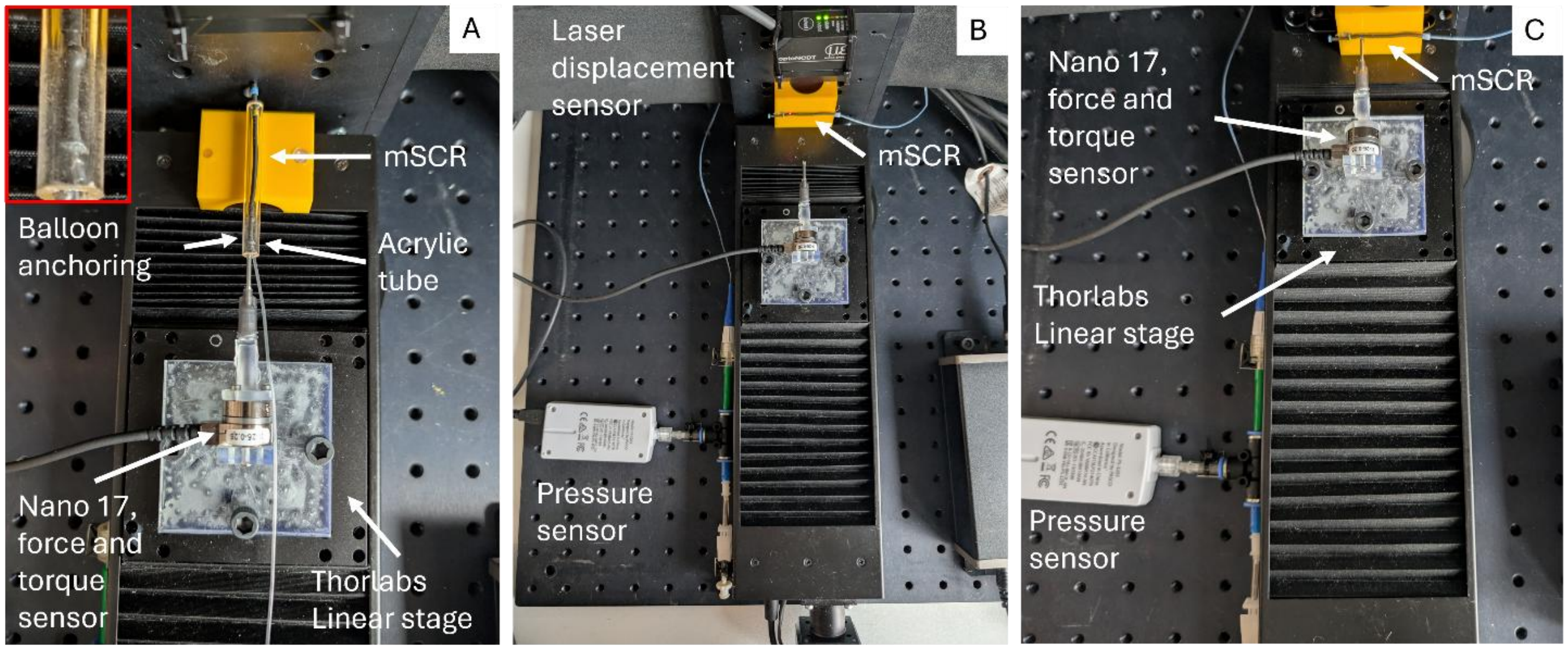

*Figure S8 Experimental setups used for anchoring force characterization (A), deformation quantification (B), and measurement of the inflation pressure while the balloon presses on the force sensor (C).*

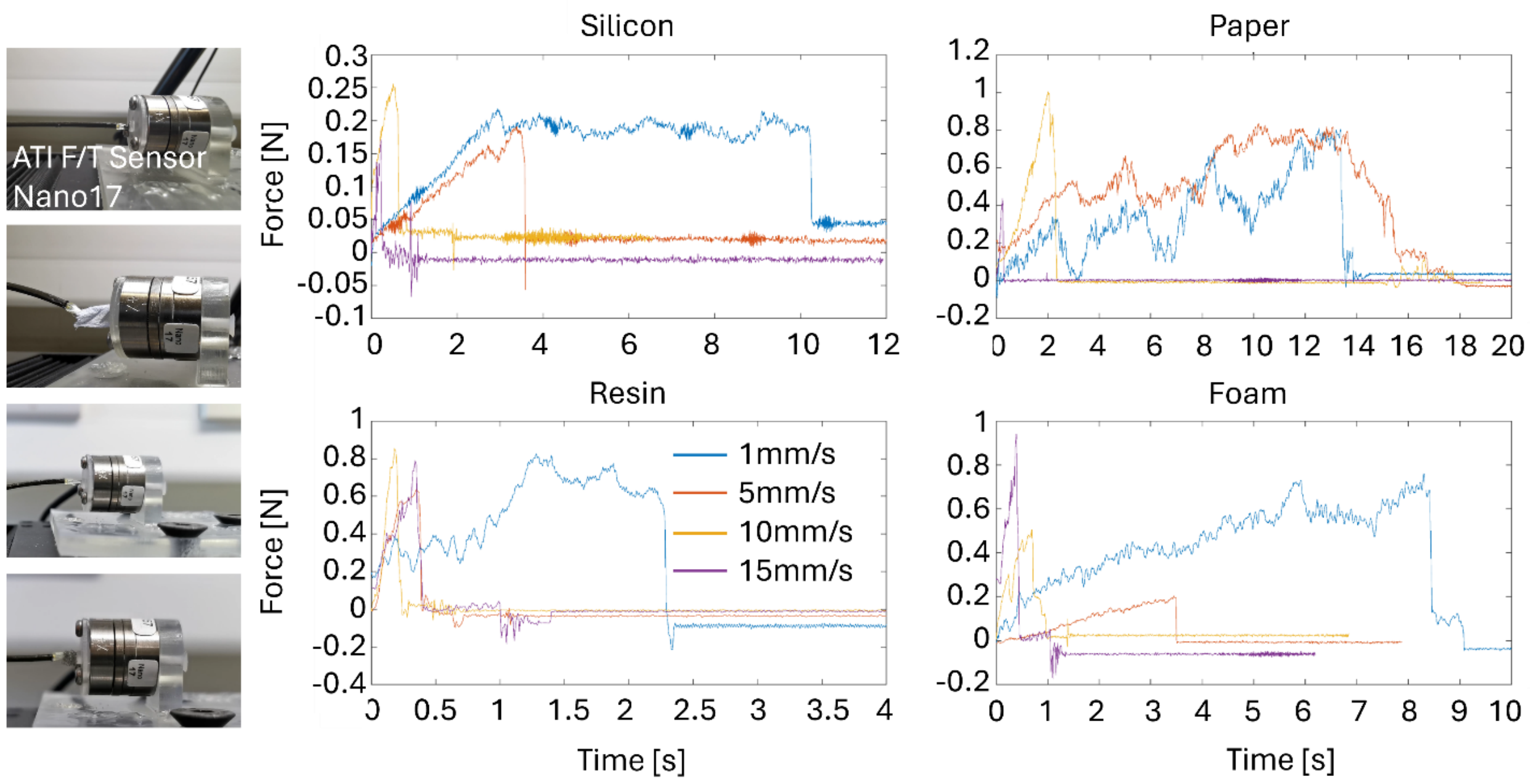

*Figure S9 Validation of gripping performance on materials spanning different stiffness properties and tested at multiple movement speeds. The mSCR is clamped at its proximal end, and the gripper is closed around the object by tensioning the nylon wire. The object is connected to a load cell (F/T Sensor Nano17 Titanium, ATI, MA, USA) fixed to a linear translation stage (NRT150, Thorlabs, NJ, USA). The platform is moved at different speeds to simulate dynamic manipulation.*

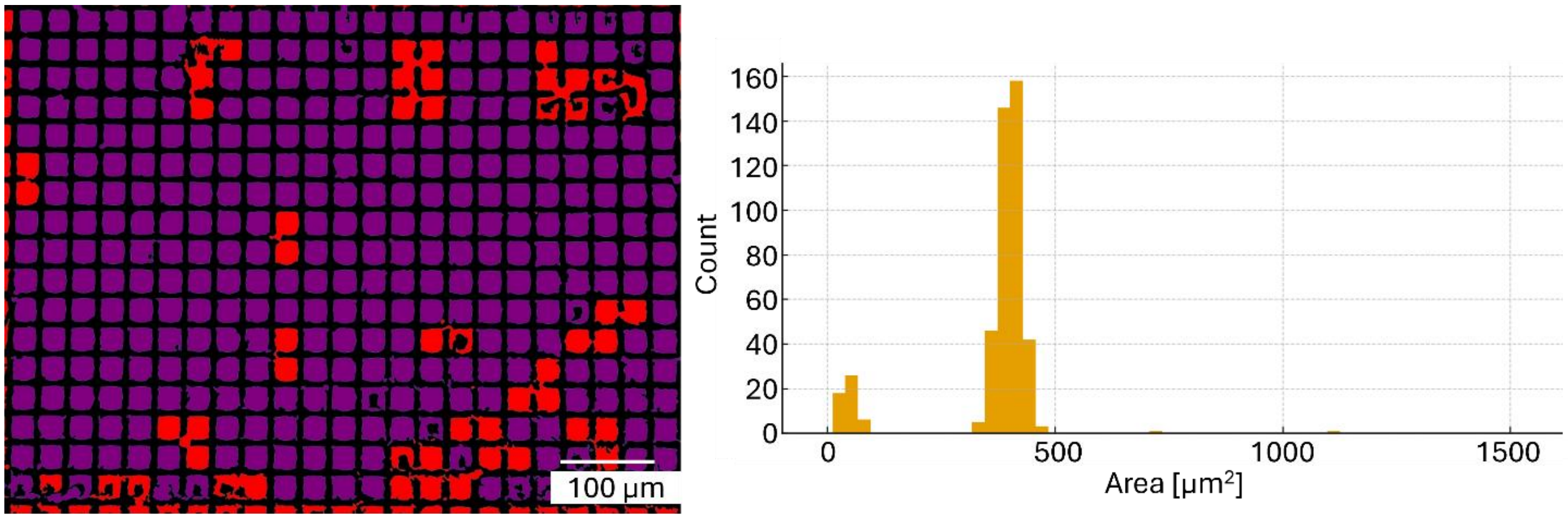

*Figure S10 Microscopic analysis of the molded PLA shell to validate the size of the pattern.*

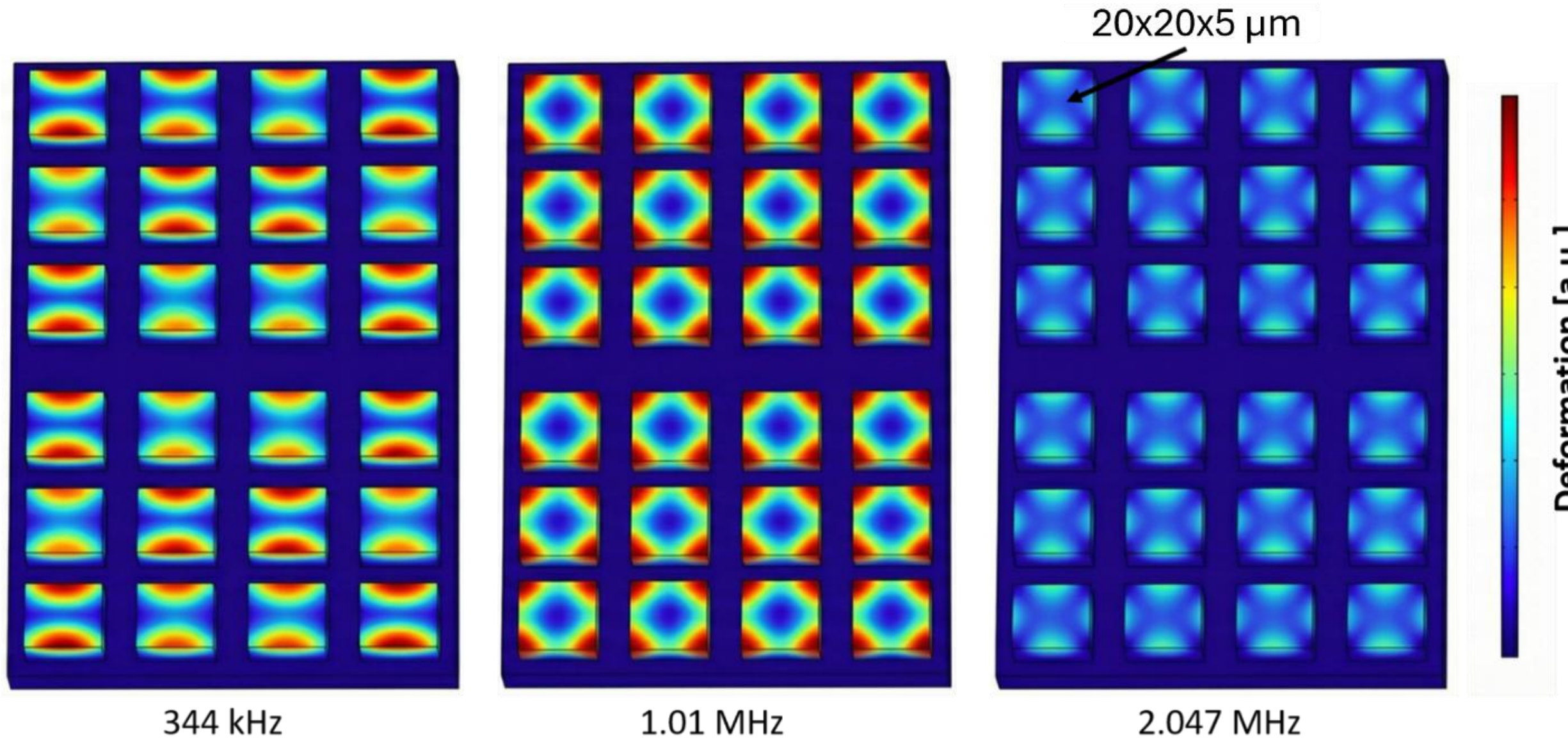

*Figure S11 Finite Element Analysis results illustrating how the PLA shell pattern deforms as a function of the applied ultrasound frequency.*

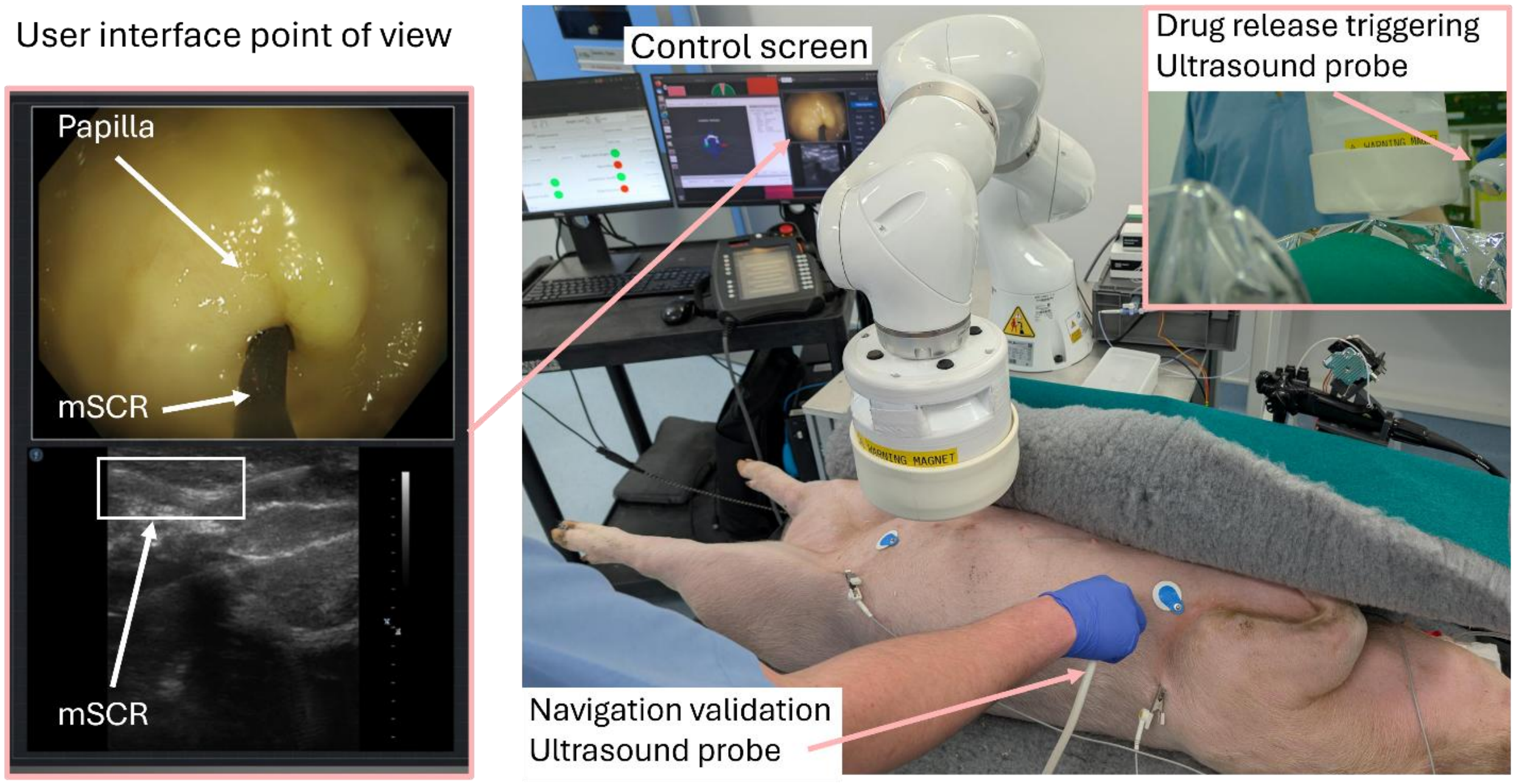

*Figure S12:* ***In vivo experimental setup for drug release*** *(Left) In vivo endoscopic view alongside ultrasound feedback displayed on the user interface. (Right) Verification of the mSCR position within the pancreatic duct, followed by activation of the drug-release mechanism using the secondary probe (top right detail).*

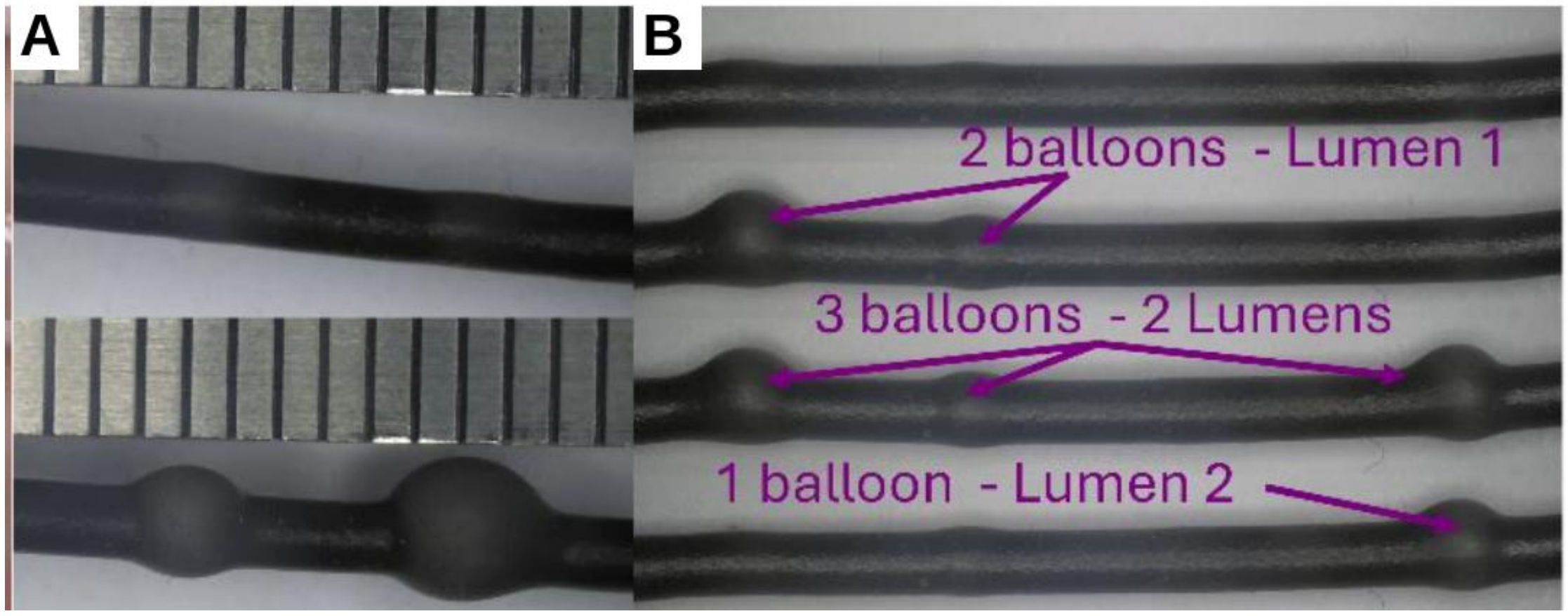

***Figure S13*** *Multi-balloon mSCR. (A) Two interconnected balloons with different geometric parameters, deflated (top) and inflated (bottom). (B) Three-balloon mSCR: two interconnected balloons sharing lumen 1 and one balloon with an independent lumen 2.*

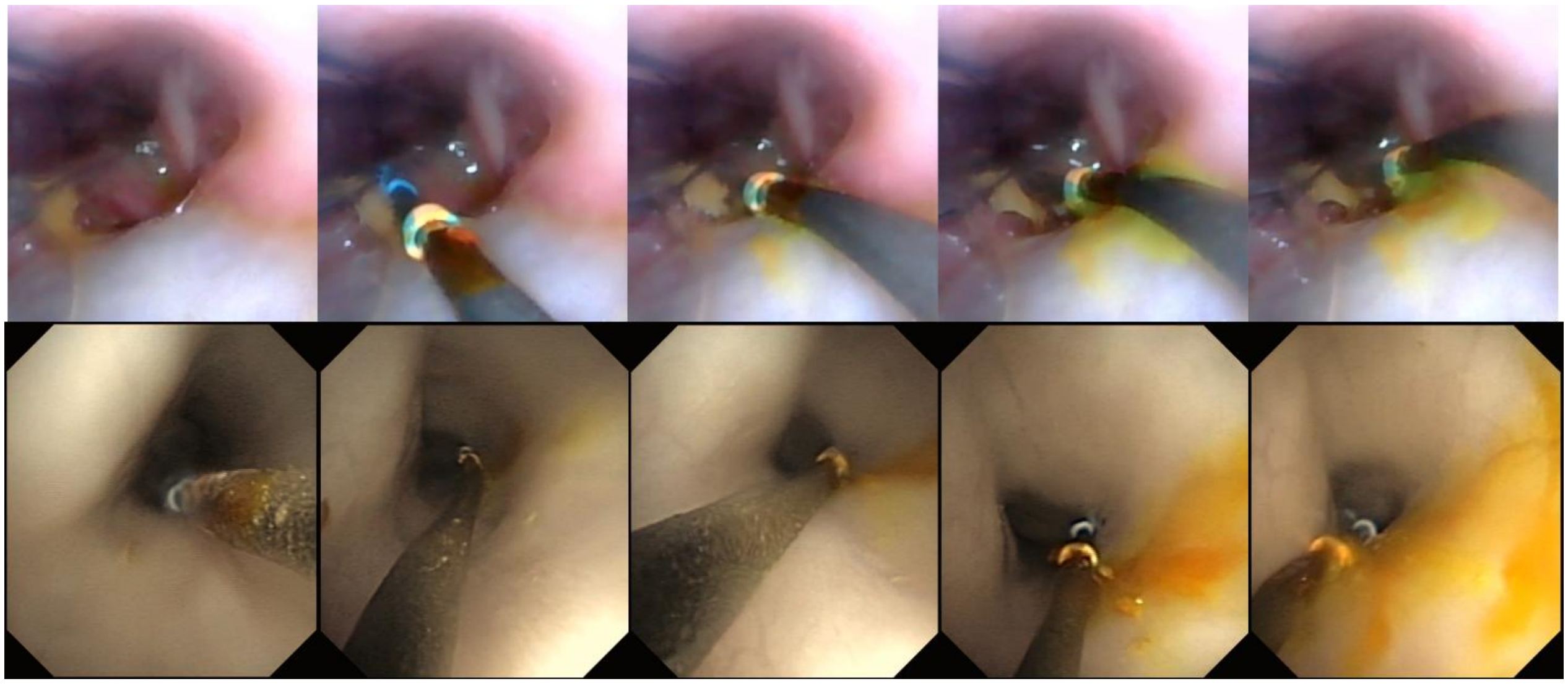

*Figure S14 In vivo demonstration of drug delivery within the esophagus (top) and lung (bottom).*

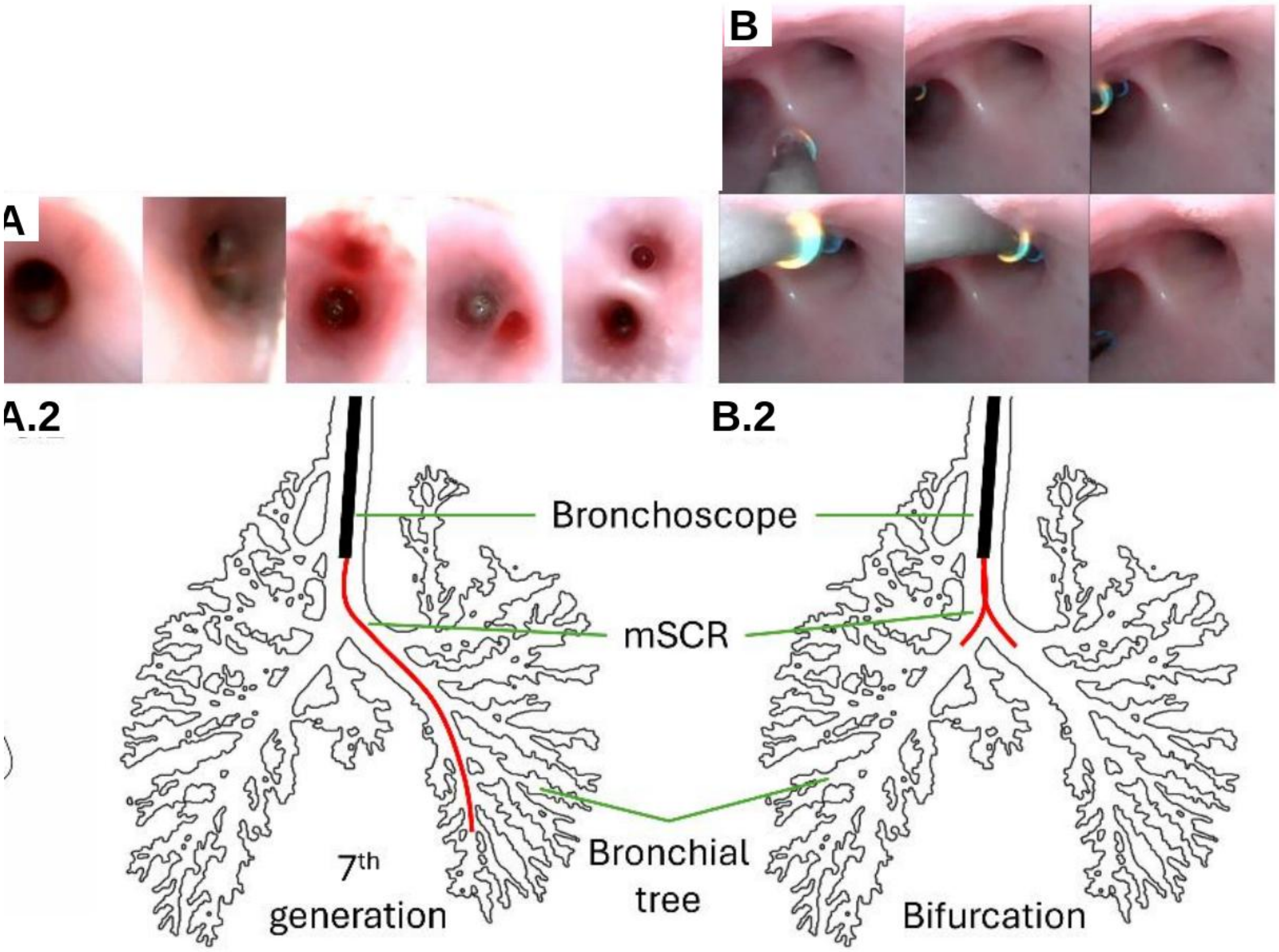

***Figure S15*** *Navigation in the porcine lung with the bronchoscope. (A) Navigation to seventh-generation segmental branches seen from the onboard camera, with schematic (A.2). (B) Sequential steering into two distinct airway branches seen from the bronchoscope, with schematic (B.2).*

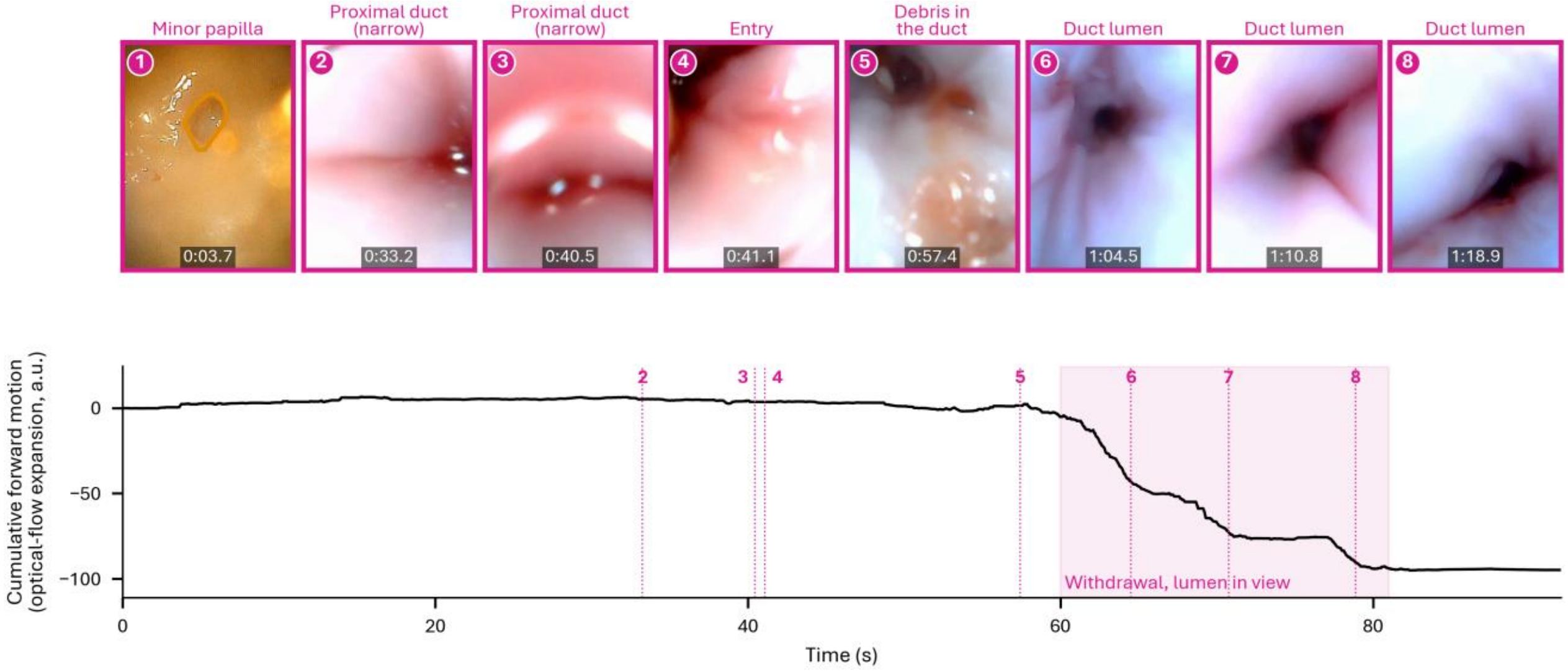

***Figure S16*** *Navigation in the pancreatic duct (same videos as Fig. 5D). Top: onboard frames 1 to 8 (times in min:s; frame 1 from a separate recording of the minor papilla). Bottom: cumulative forward motion estimated from the expansion of the optical flow (arbitrary units) over the 92-s video of the pancreatic duct. The lumen views (6 to 8) coincide with steady backward motion between 0:59.5 and 1:21.5, showing that they were acquired during withdrawal; forward insertion took place while the image was blurred.*

*Table S1 Procedural requirements of transpapillary pancreatobiliary access (R1 to R6), the corresponding design responses of the mSCR, and where each is evidenced.*

| **Requirement** | **Procedural need** | **Design response** | **Evidence** |
|---|---|---|---|
| R1 Outer diameter | Pass through the endoscope working channel and enter ducts that narrow from about 3.3 mm (pancreatic head) to about 1.5 mm (pancreatic tail) and 2.5 mm (cystic duct) | Monolithic 1.47-mm magneto-elastic body; all modules within this diameter | Fig. 1F, Fig. 2B |
| R2 Magnetic steerability | Align with the papilla and select branches under fields achievable at the bedside (8–16 mT) | Magnetized body; cost of each module measured; each module added after imaging and sensing costs 4.6° or less at 8 mT | Table 1, Fig. 2E, Fig. 4 |
| R3 Compliance during navigation | Gentle contact with the papilla and the duct wall | Soft silicone–NdFeB body; stiffness supplied only at the target by the balloon | Fig. 3 |
| R4 Onboard visualization | See the lumen and choose between branches instead of advancing blindly | Camera and LED fiber in molded guides; channel for flushing, aspiration, and insufflation | Fig. 5 |
| R5 Stabilization and action at the target | Hold position and act on tissue or deliver a payload | Magnetic anchoring balloon; compliant gripper on the camera footprint; ultrasound-triggered release shell | Fig. 3, Fig. 6, Fig. 7 |
| R6 Compatibility with standard endoscopes and the bedside workflow | Deploy through standard endoscopes without changing the endoscopic workflow | Field generated outside the patient by a bedside EPM; single joystick; tested with five endoscopes | Fig. 2, C and F; Table S3 |

| Endoscope | Model | Organ | Test demonstration | Task |
|---|---|---|---|---|
| Duodenoscope | Olympus JF-130 | Duodenum | In vivo | Operator-controlled and task-autonomous cannulation of the papilla |
| Gastroscope | Olympus GIF-H180 | Esophagus, Stomach, Duodenum | In vivo | Operator-controlled and task-autonomous cannulation of the papilla. Gripper and drug delivery in different environment |
| Colonoscope | Olympus CF-Q160 | Colon | In vivo | drug delivery in different environment |
| Colonoscope | Olympus Evis X1 – CV1500 | Duodenum | Ex vivo | Validation of the drug delivery tracking using narrow band imaging |
| Bronchoscope | Olympus BF-Q180-AC | Lung | In vivo | Navigation demonstration and drug delivery in different environment |

*Table S3 Standard medical endoscopes illustrating the cross-compatibility of our approach when applied in different settings.*

| **Name** | **Definition** | **Value** | **Unit** |
|---|---|---|---|
| $B_{epm}$ | Generated magnetic field | 8–16 | mT |
| | Workspace | 0.05 | $m^3$ |
| τ | Magnetic torque at the base of the mSCR for an applied field of 16 mT | $3.4 \times 10^{-4}$ | N·m |
| L | mSCR free (bending) length | 60 | mm |
| r | mSCR radius | 0.74 | mm |
| Br | Remanence of the magneto-elastic composite | 0.31 | T |
| E | Elastic modulus of the composite (from C1) | 3.04 | MPa |
| – | NdFeB:silicone mass ratio (volume fraction) | 3.7:1 (34.5%) | – |
| mEPM | EPM dipole moment (101.6 × 101.6 mm, N52) | 950 | $A \cdot m^2$ |
| $\mu_0$ | Magnetic permeability of free space | $4\pi \times 10^{-7}$ | $N \cdot A^{-2}$ |
| C1 | Constant related to the small-strain shear response | 0.507 | MPa |
| C2 | Constant related to intermediate-strain behavior | 0.0674 | MPa |
| C3 | Constant related to strain hardening at larger deformations | 0.0038 | MPa |

*Table S4 Summary of model parameters used in the mSCR design and control framework*